%% file: main.tex
\documentclass[runningheads]{llncs}

\usepackage{eccv}

\usepackage{eccvabbrv}

\usepackage{graphicx}
\usepackage{booktabs}

\usepackage[accsupp]{axessibility}  
\usepackage{booktabs}
\usepackage{multirow}
\usepackage{tabularx}
\usepackage[table]{xcolor}

\usepackage{algorithm}
\usepackage{algorithmic}
\usepackage{float}

\definecolor{clcolor}{HTML}{249087} 
\newcolumntype{Y}{>{\centering\arraybackslash}X}

\usepackage{tikz}

\usepackage{hyperref}

\usepackage{orcidlink}

\begin{document}

\title{From Open Loop to Closed Loop: A Test-Time Iterative Optimization Framework for Reference-Consistent Image Generation} 

\titlerunning{From Open Loop to Closed Loop}

\author{Baixuan Zhao\inst{1}\orcidlink{0000-0001-8929-8322} \and
Xinyu Zhang\inst{3}\orcidlink{0009-0007-3608-1673} \and
Huayu Zheng\inst{1}\orcidlink{0009-0002-4964-6761}  \and
Shuaicheng Liu\inst{3}\orcidlink{0000-0002-8815-5335}  \and \\
Xiongkuo Min\inst{1}\orcidlink{0000-0001-5693-0416}  \and
Guangtao Zhai\inst{1}\orcidlink{0000-0001-8165-9322}  \and
Xiaohong Liu\inst{1,2}\thanks{Corresponding author. Email: xiaohongliu@sjtu.edu.cn}\orcidlink{0000-0001-6377-4730}
}

\authorrunning{B.~Zhao et al.}

\institute{
Shanghai Jiao Tong University, Shanghai, China
\and
Shanghai Innovation Institute, Shanghai, China
\and
University of Electronic Science and Technology of China, Chengdu, Sichuan, China
}

\maketitle
\setcounter{footnote}{0}

\input{0_Abstract}
\input{1_Introduction}

\input{2_related_works}
\input{3_Preliminaries}
\input{4_Method}
\input{5_Experiment}
\input{6_Conclusion}

%
%
\newpage
\bibliographystyle{splncs04}
\bibliography{reference}

\newpage
\input{X_Appendix_in_main}

\end{document}

%% file: 0_Abstract.tex
\begin{abstract}
While controllable image generation has made significant strides by incorporating visual reference conditions, existing methods predominantly operate as open-loop systems. They inject control signals in a strictly feed-forward manner, failing to guarantee strict fidelity to the reference due to the absence of active feedback and error correction mechanisms. To address this fundamental limitation, we propose a novel test-time iterative optimization framework that reformulates reference-consistent generation as a closed-loop dynamic tracking problem. By treating the pre-trained generative model as a control plant, our framework employs a sensor-controller architecture driven by a modified Proportional-Integral-Derivative (PID) algorithm. This mechanism iteratively optimizes the latent control signals at test time based on the sensed discrepancy between the generated output and the reference target. Notably, this approach is entirely training-free, model-agnostic, and integrates seamlessly around existing diffusion pipelines. Extensive evaluations across ID-preserving, pose-controlled, and depth-controlled generation tasks validate the universality of our method. Empirical results demonstrate improvements over computation-matched open-loop baselines, achieving relative performance gains of up to 25.36\% for facial similarity, alongside spatial error reductions of up to 27.71\% for pose alignment and 28.50\% for depth consistency. More broadly, this work offers a new conceptual perspective: it demonstrates that controllable generation can be effectively managed as a dynamic feedback system, bringing the rigorous principles of classical control theory into the optimization of generative models. Code is available at \url{https://github.com/zzdrill/From-Open-Loop-to-Closed-Loop}.
\end{abstract}

%% file: 1_Introduction.tex
\section{Introduction}
\label{sec:intro}

Generative AI has reshaped digital content creation \cite{LuminaImage2025,LuminaMGPT2026,LuminaDiMOO2025}, with text-to-image diffusion models \cite{DALLE2022, StableDiffusion2022, flux2024} enabling high-fidelity image synthesis from textual descriptions.
To reduce the ambiguity of text-only prompts, recent controllable generation methods introduce visual references \cite{A2Edit2026}, enabling applications such as ID-preserving generation \cite{IPAdapter2023,instantid2024,photomaker2024}, pose/depth-controlled generation \cite{ControlNet2023, uni_controlnet, controlnext2024}, image restoration \cite{wang2025learning,jiang2023low}, aesthetic QR codes \cite{wu2024text2qr,cui2024face2qr,AnimateQR2025}, and virtual try-on \cite{dam2024time,choi2024improving}.

Despite these advances, many real-world applications require high consistency between generated outputs and reference conditions.
In ID-preserving portrait generation, even subtle feature drift may change the perceived identity.
In pose- or depth-controlled generation, small spatial deviations can lead to visible artifacts such as shifted limbs or distorted geometry.
Existing controllable generation methods usually inject reference conditions into the generative model in a feed-forward manner.
Once the condition is encoded, the model lacks an explicit mechanism to observe the generated result and correct deviations from the reference.
This motivates a control-theoretic question: \textbf{Can reference-consistent generation be formulated as a closed-loop dynamic tracking problem?}

\begin{figure*}[t]
    \centering
    \includegraphics[width=\textwidth]{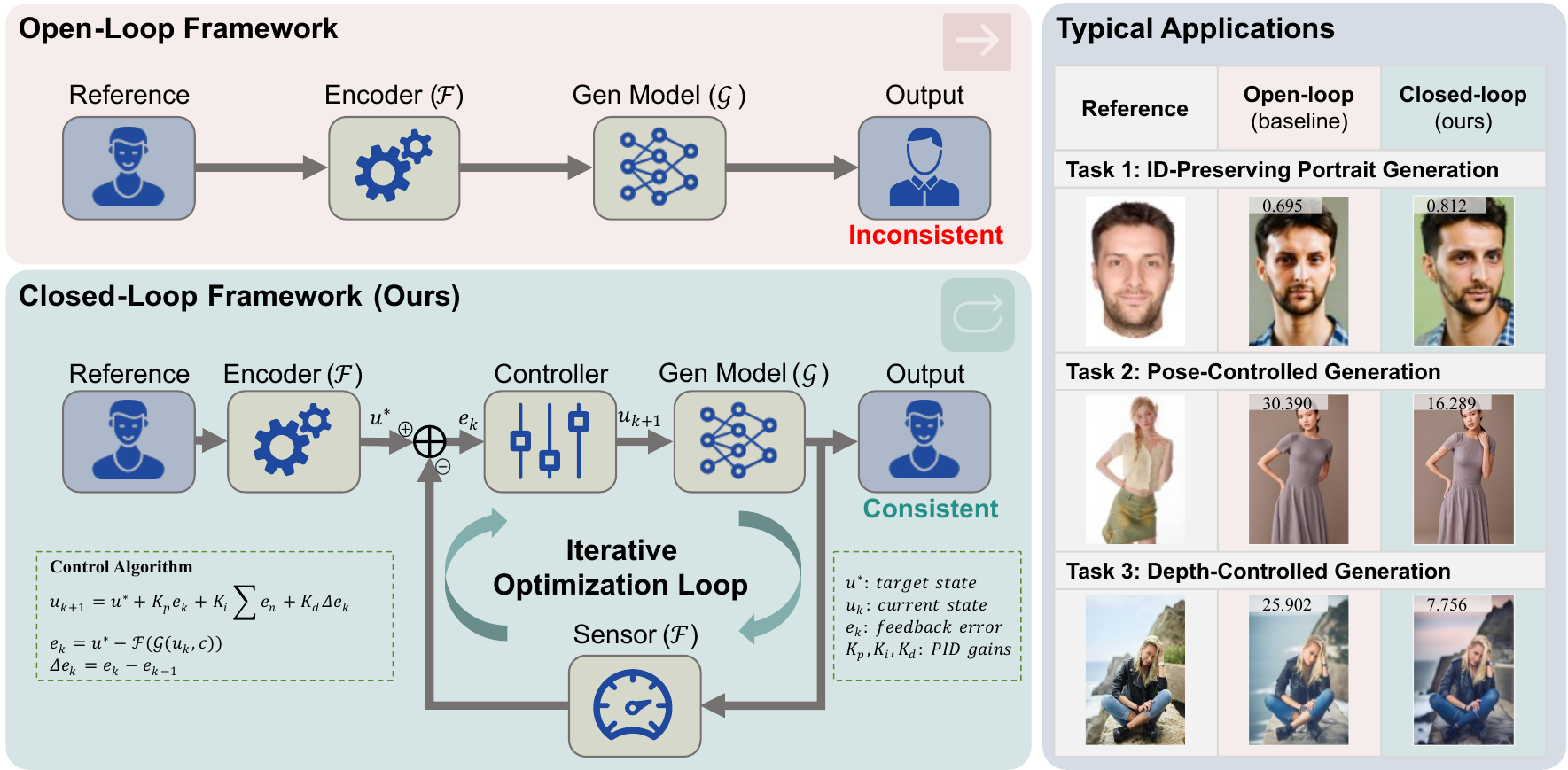}
    \caption{Overview of the proposed closed-loop optimization framework.
    We reformulate reference-consistent generation as feedback control over a generative model.
    The encoder and sensor are the same pre-trained module $\mathcal{F}$, reused in different roles for target extraction and feedback measurement.
    A modified PID controller iteratively updates the latent control input $u_k$ to reduce the generation error.
    This closed-loop optimization improves reference consistency over open-loop baselines across ID-, pose-, and depth-controlled generation.}
    \label{fig:main}
    \vspace{-10pt}
\end{figure*}

Under this formulation, current reference-consistent methods can be viewed as open-loop systems, where reference embeddings or spatial conditions are passed to the generative model without feedback verification.
To address this limitation, we propose a training-free test-time optimization framework for reference-consistent image generation, as illustrated in Figure~\ref{fig:main}.
Our framework reuses an off-the-shelf pre-trained module in two roles: as an encoder to extract the target state from the reference, and as a sensor to measure the generated state from the current output.
A modified PID controller then updates the latent control input according to the sensed tracking error, forming an explicit generation-and-check loop.
The framework is model-agnostic and can be plugged into existing diffusion pipelines without modifying their weights.

We evaluate the proposed framework on three representative tasks: (1) ID-preserving portrait generation, (2) pose-controlled generation, and (3) depth-controlled generation.
The main contributions of this work are summarized as follows:

\vspace{1mm}
\noindent -\textbf{Closed-loop formulation.}
We formulate reference-consistent generation as a closed-loop dynamic tracking problem, providing a control-theoretic perspective for improving alignment between generated outputs and visual references.

\vspace{1mm}
\noindent -\textbf{Training-free test-time framework.}
We introduce a lightweight plug-and-play optimization framework driven by a modified PID controller, which iteratively updates latent control inputs based on sensed tracking errors without additional training or model weight modification.

\vspace{1mm}
\noindent -\textbf{Broad applicability and empirical gains.}
Extensive experiments demonstrate consistent improvements across ID-preserving, pose-controlled, and depth-controlled generation.
Under computation-matched comparison with best-of-$N$ open-loop baselines, our framework achieves relative gains of up to 25.36\% in ID preservation and spatial error reductions of up to 27.71\% and 28.50\% for pose and depth control, respectively.

%% file: 2_related_works.tex
\section{Related Works}
\label{sec:related_works}

\noindent \textbf{Controllable Image Generation.}
Diffusion models have achieved remarkable progress in image synthesis, motivating a wide range of controllable generation methods \cite{StableDiffusion2022, flux2024,LayerT2V2026,FlowDirector2026}.
Adapter-based approaches, such as ControlNet \cite{ControlNet2023} and IP-Adapter \cite{IPAdapter2023}, introduce additional conditioning branches for flexible spatial or reference guidance \cite{zhang2025survey}.
Recent methods further improve parameter efficiency and multi-modal control \cite{controlnext2024, omini2025,LuminaDiMOO2025}, while ControlNet++ \cite{controlnet++} enhances reference consistency through training-time consistency losses.
Despite these advances, most methods remain open-loop at inference time: once the condition is encoded, the model does not actively observe and correct deviations from the reference.
Our work addresses this limitation by introducing a closed-loop test-time optimization framework for reference-consistent generation.

\vspace{1mm}
\noindent \textbf{Test-Time Optimization and Refinement.}
Cascaded architectures \cite{podell2023sdxl} and post-processing refiners \cite{wang2025learning,Li2024GRefine} are commonly used to improve resolution and visual fidelity.
Prompt optimization methods \cite{yang2024batch} iteratively refine textual inputs for better semantic coverage.
Training-free guidance methods such as UGD \cite{UGD}, FreeDoM \cite{FreeDoM}, and DAS \cite{DAS,DiffStega2024} also perform test-time optimization by injecting task-specific guidance or constraint signals into the sampling process.
However, these methods are mainly formulated as sampling-time guidance, where optimization is coupled with the denoising trajectory or task-specific energy functions.
In contrast, our method targets reference-consistent alignment through an explicit feedback loop.
A pre-trained sensor measures the tracking error between the reference and generated states, and a modified PID controller iteratively updates the latent control input, enabling generation-and-check refinement without additional training or independent post-processing.

\vspace{1mm}
\noindent \textbf{Control-Theoretic Interpretation of Diffusion.}
Another line of work interprets diffusion sampling through the lens of stochastic control \cite{song2021scorebased, richter2023optimal}, where the reverse denoising process can be viewed as an implicit regulation mechanism that preserves distributional fidelity.
Recent analyses also relate controllable adapters to delayed or low-frequency feedback systems \cite{controlnetxs2024}.
These studies mainly explain the internal dynamics of diffusion models or existing adapters, but they do not provide an external task-level feedback mechanism for correcting reference-specific semantic drift.
Our framework differs by placing a sensor-controller loop outside the generative model, turning reference-consistent generation into an explicit dynamic tracking problem.

%% file: 3_Preliminaries.tex
\section{Preliminaries}
\label{sec:preliminaries}

\subsection{Closed-loop Control System}
A closed-loop control system is an automated regulation architecture that continuously adjusts its behavior through active feedback. It typically consists of four essential components: (1) a \emph{target state} that defines the desired condition; (2) a \emph{controller} that calculates actions based on the \emph{tracking error} (the discrepancy between the target and actual state); (3) a \emph{controlled plant} representing the system requiring regulation (e.g., a generative model); and (4) a \emph{sensor} that monitors the actual state of the output to provide continuous feedback signals. Compared to open-loop counterparts, closed-loop architectures demonstrate superior robustness, higher precision, and effective rejection of external disturbances.

\subsection{PID Control Algorithm}
Generative models often exhibit a black-box nature with extremely high-dimensional complexity, making it intractable to derive explicit transfer functions or state-space equations required by modern optimal control methods like LQR \cite{astrom2010feedback} or MPC \cite{camacho2013model}. In contrast, the Proportional-Integral-Derivative (PID) \cite{borase2021review} algorithm is a fundamentally model-free controller that requires no prior knowledge of the plant dynamics, making it an ideal candidate for iterative generative refinement. 

PID achieves precise regulation through three synergistic terms: the \emph{proportional} (P) term provides an instantaneous response to the current error, the \emph{integral} (I) term accumulates past errors to eliminate steady-state offsets, and the \emph{derivative} (D) term predicts future trends to suppress overshoot. For discrete-time systems, the standard positional PID control signal is expressed as:
\begin{equation}
u_{k+1} = K_p e_k + K_i T_s \sum_{n=0}^{k} e_n + K_d \frac{e_k - e_{k-1}}{T_s},
\label{eq:pid_disc}
\end{equation}
where $k$ denotes the discrete step index; $u_{k+1}$ is the control signal computed for the $(k+1)$-th step; $e_k$ represents the tracking error between the target state and the system output at step $k$; and $K_p$, $K_i$, and $K_d$ correspond to the proportional, integral, and derivative gains, respectively. The variable $T_s$ denotes the sampling period.

%% file: 4_Method.tex
\section{Method}
\label{sec:method}

\subsection{Framework Overview}
Following the principles of control theory, we formulate the reference-consistent generation task as a discrete-time dynamic tracking problem. Our closed-loop framework (illustrated in Figure \ref{fig:main}) comprises five key components: \textit{reference}, \textit{encoder}, \textit{controller}, \textit{controlled plant}, and \textit{sensor}. 

In this architecture, the reference image $I^*$ defines the target state $u^* = \mathcal{F}(I^*)$. A shared feature extractor $\mathcal{F}$ is employed with dual functionality: as the \textit{encoder}, it maps the reference image to the target state; as the \textit{sensor}, it monitors the generated image $I_k$ at step $k$ to provide feedback on the actual state directly via $\mathcal{F}(I_k)$. The \textit{controller} quantifies the tracking error $e_k = u^* - \mathcal{F}(I_k)$ and computes the updated \textit{control input} $u_{k+1}$ via the control algorithm $\mathcal{C}$. The generative model, integrated with a controllable adapter (denoted as $\mathcal{G}$), acts as the \textit{controlled plant} to synthesize images based on the latent control input $u$ and text prompt $c$.

As shown in Equation (\ref{eq:close_loop_prog}), the system operates through successive generation-and-check cycles. In each iteration $k$, the actual state is observed directly from the generated image as $\mathcal{F}(I_k)$. The controller then updates the control input $u_{k+1}$ to compensate for the sensed discrepancy. Specifically, given the target state $u^* = \mathcal{F}(I^*)$, the iterative optimization loop is formally defined as:
\begin{equation}
\left\{
\begin{aligned}
    I_k &= \mathcal{G}(u_k, c), \\
    e_k &= u^* - \mathcal{F}(I_k), \\
    u_{k+1} &= \mathcal{C}(u^*, e_k).
\end{aligned}
\right.
\label{eq:close_loop_prog}
\end{equation}
The subsequent output $I_{k+1} = \mathcal{G}(u_{k+1}, c)$ thus aligns progressively closer with the reference.

\vspace{1mm}
\noindent \textbf{Discussion on Inference Overhead.} It is worth noting that each iteration $k$ in Equation (\ref{eq:close_loop_prog}) executes a complete diffusion denoising process. While this inherently increases the overall inference time compared to open-loop baselines, it trades a manageable latency for a paramount engineering advantage: \textit{zero intrusion}. By treating the base model as a black box, our framework requires no modifications to the original generation pipeline. Implementing this lightweight wrapper merely involves adding a few lines of code to establish the external feedback loop.

\subsection{Controller Formulation}
Classical dynamic systems are characterized by their output states being dependent on both instantaneous inputs and cumulative historical states. Conventional PID algorithms are specifically designed for such dynamics, regulating the system by calculating control signals solely based on the tracking error. In contrast, image generation models behave as static (memoryless) systems: the output depends exclusively on the current control input $u_k$ and prompt $c$, with no inherent mechanism to retain internal states across inference iterations. 

Consequently, a traditional PID controller---where the control signal diminishes as the error approaches zero---is fundamentally unsuitable for generative tasks, as the model requires a persistent baseline signal to maintain the presence of the target subject. To address this mismatch, we propose a \textit{Modified PID Algorithm} that introduces the target state as a persistent bias term at each iteration. Given that the condition input space and the sensory feedback space are aligned in our tasks, we directly utilize the target state $u^* = \mathcal{F}(I^*)$ as the persistent base control input.

By reformulating the control task as a residual adjustment relative to $u^*$, the system is naturally initialized by setting $u_1 = u^*$. This initial pass generates the base image $I_1$, yielding the first tracking error $e_1$. With the sampling period normalized to unit steps, the modified control law for all $k \geq 1$ is formulated as:
\begin{equation}
\left\{
\begin{aligned}
    e_k &= u^* - \mathcal{F}(\mathcal{G}(u_k, c)), \\
    u_{k+1} &= u^* + K_p e_k + K_i \sum_{n=1}^{k} e_n + K_d (e_k - e_{k-1}),
\end{aligned}
\right.
\label{eq:pid_mod}
\end{equation}
where $e_0 \equiv 0$ is defined for the initial derivative calculation. In this formulation, the PID terms act as a residual compensator, refining the base signal $u^*$ through historical feedback to keep $u_{k+1}$ grounded in the target condition while driving the output toward asymptotic convergence.

\subsection{Task Formulations}
Our closed-loop framework serves as a universal wrapper that can be instantiated across various controllable generation tasks. The versatility of the framework relies on defining appropriate state spaces and sensors for specific conditions.

\vspace{1mm}
\noindent \textbf{ID-preserving portrait generation.} In this task, the control objective is to maintain high fidelity to the subject's identity. We formulate the state space in the latent facial embedding domain, $\mathbb{R}^d$. The sensor $\mathcal{F}_{id}$ is defined as a facial recognition encoder. Given the target reference embedding $u^*_{id} = \mathcal{F}_{id}(I^*)$, the tracking error at step $k$ is quantified as the element-wise difference: $e_k = u^*_{id} - \mathcal{F}_{id}(I_k) \in \mathbb{R}^d$. This error vector drives the controller to iteratively optimize the identity-conditioned base signal along each feature dimension.

\vspace{1mm}
\noindent \textbf{Pose-controlled generation.} To achieve precise structural alignment, the state space is defined in the 2D spatial coordinate domain, $\mathbb{R}^{H \times W \times 2}$. The sensor $\mathcal{F}_{pose}$ functions as a pose estimator, extracting structural keypoint representations. Given the target reference pose map $u^*_{pose} = \mathcal{F}_{pose}(I^*)$, the tracking error is formulated as the spatial deviation: $e_k = u^*_{pose} - \mathcal{F}_{pose}(I_k) \in \mathbb{R}^{H \times W \times 2}$. The resulting control signals are subsequently added as spatial residuals to update the explicit pose guidance map in each iteration.

\vspace{1mm}
\noindent\textbf{Depth-controlled generation.} For depth-constrained tasks, the state space operates in the dense pixel-wise geometric domain, $\mathbb{R}^{H \times W}$. The sensor $\mathcal{F}_{depth}$ acts as a monocular depth estimator. Given the target reference depth map $u^*_{depth} = \mathcal{F}_{depth}(I^*)$, the tracking error evaluates the pixel-level deviations against the estimated depth of the current output: $e_k = u^*_{depth} - \mathcal{F}_{depth}(I_k) \in \mathbb{R}^{H \times W}$. The controller iteratively compensates for these geometric deviations, ensuring the output structure progressively converges to the target spatial configuration.

%% file: 5_Experiment.tex
\section{Experiments}
\label{sec:experiment}

\subsection{Experimental Setup}
\label{subsec:setup}

\vspace{-8mm}
\begin{table}[h]
\begin{minipage}[c]{0.52\textwidth}
\textbf{Implementation Details.} All experiments are conducted on an NVIDIA GeForce RTX 4090 GPU. As shown in Table~\ref{tab:params}, PID coefficients, resolutions, and iteration numbers are task-specific: ID-preserving uses a fixed $1024^2$ resolution for 20 iterations, while pose/depth control uses adaptive aspect-ratio scaling with the shorter edge set to 1024 pixels for 15 iterations.

\end{minipage}\hfill
\begin{minipage}[c]{0.45\textwidth}
\centering
\caption{Parameter configurations.}
\label{tab:params}

\renewcommand{\arraystretch}{1.3} 

\resizebox{\linewidth}{!}{
\begin{tabular}{lccccc}
\toprule
Task & $K_p$ & $K_i$ & $K_d$ & Res. & Iter. \\ \midrule
ID & 0.30 & 0.05 & 0.01 & $1024^2$ & 20 \\
Pose  & 0.05 & 0.02 & 0.01 & Adaptive & 15 \\
Depth & 0.01 & 0.005& 0.005& Adaptive & 15 \\
\bottomrule
\end{tabular}
}
\end{minipage}
\end{table}
\vspace{-6mm}

\vspace{1mm}
\noindent \textbf{Datasets.} For ID-preserving generation, we evaluate on: (1) \textit{Web100}, a multi-reference set containing 130 non-celebrity identities with 2--5 photos each; (2) \textit{CelebA300}, comprising 300 random identities from CelebA-HQ \cite{CelebA}. For pose and depth control, we utilize a filtered subset of 477 single-person photos from a public repository\footnote{\url{https://huggingface.co/datasets/raulc0399/open_pose_controlnet}, accessed on March 5, 2026.}.

\vspace{1mm}
\noindent \textbf{Metrics.}
We evaluate our framework across three primary dimensions.
To avoid sensor-evaluator coupling, we use different models for feedback and evaluation:
InsightFace\footnote{\url{https://github.com/deepinsight/insightface}, accessed on March 5, 2026.} ResNet100@Glint360K / ResNet50@WebFace600K for ID,  
OpenPose / DWPose for pose \cite{openpose,dwpose},
and MiDaS / ZoeDepth for depth \cite{midas,zoedepth}, respectively.
Specifically, the first model in each pair is used as the feedback sensor, while the second one is used only for evaluation.

\textit{Consistency.}
We measure the alignment between the generated output and the target reference.
For ID-preserving tasks, we compute Facial Similarity via the cosine similarity between identity embeddings:
\begin{equation}
S_{face} = \cos(\mathbf{v}^*, \mathbf{v}^{gen}),
\end{equation}
where $\mathbf{v}$ denotes the identity embedding vector.

For the pose-controlled generation task, the consistency metric is evaluated using the Mean Per Joint Position Error (MPJPE).
We formulate this as the average $L_2$ distance between corresponding structural keypoints:
\begin{equation}
E_{pose} = \frac{1}{M} \sum_{m=1}^{M} \|\mathbf{p}_{m}^* - \mathbf{p}_{m}^{gen}\|_2,
\end{equation}
where $M$ represents the total number of matched keypoints, and $\mathbf{p}_{m}^*, \mathbf{p}_{m}^{gen} \in \mathbb{R}^2$ denote the 2D spatial coordinates of the $m$-th joint in the target and generated pose maps, respectively.

For the depth-controlled generation task, consistency is assessed via the Mean Absolute Error (MAE) of corresponding pixels in the depth maps:
\begin{equation}
E_{depth} = \frac{1}{H \times W} \sum_{i=1}^{H} \sum_{j=1}^{W} |d_{i,j}^* - d_{i,j}^{gen}|,
\end{equation}
where $H$ and $W$ represent the height and width of the image, while $d$ denotes the estimated depth value at a specific pixel coordinate.

\textit{Diversity.}
To quantify facial variation and address the generic ``pose-copying'' issue in ID tasks, we propose Structure Diversity ($D_{sd}$).
This metric measures the structural deviation of the generated image from the reference set.
Specifically, we first apply coordinate normalization $\mathcal{N}(\cdot)$ to eliminate scale and translation differences.
We then calculate the mean Euclidean distance over all $K$ facial landmarks, averaged across the $N$ reference images:
\begin{equation}
D_{sd} = \frac{1}{N} \sum_{n=1}^{N} \left( \frac{1}{K} \sum_{j=1}^{K} \|\mathcal{N}(\mathbf{p}_{j}^{gen}) - \mathcal{N}(\mathbf{p}_{j, n}^{ref})\|_2 \right).
\end{equation}

\textit{General Quality.}
Semantic similarity is evaluated via CLIP-I \cite{clip} and DINO \cite{dino}, while perceptual generation quality is assessed via Q-Align \cite{q_align}.

\subsection{Qualitative Evaluation}

\begin{figure}[t]
    \centering
    \includegraphics[width=\textwidth]{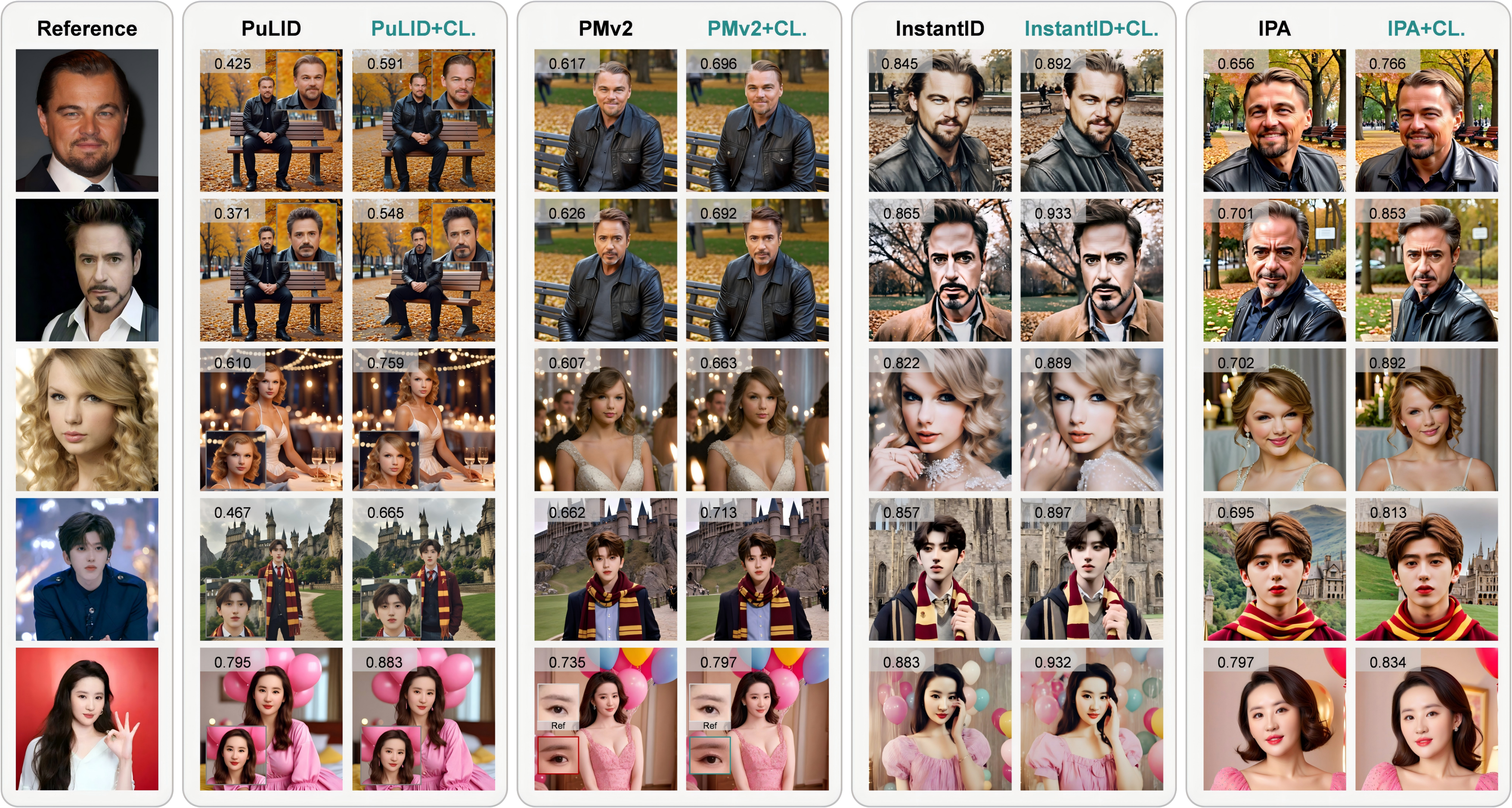}
    \caption{Qualitative comparisons of ID-preserving portrait generation. We integrate our closed-loop optimization \textcolor{clcolor}{(+CL.)} into four representative baselines. The results demonstrate noticeable improvements in both identity fidelity and fine-grained detail preservation compared to the standard open-loop generations.}
    \label{fig:vis_portrait_comp}
    \vspace{-10pt}
\end{figure}

\textbf{ID-Preserving Generation.} 
To evaluate the effectiveness of our approach, we integrate the proposed closed-loop optimization with four state-of-the-art ID-preserving generation baselines: PuLID \cite{pulid2024}, PhotoMaker v2 (PMv2) \cite{photomaker2024}, InstantID \cite{instantid2024}, and IP-Adapter-FaceID-Portrait (IPA)\footnote{\url{https://huggingface.co/h94/IP-Adapter-FaceID}, accessed on March 5, 2026.}. As illustrated in Fig. \ref{fig:vis_portrait_comp}, the visual comparisons indicate that the closed-loop integration \textcolor{clcolor}{(+CL.)} consistently improves identity fidelity across diverse base architectures. Supported by the facial similarity scores, our method facilitates a closer alignment with the reference subjects. Furthermore, as highlighted by the local zoom-in patches, the optimization process recovers structural proportions and better preserves fine-grained facial details (e.g., eye characteristics and subtle skin textures) that are occasionally smoothed out by standard open-loop methods. Overall, by establishing an effective closed-loop system, our approach significantly enhances identity consistency without compromising the underlying photorealistic generation quality, proving its plug-and-play adaptability across fundamentally different adapter designs.

\begin{figure}[htbp]
    \centering
    \includegraphics[width=\columnwidth]{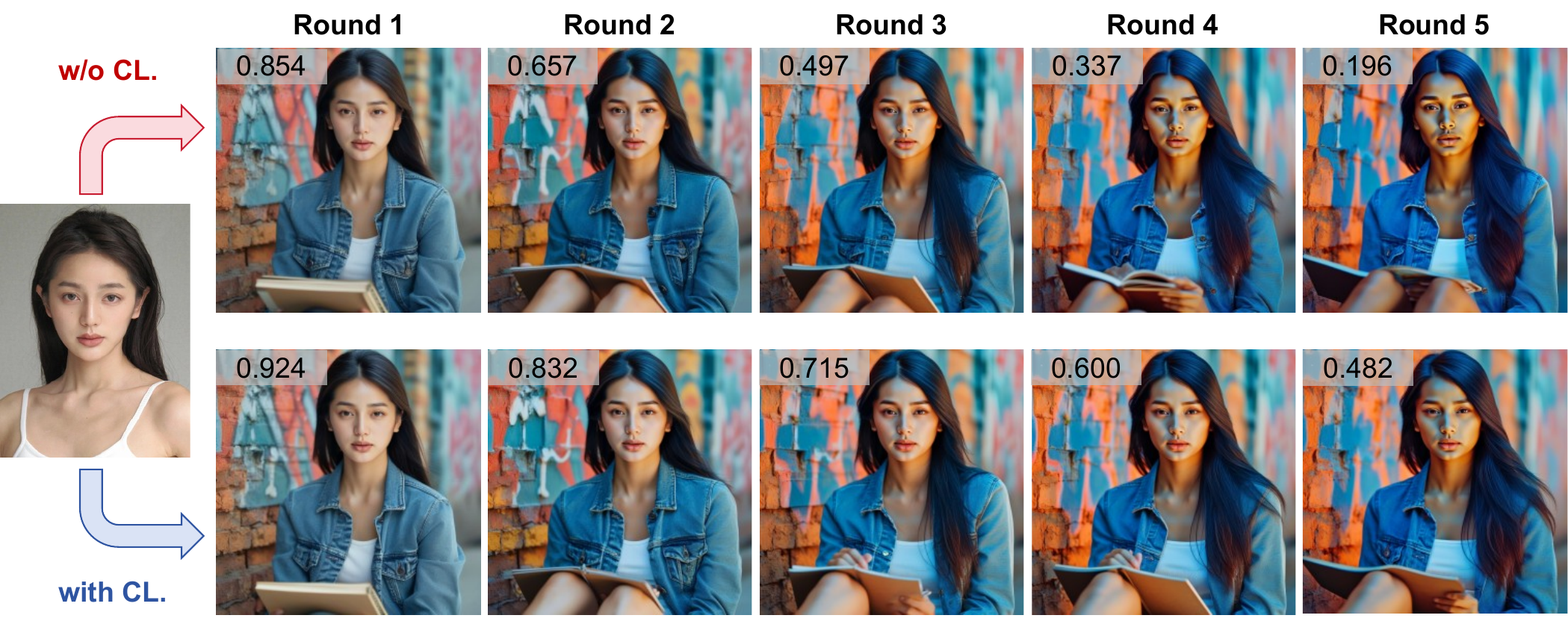}
    \caption{Recurrent generation experiment. The output of the previous round serves as the reference for the next. Our closed-loop optimization effectively mitigates identity degradation across iterations. Facial similarity ($\uparrow$) are provided below each image.}
    \label{fig:vis_portrait_recurrent}
    \vspace{-10pt}
\end{figure}

We further evaluate the robustness of our framework through a \textit{recurrent generation experiment}, where the output of the current iteration serves as the reference image for the subsequent generation. While standard feed-forward methods typically suffer from rapid identity degradation due to error accumulation across iterations, our closed-loop approach effectively mitigates this issue. As shown in Fig. \ref{fig:vis_portrait_recurrent}, the feedback mechanism maintains high facial similarity across multiple rounds, demonstrating enhanced stability against identity drift.

\begin{figure}[t]
    \centering
    
    \begin{subfigure}[b]{0.325\columnwidth}
        \centering
        \includegraphics[width=\columnwidth]{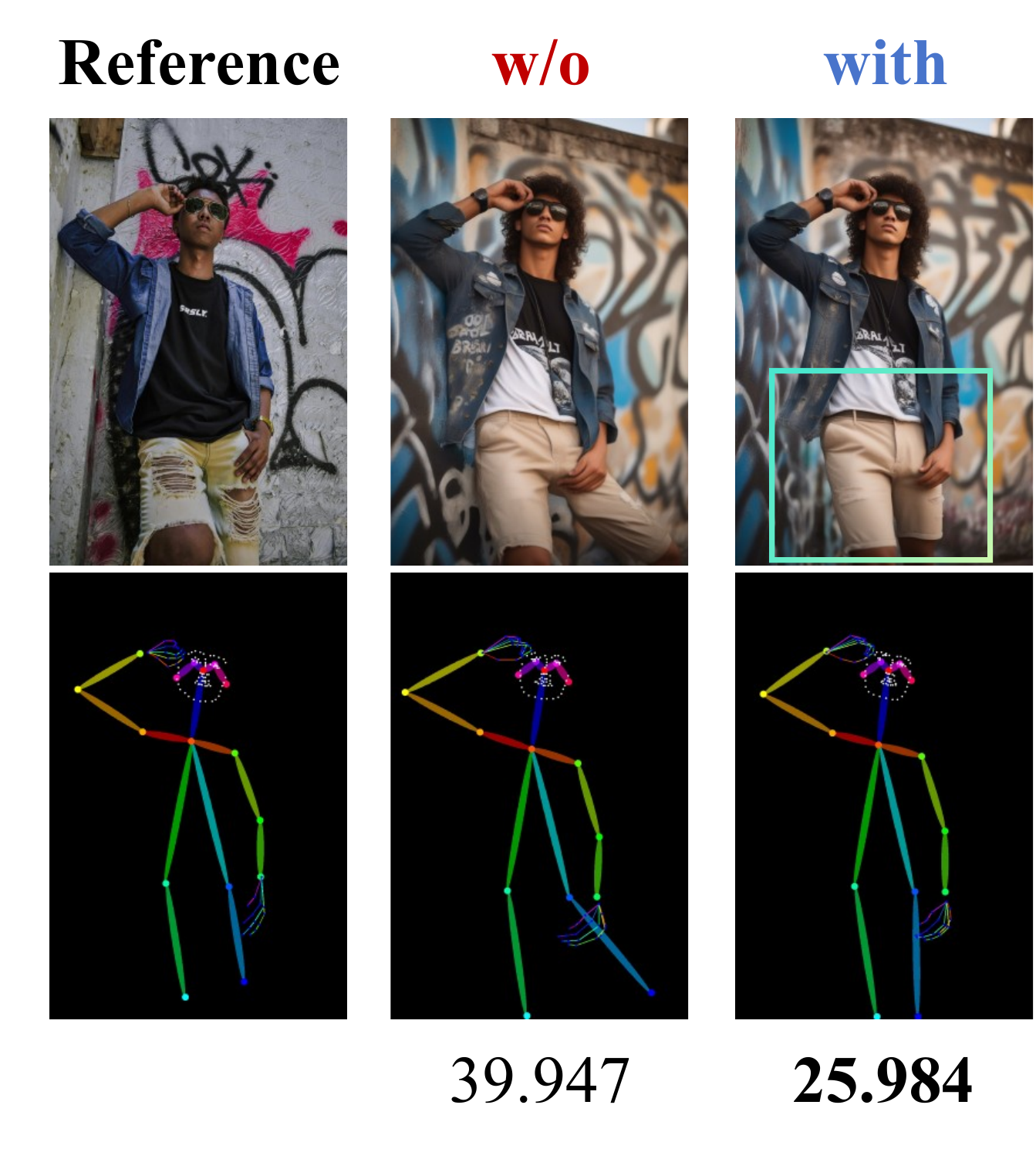} 
    \end{subfigure}
    \hfill
    \begin{subfigure}[b]{0.325\columnwidth}
        \centering
        \includegraphics[width=\columnwidth]{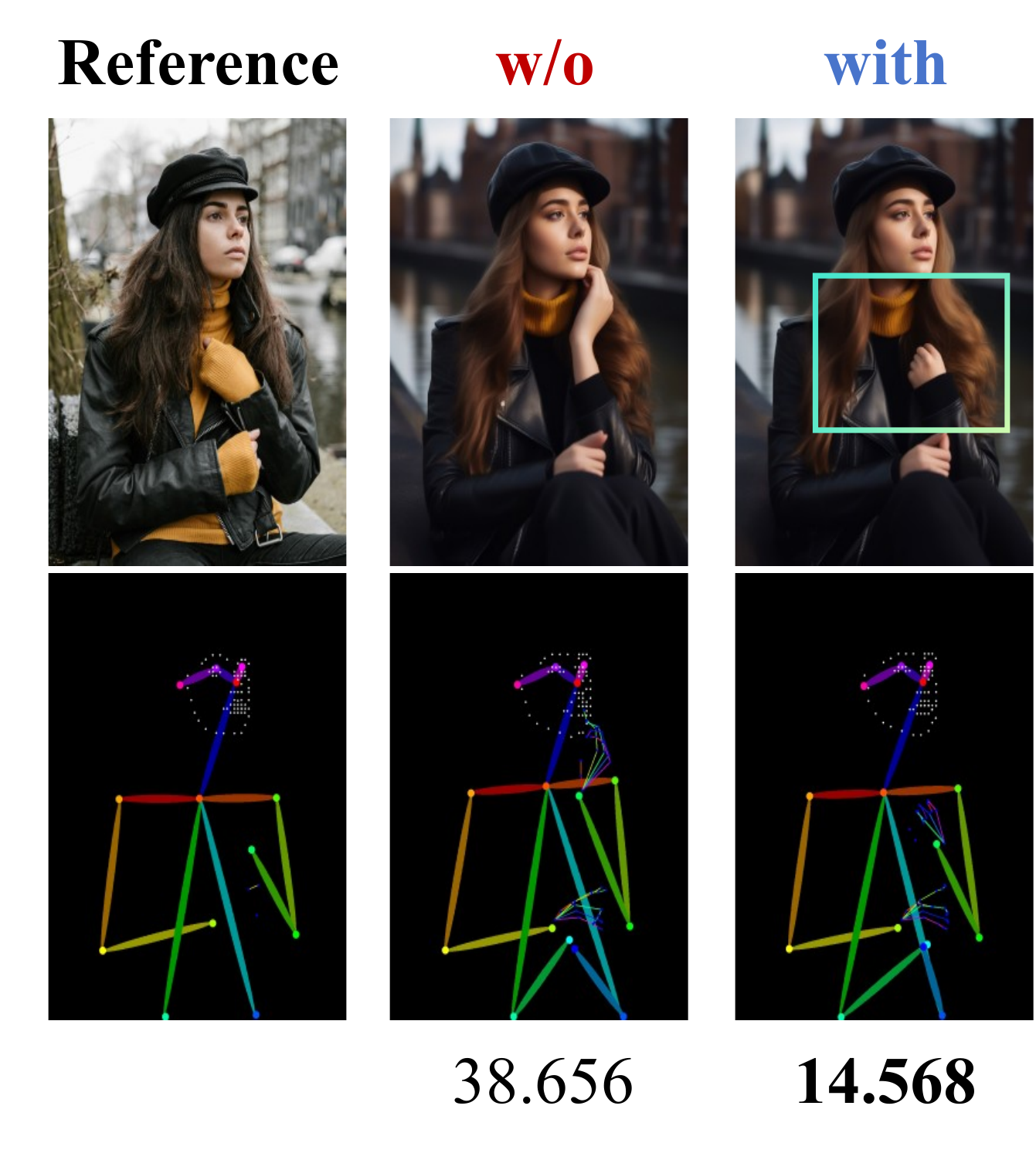} 
    \end{subfigure}
    \hfill
    \begin{subfigure}[b]{0.325\columnwidth}
        \centering
        \includegraphics[width=\columnwidth]{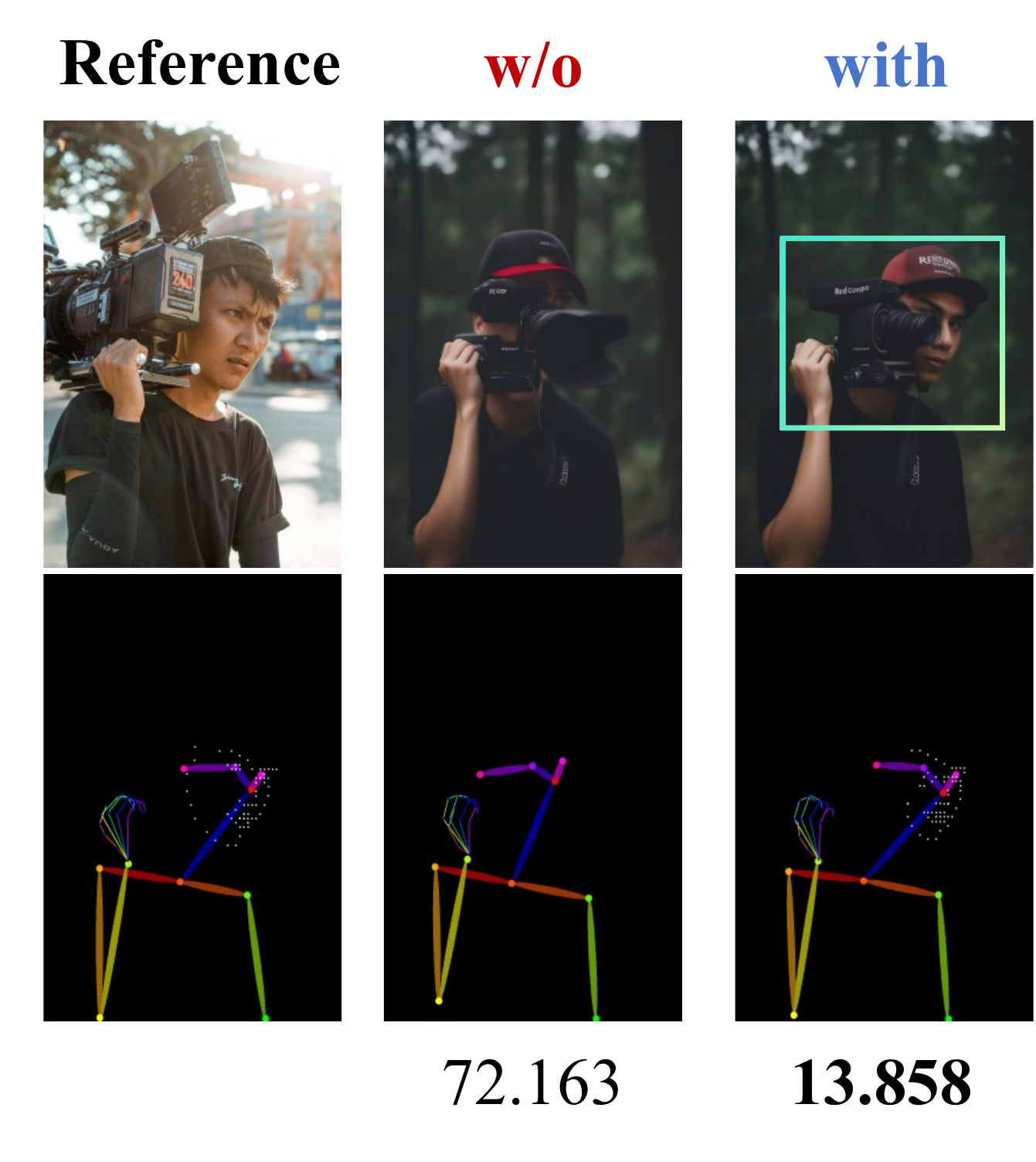} 
    \end{subfigure}
    \begin{subfigure}[b]{0.325\columnwidth}
        \centering
        \includegraphics[width=\columnwidth]{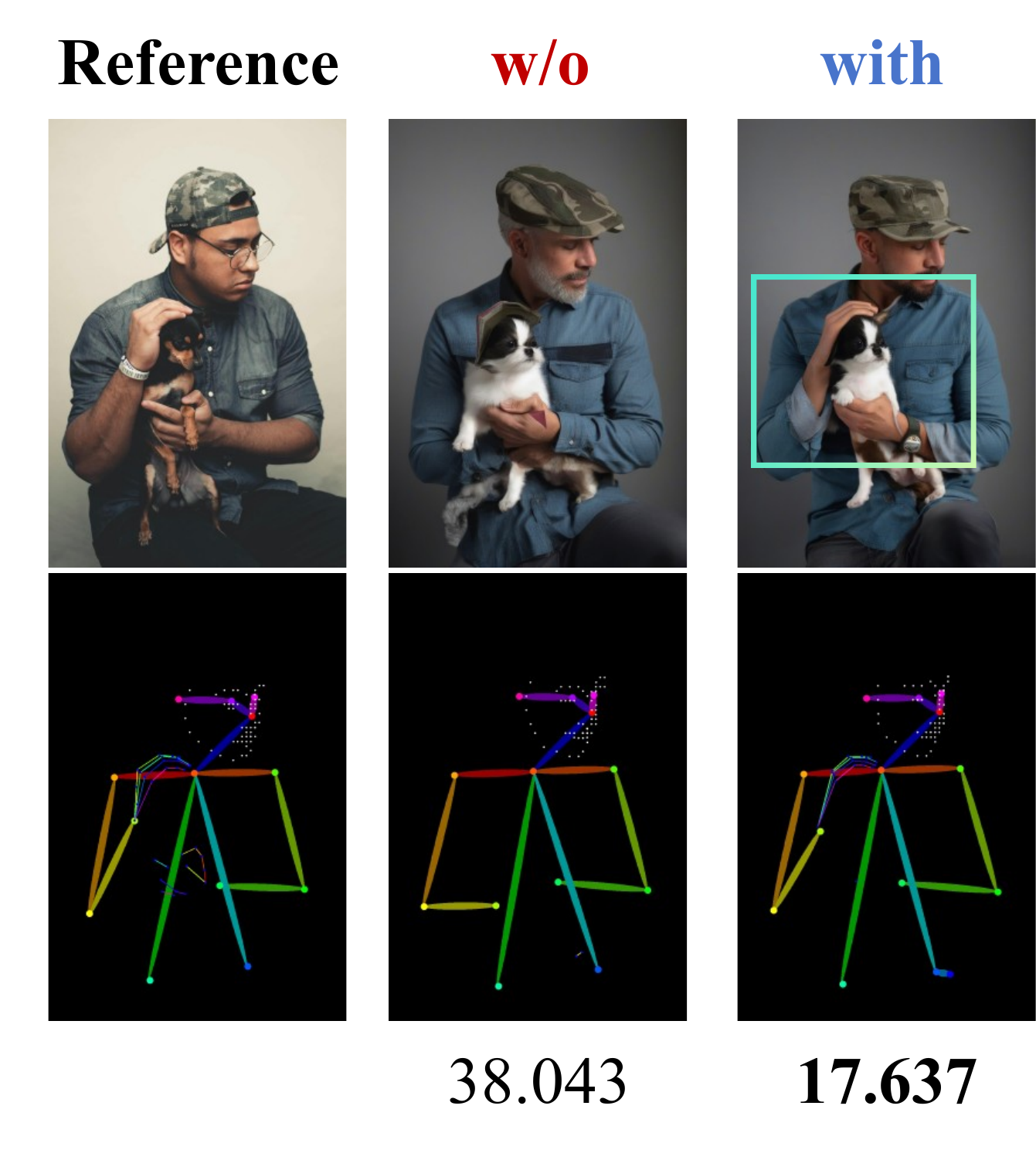} 
    \end{subfigure}
    \hfill
    \begin{subfigure}[b]{0.325\columnwidth}
        \centering
        \includegraphics[width=\columnwidth]{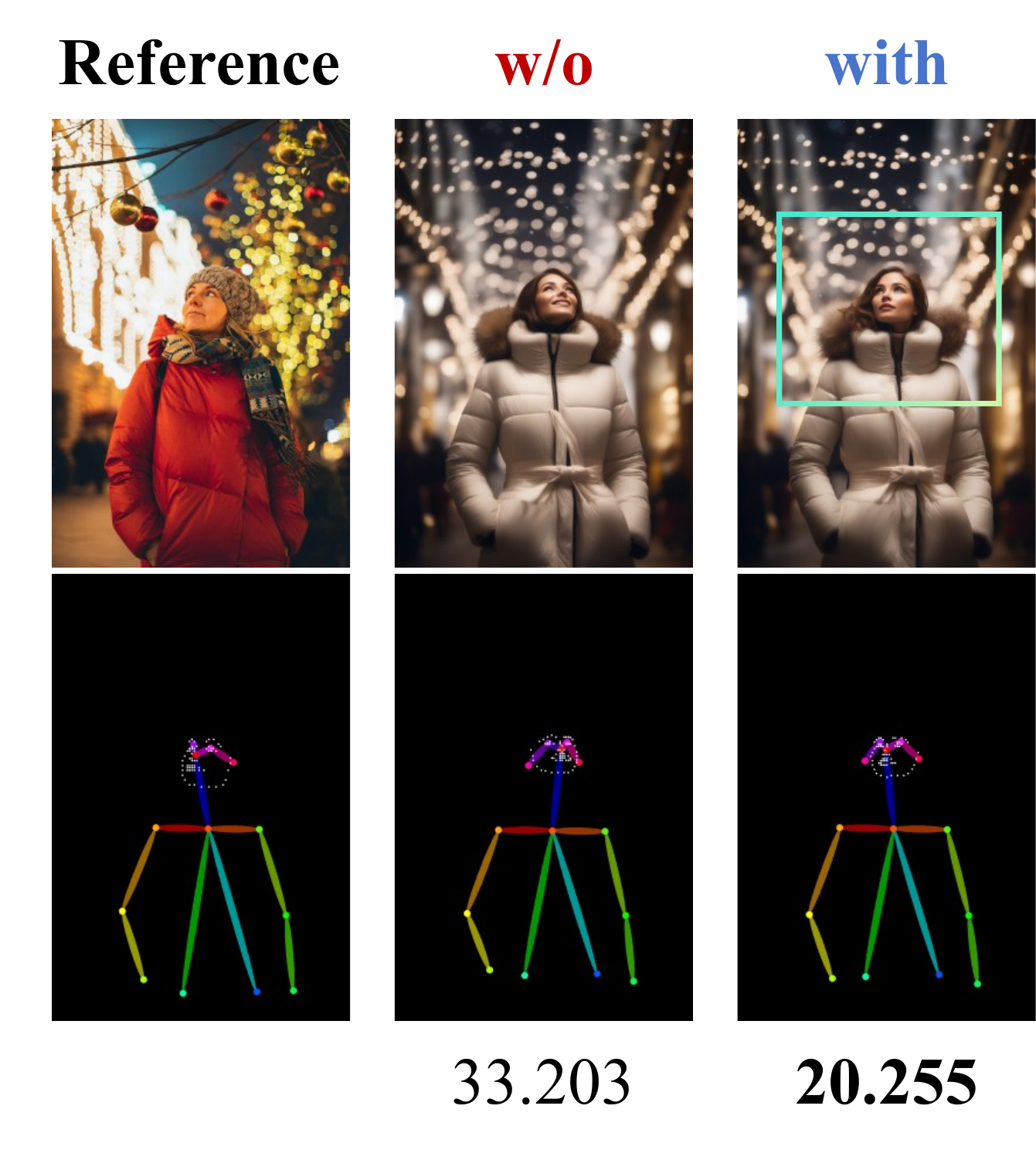} 
    \end{subfigure}
    \hfill
    \begin{subfigure}[b]{0.325\columnwidth}
        \centering
        \includegraphics[width=\columnwidth]{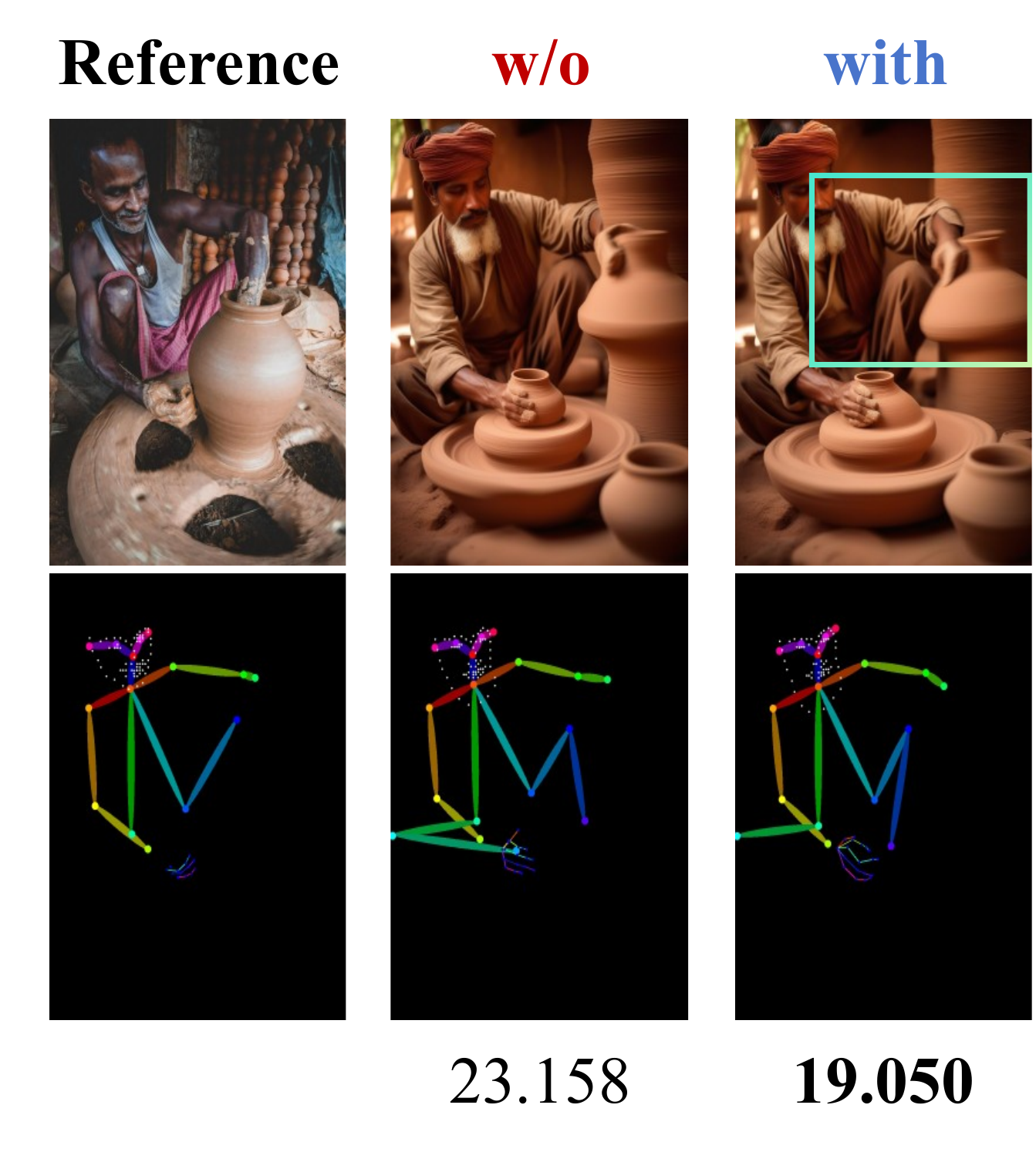} 
    \end{subfigure}
    
    \caption{Qualitative comparisons for pose-controlled generation. Compared to the standard open-loop baseline (\textbf{w/o}), our closed-loop optimization (\textbf{with}) drastically rectifies structural misalignments and complex articulations (highlighted in boxes). Extracted skeletal maps and the corresponding Mean Per Joint Position Error (MPJPE, $\downarrow$) are provided below each result to quantify the geometric improvements.}
    \vspace{-10pt}
    \label{fig:vis_pose}
\end{figure}

\begin{figure}[t]
    \centering
    
    \begin{subfigure}[b]{0.49\columnwidth}
        \centering
        \includegraphics[width=\columnwidth]{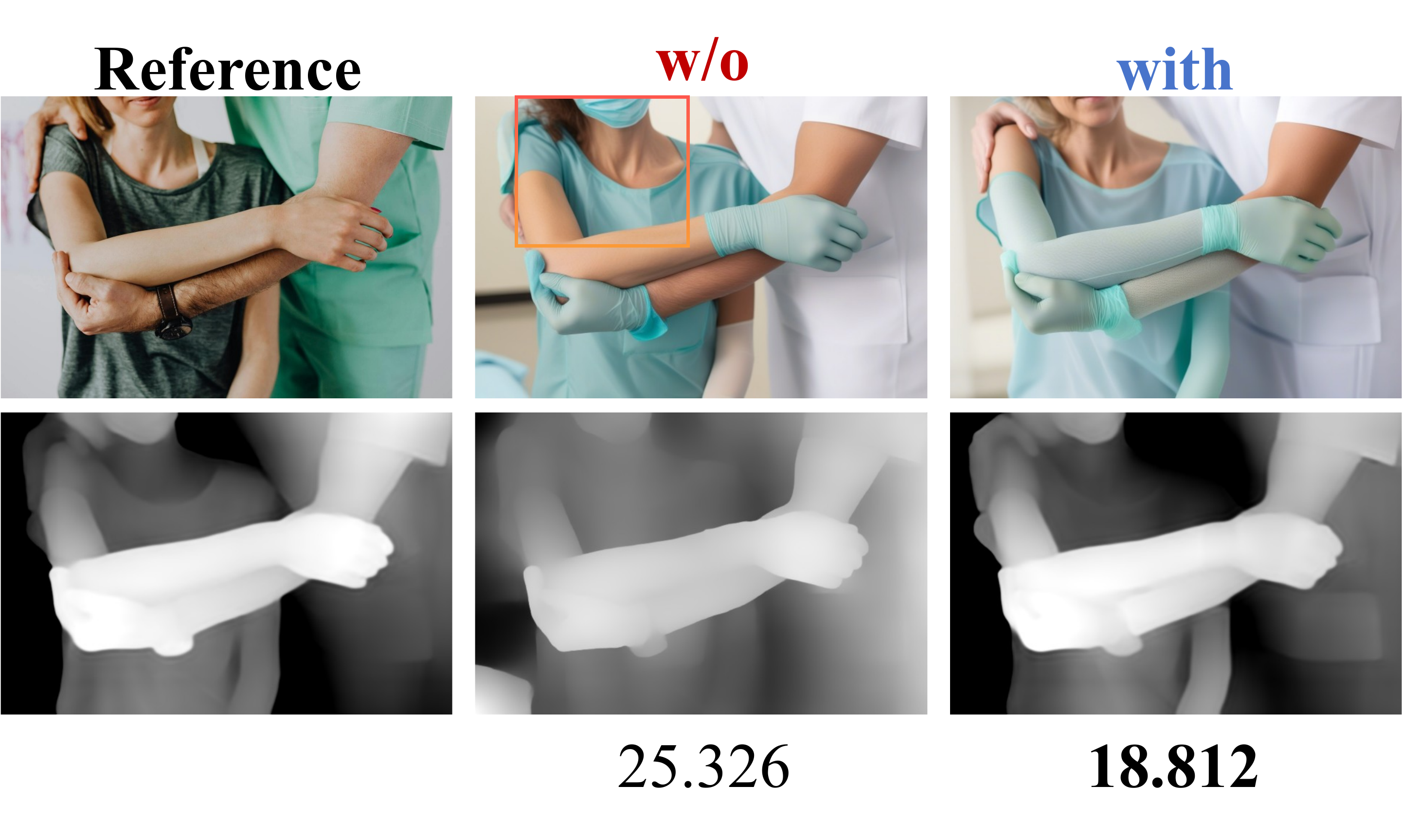} 
    \end{subfigure}
    \hfill
    \begin{subfigure}[b]{0.49\columnwidth}
        \centering
        \includegraphics[width=\columnwidth]{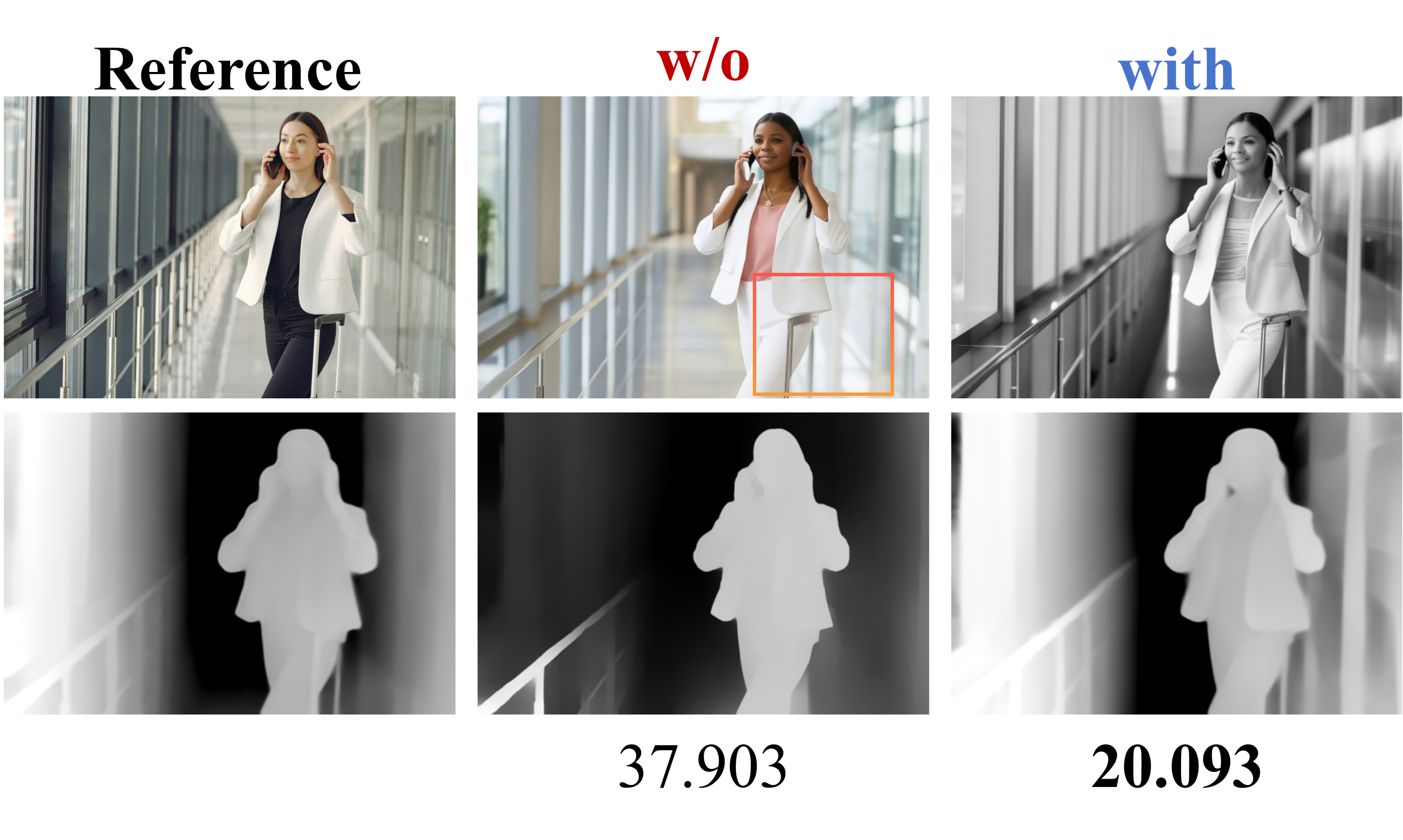} 
    \end{subfigure}
    \begin{subfigure}[b]{0.325\columnwidth}
        \centering
        \includegraphics[width=\columnwidth]{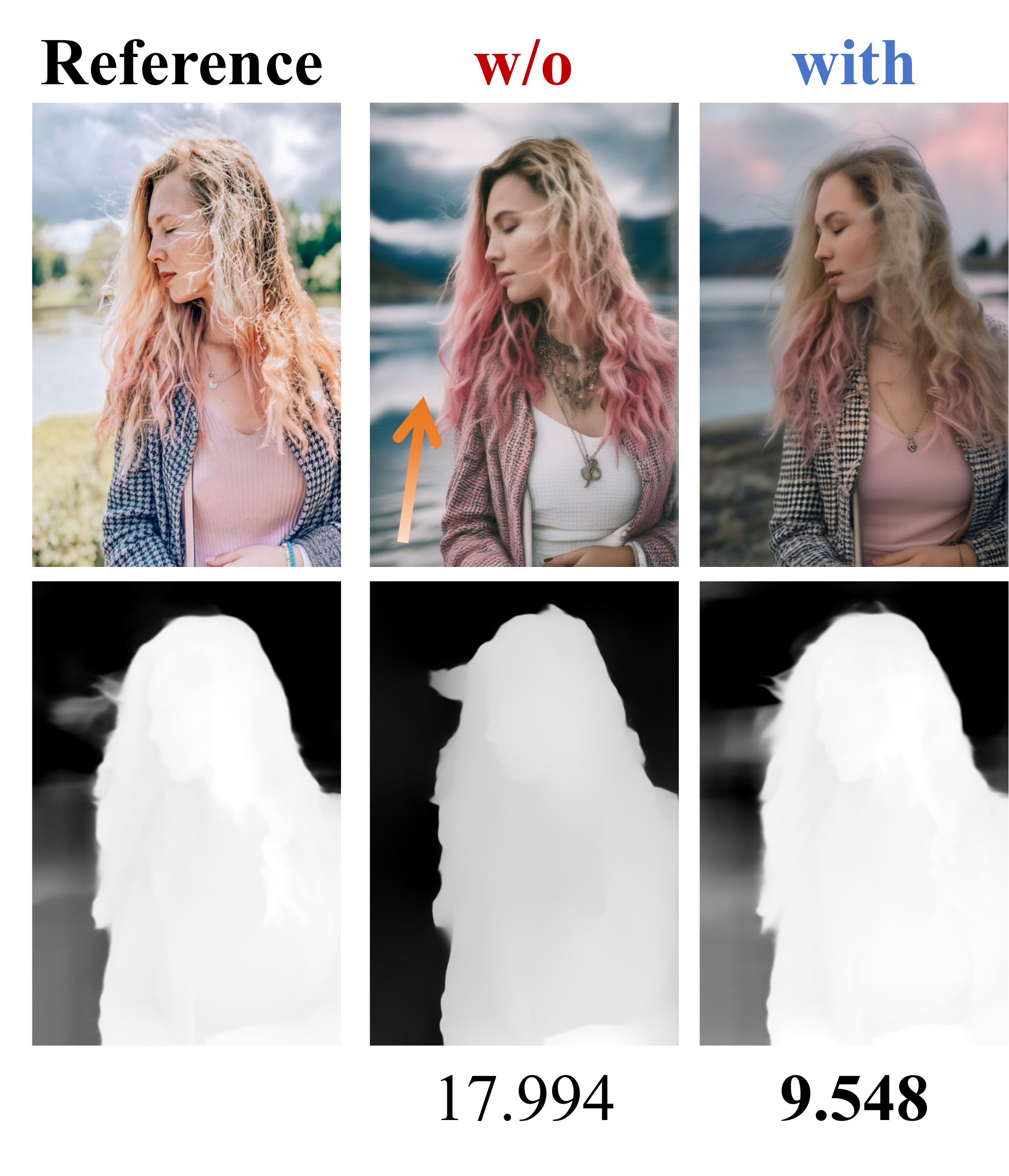} 
    \end{subfigure}
    \hfill
    \begin{subfigure}[b]{0.325\columnwidth}
        \centering
        \includegraphics[width=\columnwidth]{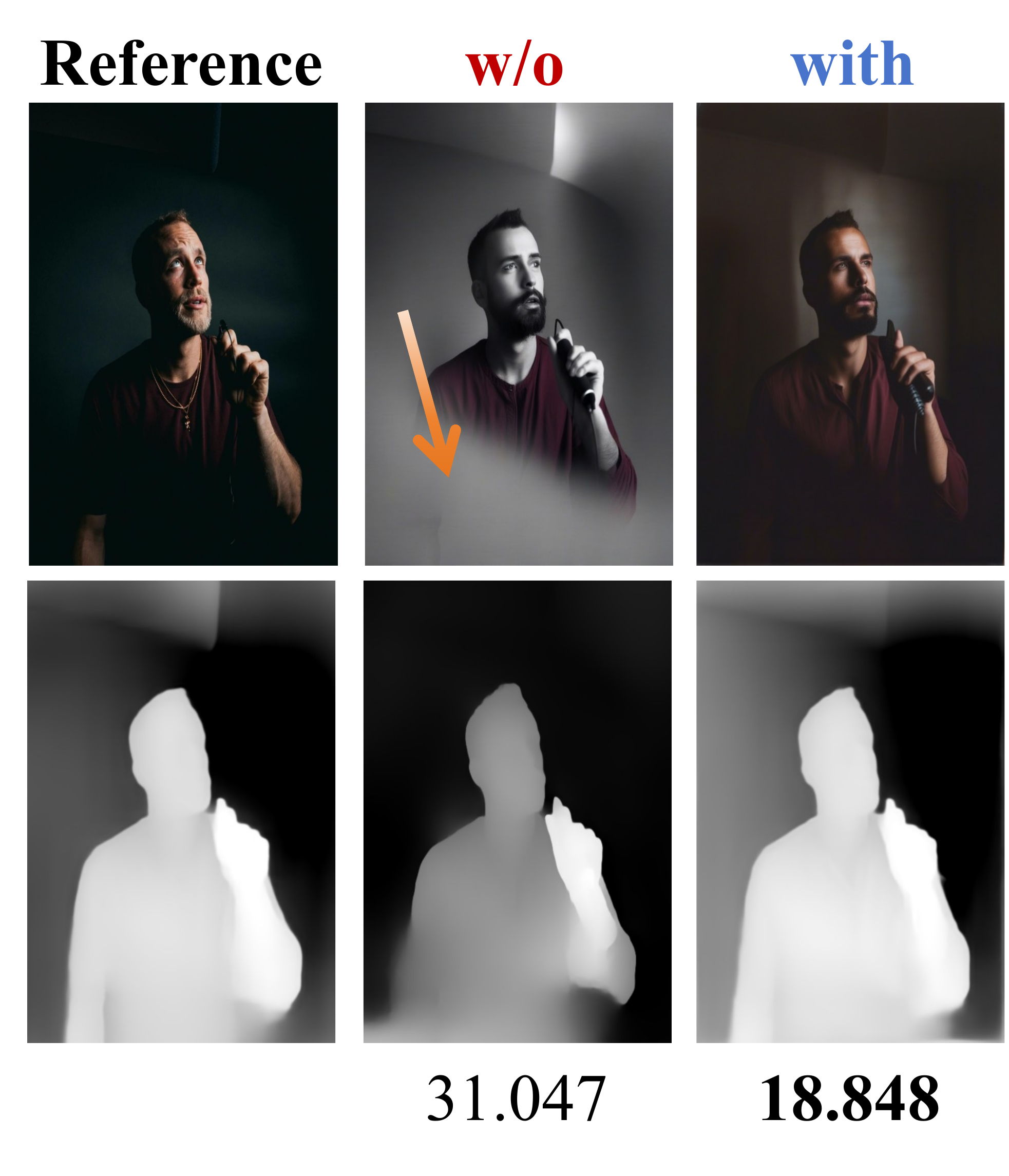} 
    \end{subfigure}
    \hfill
    \begin{subfigure}[b]{0.325\columnwidth}
        \centering
        \includegraphics[width=\columnwidth]{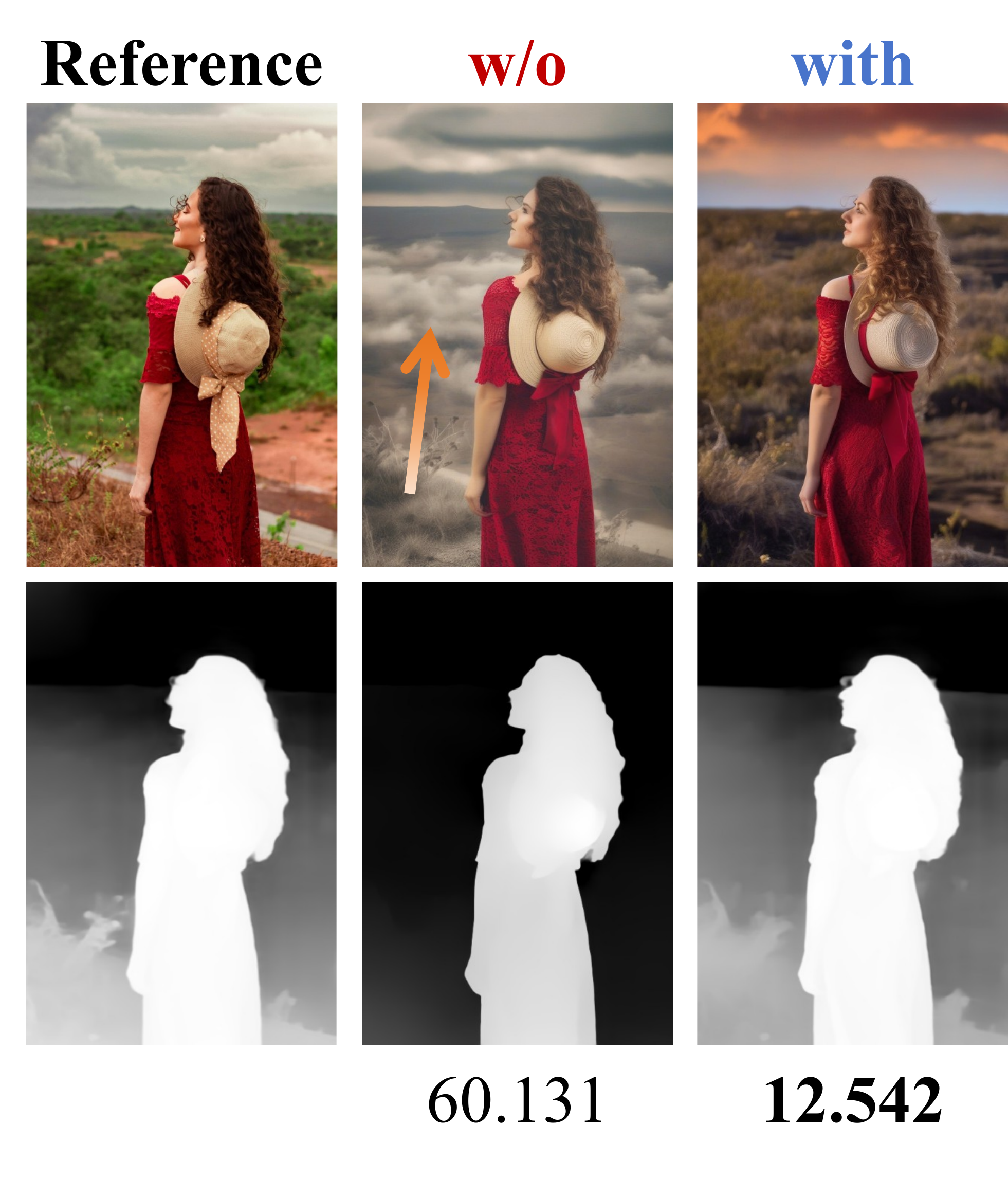} 
    \end{subfigure}
    
    \caption{Qualitative comparisons for depth-controlled generation. The open-loop baseline (\textbf{w/o}) frequently struggles with ambiguous geometric boundaries. In contrast, our closed-loop system (\textbf{with}) ensures precise depth alignment and sharp foreground-background separation (highlighted by arrows/boxes). The Mean Absolute Error (MAE, $\downarrow$) evaluated on the extracted depth maps is reported below.}
    \label{fig:vis_depth}
    \vspace{-10pt}
\end{figure}

\vspace{1mm}
\noindent \textbf{Spatial Control Tasks.} 
Beyond identity preservation, we evaluate the applicability of our closed-loop system on spatial control tasks, specifically pose- and depth-guided generation. As illustrated in Figures \ref{fig:vis_pose} and \ref{fig:vis_depth}, standard open-loop baselines can exhibit structural artifacts under strict geometric constraints, such as misplaced limb joints or blurred depth boundaries. By integrating the iterative feedback mechanism, the system monitors and reduces these spatial deviations. Furthermore, these visual improvements are quantitatively supported by the reductions in spatial error metrics (i.e., MPJPE for pose and MAE for depth) reported below each image pair, indicating the effectiveness of the proposed spatial tracking approach.

\subsection{Quantitative Evaluation}
\label{subsec:quan_eval}

\begin{figure}[t]
    \centering
    \begin{subfigure}[b]{0.32\linewidth}
        \centering
        \includegraphics[width=\textwidth]{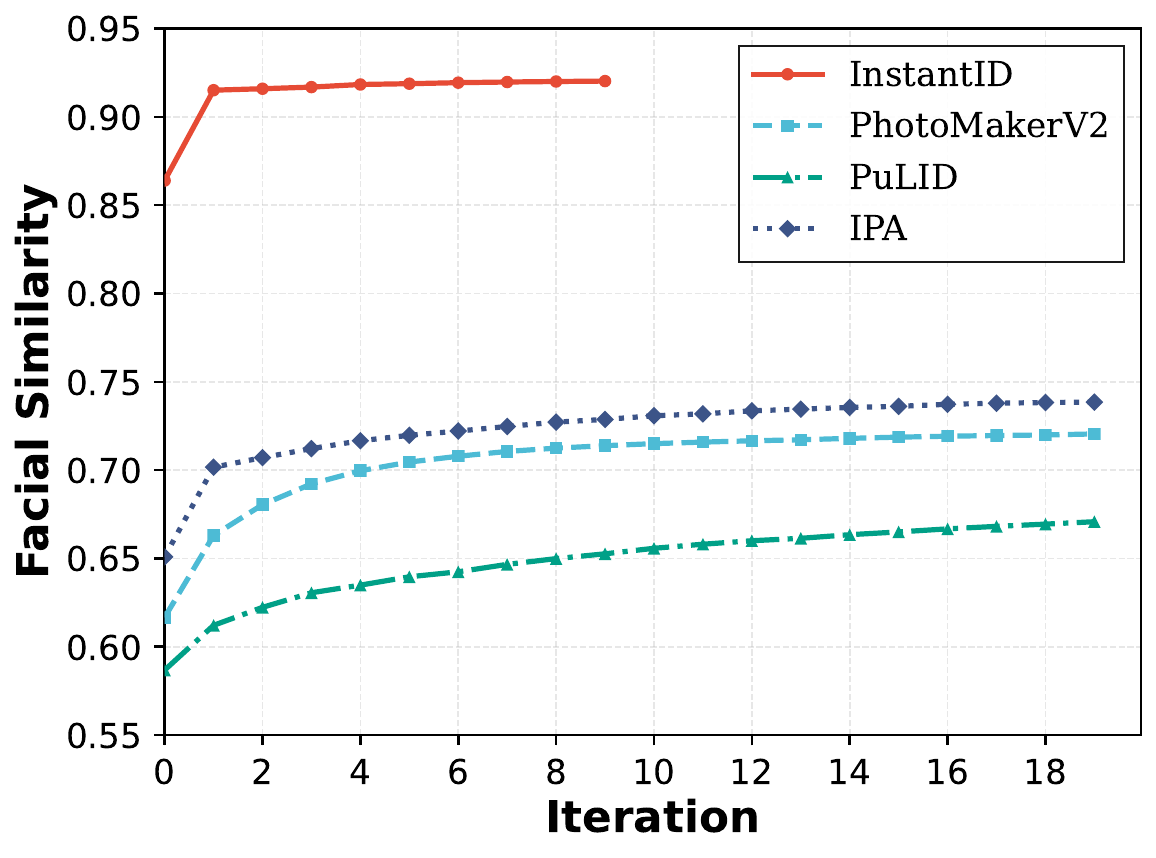}
        \caption{ID-preserving}
        \label{fig:opt_id}
    \end{subfigure}
    \hfill
    \begin{subfigure}[b]{0.32\linewidth}
        \centering
        \includegraphics[width=\textwidth]{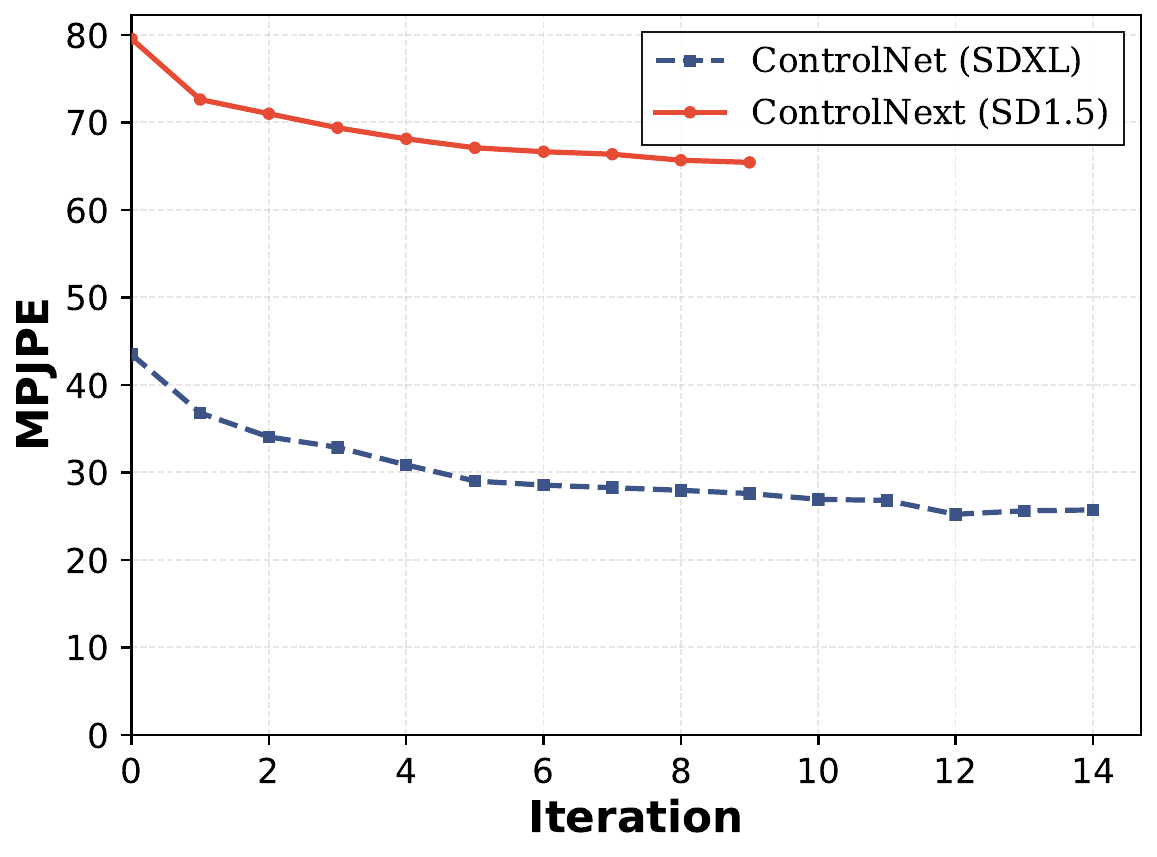}
        \caption{Pose-controlled}
        \label{fig:opt_pose}
    \end{subfigure}
    \hfill
    \begin{subfigure}[b]{0.32\linewidth}
        \centering
        \includegraphics[width=\textwidth]{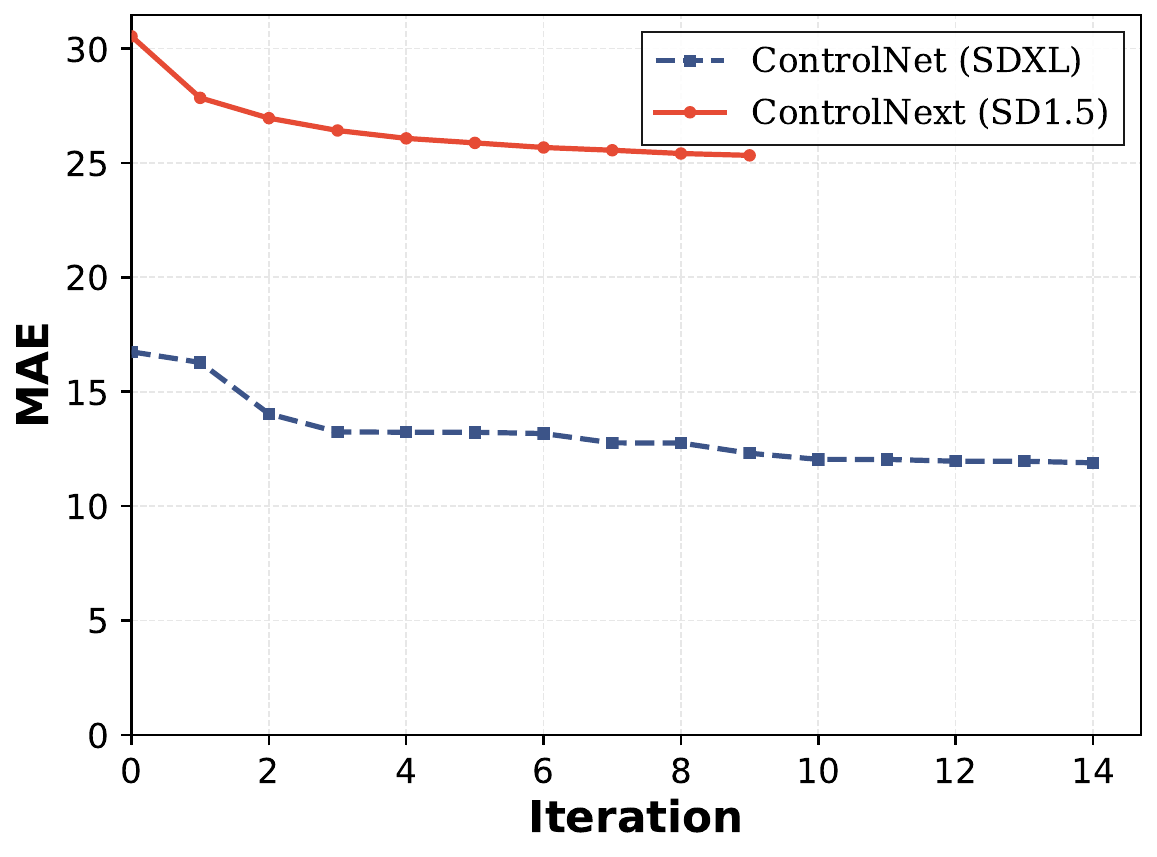}
        \caption{Depth-controlled}
        \label{fig:opt_depth}
    \end{subfigure} 

    \caption{Optimization dynamics across three generation tasks: (a) ID-preserving, (b) pose-controlled, and (c) depth-controlled generation. The plots illustrate the progression of Facial Similarity ($\uparrow$), MPJPE ($\downarrow$), and MAE ($\downarrow$) over iterations, demonstrating stable and monotonic convergence across diverse baselines.}
    \label{fig:optimization_process}
\end{figure}

\textbf{Optimization Dynamics.}
Figure \ref{fig:optimization_process} illustrates the convergence behavior of the proposed closed-loop system. Across all three tasks (identity, pose, and depth), the metrics exhibit a consistent two-phase trajectory: rapid improvement during early iterations (typically 1--5) to rectify discrepancies, followed by a stable fine-tuning phase. The system generally reaches a steady state within 20 iterations for identity-preserving tasks and 15 for spatial control tasks. These trends indicate that the iterative feedback effectively guides generation toward the target constraints without noticeable divergence.

\vspace{1mm}
\noindent \textbf{ID-Preserving Generation.}
Table~\ref{tab:quantity_ID} reports the quantitative results for ID-preserving generation.
Each method group compares the original open-loop baseline, a stronger best-of-20 open-loop variant, and our closed-loop optimization.
Compared with the best-of-20 baselines, our method consistently improves facial similarity across all models and datasets.
Specifically, PuLID obtains the largest gains, with relative improvements of 25.36\% on CelebA300 and 12.98\% on Web100.
PMv2 also benefits substantially, improving by 14.70\% and 14.38\% on the two datasets, respectively.
For stronger identity-preserving baselines, our framework still brings clear gains, improving InstantID by 7.47\% / 5.52\% and IPA by 7.62\% / 6.87\% on CelebA300 / Web100.
Beyond identity fidelity, our method maintains competitive CLIP-I, DINO, Q-Align, and Structure Diversity scores compared with both the original and best-of-20 open-loop baselines.
These results indicate that closed-loop feedback effectively enhances identity consistency without causing a notable degradation in semantic alignment, perceptual quality, or structural diversity.

\vspace{1mm}
\noindent \textbf{Spatial Control Tasks.}
For spatial control tasks, closed-loop optimization consistently improves geometric precision over strong open-loop baselines.
As shown in Table~\ref{tab:quant_spatial}, compared with the best-of-15 variant, applying our method to ControlNet reduces pose error (MPJPE) by 27.71\% and depth error (MAE) by 28.50\%.
The framework also generalizes to ControlNext, achieving error reductions of 13.46\% for pose control and 14.42\% for depth control.
These results demonstrate that iterative feedback correction improves alignment with geometric control signals beyond repeated open-loop sampling.
Meanwhile, DINO, CLIP-I, and Q-Align scores remain comparable in most cases, indicating that the enhanced spatial consistency does not come at the cost of semantic alignment or perceptual quality.

\input{tables/Quant_eval_portrait}

\input{tables/Quant_eval_pose_depth}

\subsection{Ablation Study}
\textbf{Impact of PID Components.}
Table~\ref{tab:ablation} ablates the P, I, and D terms using 50 random samples per task.
The P term alone brings clear improvements over the open-loop baseline on the main alignment metrics.
Adding the I term further improves final alignment, yielding competitive or second-best results across tasks.
The D term mainly accelerates convergence, as the PD variant requires the fewest iterations to reach 95\% of the final gain.
Overall, the full PID controller achieves the best final performance, including the highest identity similarity, the lowest pose MPJPE, the lowest depth MAE, and the best CLIP-I scores for identity and depth.
These results show that P, I, and D play complementary roles in error correction, steady-state refinement, and convergence acceleration.

\input{tables/Ablation_PID}



\vspace{1mm}
\noindent \textbf{Robustness to Prompt Complexity.} 
Table \ref{tab:text_complexity} evaluates generation under four prompt levels: \textit{no prompt}, \textit{simple}, \textit{moderate}, and \textit{complex}. Our closed-loop framework \textcolor{clcolor}{(+CL.)} consistently outperforms the open-loop baseline across all conditions. Notably, even without any prompt, our method achieves the highest identity similarity and significantly reduces spatial errors. This confirms that our iterative feedback effectively overrides textual ambiguities, strictly anchoring the generation to the visual reference.

\noindent \textbf{Robustness to Random Seeds.} 
To assess stability against the inherent stochasticity of diffusion models, we compute performance statistics across multiple random seeds (Table \ref{tab:rand_seed}). Our closed-loop method consistently yields superior mean performance ($\mu$) while drastically reducing standard deviation ($\sigma$). For example, the depth error variance is more than halved. This confirms that dynamic optimization effectively mitigates stochastic fluctuations, ensuring a highly stable generation process.

\input{tables/Ablation_text_seed}

\subsection{Coefficient Sensitivity Analysis}

\begin{figure*}[t]
    \centering
    \begin{subfigure}[t]{0.32\textwidth}
        \centering
        \includegraphics[width=\linewidth]{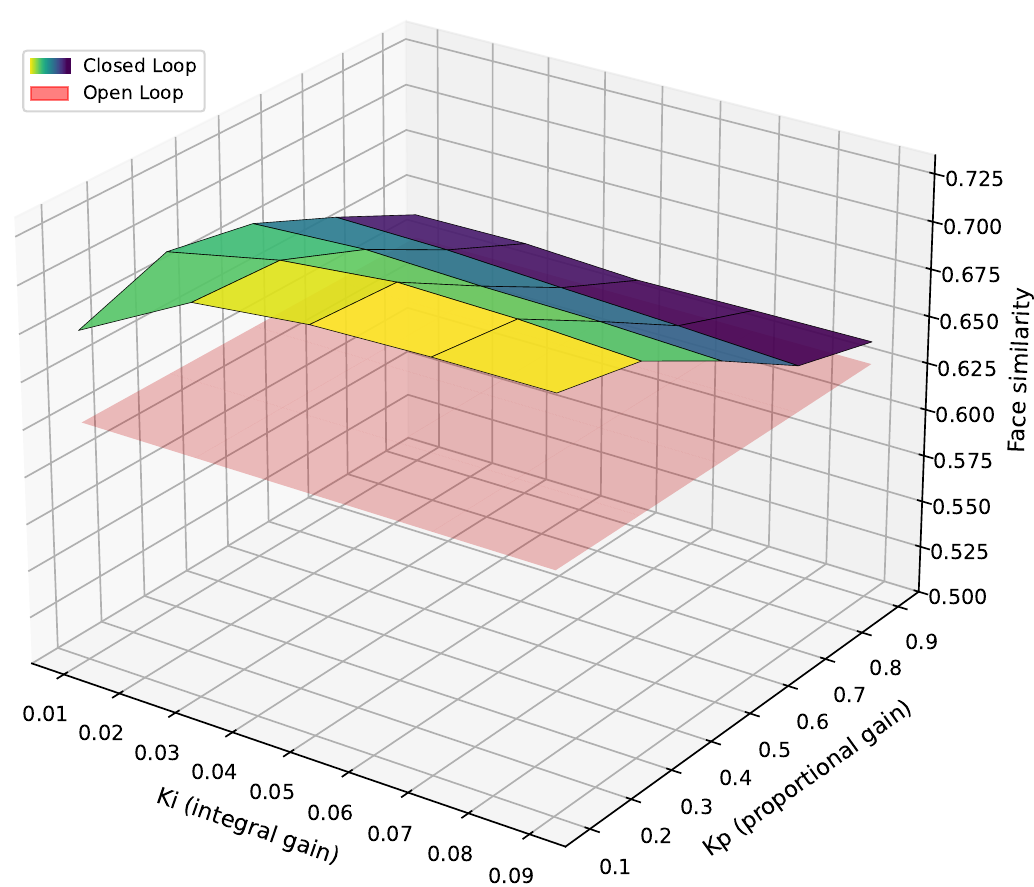}
        \caption{ID preserving}
        \label{fig:coef_sensitivity_id}
    \end{subfigure}
    \hfill
    \begin{subfigure}[t]{0.32\textwidth}
        \centering
        \includegraphics[width=\linewidth]{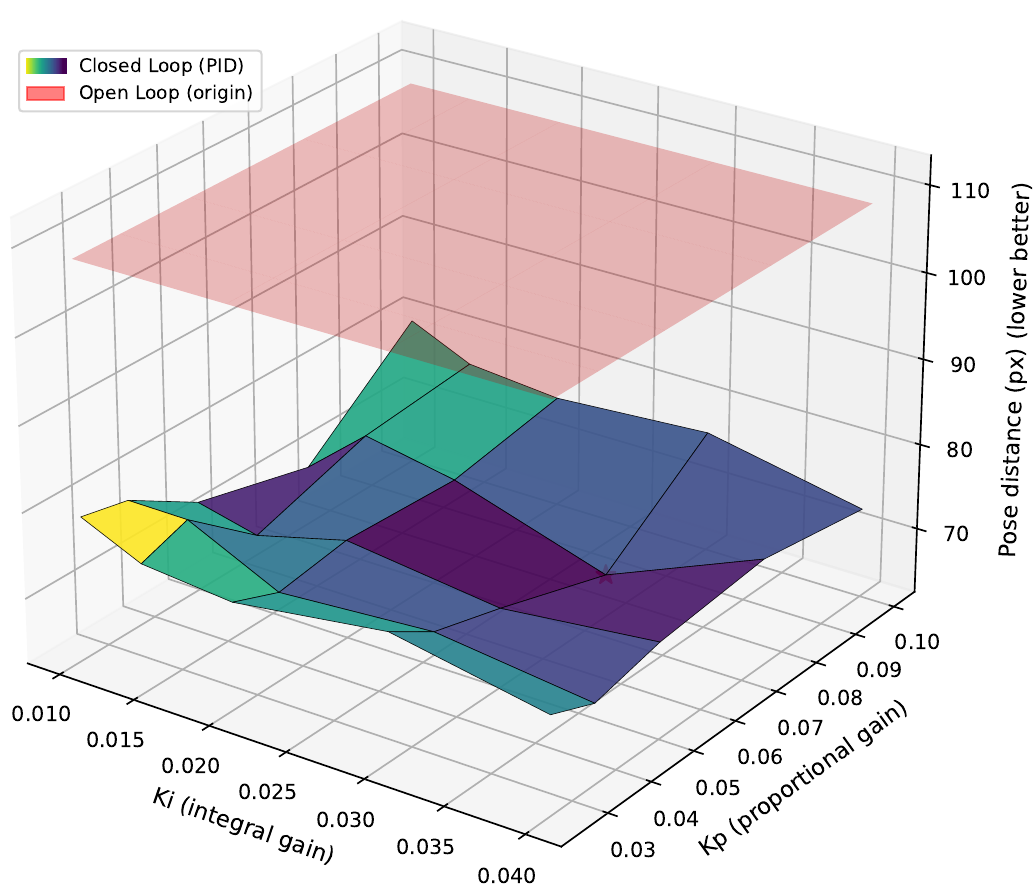}
        \caption{Pose control}
        \label{fig:coef_sensitivity_pose}
    \end{subfigure}
    \hfill
    \begin{subfigure}[t]{0.32\textwidth}
        \centering
        \includegraphics[width=\linewidth]{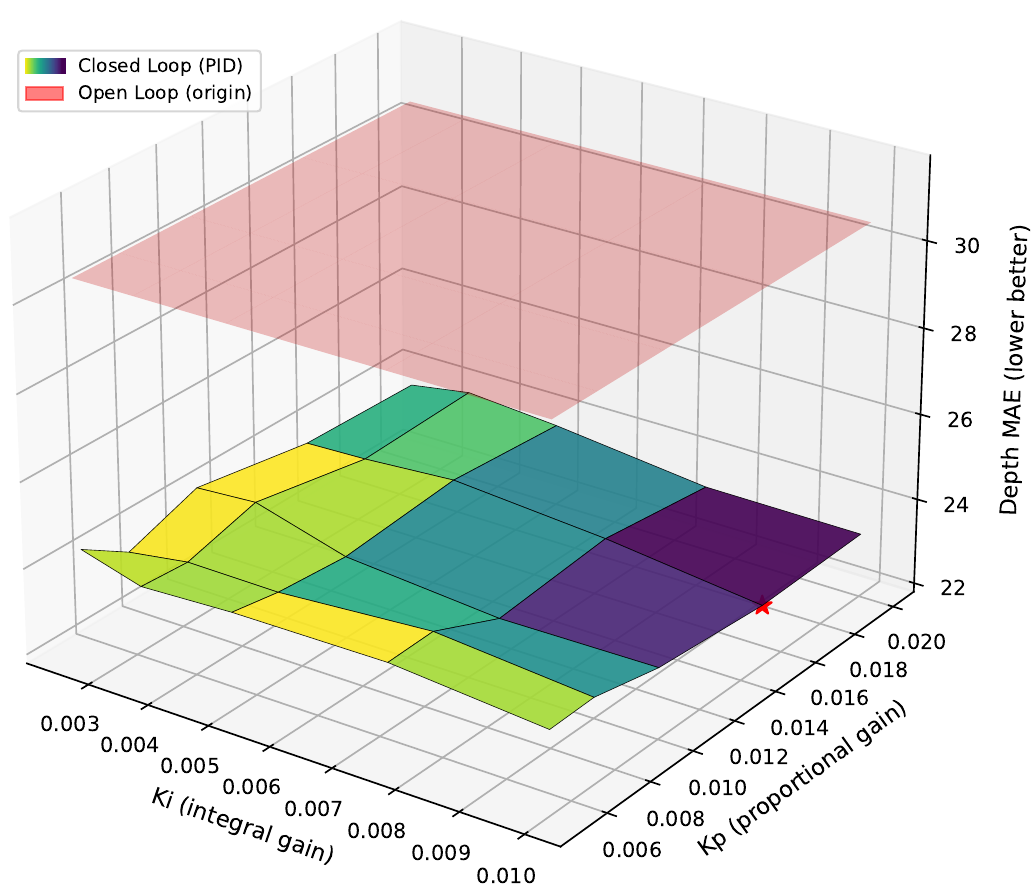}
        \caption{Depth control}
        \label{fig:coef_sensitivity_depth}
    \end{subfigure}

    \caption{Coefficient sensitivity analysis.
    With $K_d$ fixed, we perform a 2D grid search over $K_p$ and $K_i$.
    The closed-loop controller consistently improves ID similarity and reduces pose/depth errors over a broad coefficient range.}
    \label{fig:coef_sensitivity}
    \vspace{-8pt}
\end{figure*}

We further analyze the sensitivity of the PID coefficients by fixing the derivative gain $K_d$ and performing a two-dimensional grid search over $K_p$ and $K_i$.
As shown in Fig.~\ref{fig:coef_sensitivity}, the closed-loop results remain consistently better than the open-loop baseline across a broad coefficient range.
For ID-preserving generation, most coefficient settings lead to higher facial similarity, suggesting that the feedback loop can robustly correct identity drift.
For pose- and depth-controlled generation, the controller achieves lower MPJPE and MAE under most settings, indicating improved geometric alignment with the reference conditions.
Although different tasks favor slightly different coefficient regions, the performance surfaces show stable effective areas rather than sharp isolated optima.
This indicates that the proposed framework does not rely on a narrow hyperparameter choice, but instead exhibits stable effectiveness within a practical range of controller gains.

\subsection{Analysis of Computational Efficiency}
We analyze the computational overhead for ID-preserving and pose-controlled generation, with depth control showing a similar cost profile to pose control.
For ID-preserving generation, each iteration takes 4.2 seconds on average with a peak GPU memory usage of 15,286 MB.
For pose-controlled generation, each iteration takes 11.6 seconds with a peak memory usage of 14,148 MB.
Since the per-iteration cost is comparable to standard single-round inference, the total latency scales approximately linearly with the maximum number of iterations.

%% file: tables/Quant_eval_portrait.tex
\newcommand{\up}[1]{\textsuperscript{\scalebox{0.7}{\textcolor{clcolor}{+#1\%}}}}

\begin{table}[t]
\centering
\caption{Quantitative comparison on ID-preserving generation.
Each method group includes the original open-loop baseline, its best-of-20 variant, and our closed-loop optimization \textcolor{clcolor}{(+CL.)}.
\textcolor{blue}{\textbf{Bold}} indicates the best result within each method.
The superscript gain reports the relative improvement of +CL. over the best-of-20 baseline in Facial Similarity.
[Sim.: Facial Similarity, Q-Al.: Q-Align, Div.: Structure Diversity.]}
\label{tab:quantity_ID}

\scriptsize
\setlength{\tabcolsep}{2pt}
\renewcommand{\arraystretch}{1.25} 

\resizebox{\textwidth}{!}{
\begin{tabular}{l | ccccc | ccccc}
\toprule
\multirow{2}{*}{\textbf{Method}} & \multicolumn{5}{c|}{\textbf{CelebA300}} & \multicolumn{5}{c}{\textbf{Web100}} \\ 
\cmidrule(lr){2-6} \cmidrule(lr){7-11}
 & Sim.$\uparrow$ & CLIP-I$\uparrow$ & DINO$\uparrow$ & Q-Al.$\uparrow$ & Div.$\uparrow$ & Sim.$\uparrow$ & CLIP-I$\uparrow$ & DINO$\uparrow$ & Q-Al.$\uparrow$ & Div.$\uparrow$ \\ 
\midrule

PuLID \cite{pulid2024} & 0.550 & 0.733 & 0.282 & \textcolor{blue}{\textbf{4.980}} & \textcolor{blue}{\textbf{0.059}} & 0.584 & 0.779 & 0.341 & \textcolor{blue}{\textbf{4.971}} & \textcolor{blue}{\textbf{0.107}} \\
PuLID-best20 \cite{pulid2024} & 0.560 & 0.737 & 0.285 & 4.972 & 0.054 & 0.593 & 0.785 & 0.342 & 4.964 & 0.101 \\
\rowcolor{blue!10} PuLID \textcolor{clcolor}{+CL.} & \textcolor{blue}{\textbf{0.702}}\up{25.36} & \textcolor{blue}{\textbf{0.743}} & \textcolor{blue}{\textbf{0.287}} & 4.975 & 0.047 & \textcolor{blue}{\textbf{0.670}}\up{12.98} & \textcolor{blue}{\textbf{0.794}} & \textcolor{blue}{\textbf{0.344}} & 4.966 & 0.105 \\
\midrule

PhotoMaker \cite{photomaker2024} & 0.548 & 0.768 & 0.329 & \textcolor{blue}{\textbf{4.974}} & 0.0738 & 0.589 & 0.823 & \textcolor{blue}{\textbf{0.412}} & 4.928 & \textcolor{blue}{\textbf{0.131}} \\
PMv2-best20 \cite{photomaker2024} & 0.551 & 0.770 & \textcolor{blue}{\textbf{0.332}} & 4.969 & 0.070 & 0.598 & 0.824 & 0.410 & 4.933 & 0.128 \\
\rowcolor{blue!10} PMv2 \textcolor{clcolor}{+CL.} & \textcolor{blue}{\textbf{0.632}}\up{14.70} & \textcolor{blue}{\textbf{0.771}} & 0.330 & 4.963 & \textcolor{blue}{\textbf{0.0741}} & \textcolor{blue}{\textbf{0.684}}\up{14.38} & \textcolor{blue}{\textbf{0.827}} & 0.407 & \textcolor{blue}{\textbf{4.965}} & 0.130 \\
\midrule

InstantID \cite{instantid2024} & 0.812 & 0.774 & 0.604 & \textcolor{blue}{\textbf{4.314}} & \textcolor{blue}{\textbf{0.023}} & 0.718 & 0.814 & \textcolor{blue}{\textbf{0.493}} & 4.035 & \textcolor{blue}{\textbf{0.101}} \\
InstantID-best20 \cite{instantid2024} & 0.817 & 0.780 & \textcolor{blue}{\textbf{0.605}} & 4.305 & 0.017 & 0.724 & 0.812 & 0.489 & 4.037 & 0.094 \\ 
\rowcolor{blue!10} InstantID \textcolor{clcolor}{+CL.} & \textcolor{blue}{\textbf{0.878}}\up{7.47} & \textcolor{blue}{\textbf{0.786}} & 0.601 & 4.313 & 0.022 & \textcolor{blue}{\textbf{0.764}}\up{5.52} & \textcolor{blue}{\textbf{0.818}} & 0.491 & \textcolor{blue}{\textbf{4.433}} & 0.100 \\
\midrule

IPA & 0.708 & 0.756 & 0.529 & 4.946 & \textcolor{blue}{\textbf{0.114}} & 0.672 & 0.798 & 0.492 & 4.973 & 0.149 \\
IPA-best20 & 0.722 & 0.757 & 0.530 & 4.941 & 0.106 & 0.684 & 0.803 & \textcolor{blue}{\textbf{0.494}} & 4.963 & 0.142 \\
\rowcolor{blue!10} IPA \textcolor{clcolor}{+CL.} & \textcolor{blue}{\textbf{0.777}}\up{7.62} & \textcolor{blue}{\textbf{0.774}} & \textcolor{blue}{\textbf{0.532}} & \textcolor{blue}{\textbf{4.950}} & 0.108 & \textcolor{blue}{\textbf{0.731}}\up{6.87} & \textcolor{blue}{\textbf{0.808}} & 0.486 & \textcolor{blue}{\textbf{4.977}} & \textcolor{blue}{\textbf{0.151}} \\ 

\bottomrule
\end{tabular}
}
\vspace{-10pt}
\end{table}

%% file: tables/Quant_eval_pose_depth.tex
\newcommand{\down}[1]{\textsuperscript{\scalebox{0.7}{\textcolor{clcolor}{-#1\%}}}}

\begin{table}[t]
\centering
\caption{Quantitative comparison on pose- and depth-controlled generation.
Each method group includes the original open-loop baseline, its best-of-15 variant, and our closed-loop optimization \textcolor{clcolor}{(+CL.)}.
\textcolor{blue}{\textbf{Bold}} indicates the best result within each method group.
The superscript reduction reports the relative error decrease of +CL. over the best-of-15 baseline for MPJPE and MAE.
[Q-Al.: Q-Align.]}
\label{tab:quant_spatial}

\scriptsize
\setlength{\tabcolsep}{3pt}
\renewcommand{\arraystretch}{1.25} 

\resizebox{\textwidth}{!}{
\begin{tabular}{l | cccc | cccc}
\toprule
\multirow{2}{*}{\textbf{Method}} & \multicolumn{4}{c|}{\textbf{Pose-Controlled Generation}} & \multicolumn{4}{c}{\textbf{Depth-Controlled Generation}} \\ 
\cmidrule(lr){2-5} \cmidrule(lr){6-9}
 & MPJPE$\downarrow$ & DINO$\uparrow$ & CLIP-I$\uparrow$ & Q-Al.$\uparrow$ & MAE$\downarrow$ & DINO$\uparrow$ & CLIP-I$\uparrow$ & Q-Al.$\uparrow$ \\ 
\midrule

ControlNet \cite{ControlNet2023} 
& 39.193 & 0.6297 & \textcolor{blue}{\textbf{0.7891}} & 4.936 
& 16.983 & 0.781 & 0.840 & 4.956 \\

ControlNet-best15 
& 36.521 & 0.626 & 0.788 & 4.928 
& 15.797 & 0.775 & 0.838 & 4.954 \\

\rowcolor{blue!10} ControlNet \textcolor{clcolor}{+CL.} 
& \textcolor{blue}{\textbf{26.401}}\down{27.71} 
& \textcolor{blue}{\textbf{0.6300}} 
& 0.7887 
& \textcolor{blue}{\textbf{4.940}} 
& \textcolor{blue}{\textbf{11.295}}\down{28.50} 
& \textcolor{blue}{\textbf{0.785}} 
& \textcolor{blue}{\textbf{0.843}} 
& \textcolor{blue}{\textbf{4.963}} \\ 
\midrule  

ControlNext \cite{controlnext2024} 
& 79.542 & 0.572 & \textcolor{blue}{\textbf{0.742}} & 4.951 
& 30.544 & \textcolor{blue}{\textbf{0.549}} & 0.733 & \textcolor{blue}{\textbf{4.930}} \\

ControlNext-best15 
& 75.827 & 0.569 & 0.738 & 4.945 
& 29.598 & 0.543 & 0.733 & 4.928 \\

\rowcolor{blue!10} ControlNext \textcolor{clcolor}{+CL.} 
& \textcolor{blue}{\textbf{65.623}}\down{13.46} 
& \textcolor{blue}{\textbf{0.575}} 
& 0.739 
& \textcolor{blue}{\textbf{4.952}} 
& \textcolor{blue}{\textbf{25.331}}\down{14.42} 
& 0.548 
& \textcolor{blue}{\textbf{0.734}} 
& 4.924 \\ 

\bottomrule
\end{tabular}
}
\end{table}

%% file: tables/Ablation_PID.tex
\begin{table}[t]
\centering
\caption{Ablation study on different PID component combinations. 
\textcolor{blue}{\textbf{Blue bold}} and \textbf{bold} indicate the best and second-best results, respectively. 
95\%Gain denotes the average number of iterations required to reach 95\% of the final performance gain.}
\label{tab:ablation}

\scriptsize
\setlength{\tabcolsep}{3.5pt}
\renewcommand{\arraystretch}{1.25} 

\resizebox{\textwidth}{!}{
\begin{tabular}{l ccc ccc ccc ccc}
\toprule
\multirow{2}{*}{\textbf{Method}} & \multirow{2}{*}{\textbf{P}} & \multirow{2}{*}{\textbf{I}} & \multirow{2}{*}{\textbf{D}} & \multicolumn{3}{c}{\textbf{Identity}} & \multicolumn{3}{c}{\textbf{Pose}} & \multicolumn{3}{c}{\textbf{Depth}} \\ 
\cmidrule(lr){5-7} \cmidrule(lr){8-10} \cmidrule(lr){11-13}
 & & & & Sim.$\uparrow$ & CLIP-I$\uparrow$ & 95\%Gain$\downarrow$ & MPJPE$\downarrow$ & CLIP-I$\uparrow$ & 95\%Gain$\downarrow$ & MAE$\downarrow$ & CLIP-I$\uparrow$ & 95\%Gain$\downarrow$ \\ 
\midrule

Open loop  
& \textcolor{red}{$\times$} 
& \textcolor{red}{$\times$} 
& \textcolor{red}{$\times$} 
& 0.629 
& 0.739 
& - 
& 108.867
& 0.762 
& - 
& 30.612 
& 0.816 
& - \\ 
\midrule

\multirow{4}{*}{\shortstack{Closed loop\\(ours)}} 
& \textcolor{green!60!black}{$\checkmark$} 
& \textcolor{red}{$\times$} 
& \textcolor{red}{$\times$} 
& 0.694 
& 0.763 
& \textbf{5.58} 
& 78.903 
& 0.763
& \textbf{3.66} 
& 24.222 
& 0.818 
& \textbf{3.98} \\

& \textcolor{green!60!black}{$\checkmark$} 
& \textcolor{green!60!black}{$\checkmark$} 
& \textcolor{red}{$\times$} 
& \textbf{0.723} 
& \textbf{0.7736} 
& 6.80 
& \textbf{67.400}
& \textcolor{blue}{\textbf{0.771}} 
& 4.70 
& \textbf{23.345} 
& 0.8189 
& 4.44 \\

& \textcolor{green!60!black}{$\checkmark$} 
& \textcolor{red}{$\times$} 
& \textcolor{green!60!black}{$\checkmark$} 
& 0.695 
& 0.765
& \textcolor{blue}{\textbf{5.52}} 
& 76.514 
& 0.766
& \textcolor{blue}{\textbf{3.42}} 
& 23.645
& \textbf{0.8194}
& \textcolor{blue}{\textbf{3.88}} \\

& \textcolor{green!60!black}{$\checkmark$} 
& \textcolor{green!60!black}{$\checkmark$} 
& \textcolor{green!60!black}{$\checkmark$} 
& \textcolor{blue}{\textbf{0.724}}
& \textcolor{blue}{\textbf{0.7738}}
& 6.58
& \textcolor{blue}{\textbf{66.231}} 
& \textbf{0.769}
& 4.62
& \textcolor{blue}{\textbf{23.213}} 
& \textcolor{blue}{\textbf{0.821}}
& 4.20 \\ 
\bottomrule
\end{tabular}
}
\vspace{-10pt}
\end{table}

%% file: tables/Ablation_text_seed.tex
\begin{table}[htbp]
    \centering
    \renewcommand{\arraystretch}{1.25} 
    
    \begin{minipage}[t]{0.52\textwidth}
        \centering
        \caption{Ablation on text prompts. \textcolor{blue}{\textbf{Bold}} indicates closed-loop improvements.}
        \label{tab:text_complexity}
        
        \resizebox{\linewidth}{!}{
        \begin{tabular}{l cc cc cc}
            \toprule
            \multirow{2}{*}{\textbf{Prompt Level}} & \multicolumn{2}{c}{\textbf{Identity}} & \multicolumn{2}{c}{\textbf{Pose}} & \multicolumn{2}{c}{\textbf{Depth}} \\ 
            \cmidrule(lr){2-3} \cmidrule(lr){4-5} \cmidrule(lr){6-7}
             & Base & \cellcolor{blue!10}\textcolor{clcolor}{+CL.} & Base & \cellcolor{blue!10}\textcolor{clcolor}{+CL.} & Base & \cellcolor{blue!10}\textcolor{clcolor}{+CL.} \\ 
            \midrule
            
            No prompt & 0.651 & \cellcolor{blue!10}\textcolor{blue}{\textbf{0.707}} & 29.952 & \cellcolor{blue!10}\textcolor{blue}{\textbf{22.066}} & 22.598 & \cellcolor{blue!10}\textcolor{blue}{\textbf{14.553}} \\
            Simple    & 0.592 & \cellcolor{blue!10}\textcolor{blue}{\textbf{0.667}} & 26.240 & \cellcolor{blue!10}\textcolor{blue}{\textbf{21.810}} & 21.638 & \cellcolor{blue!10}\textcolor{blue}{\textbf{12.610}} \\
            Moderate  & 0.596 & \cellcolor{blue!10}\textcolor{blue}{\textbf{0.659}} & 27.064 & \cellcolor{blue!10}\textcolor{blue}{\textbf{22.491}} & 22.970 & \cellcolor{blue!10}\textcolor{blue}{\textbf{13.269}} \\
            Complex   & 0.593 & \cellcolor{blue!10}\textcolor{blue}{\textbf{0.666}} & 26.677 & \cellcolor{blue!10}\textcolor{blue}{\textbf{21.324}} & 21.902 & \cellcolor{blue!10}\textcolor{blue}{\textbf{12.576}} \\ 
            
            \bottomrule
        \end{tabular}
        }
    \end{minipage}
    \hfill 
    \begin{minipage}[t]{0.46\textwidth}
        \centering
        \caption{Stability across seeds. [$\mu$: Mean, $\sigma$: Std.]}
        \label{tab:rand_seed}
        
        \resizebox{\linewidth}{!}{
        \begin{tabular}{l cc cc cc}
            \toprule
            \multirow{2}{*}{\textbf{Method}} & \multicolumn{2}{c}{\textbf{Identity}} & \multicolumn{2}{c}{\textbf{Pose}} & \multicolumn{2}{c}{\textbf{Depth}} \\ 
            \cmidrule(lr){2-3} \cmidrule(lr){4-5} \cmidrule(lr){6-7} 
             & $\mu$ & $\sigma$ & $\mu$ & $\sigma$ & $\mu$ & $\sigma$ \\ 
            \midrule
            
            Base & 0.658 & 0.021 & 23.618 & 9.050 & 20.808 & 5.348 \\
            \rowcolor{blue!10} \textcolor{clcolor}{+CL.} (ours) & \textcolor{blue}{\textbf{0.706}} & \textcolor{blue}{\textbf{0.009}} & \textcolor{blue}{\textbf{18.220}} & \textcolor{blue}{\textbf{5.629}} & \textcolor{blue}{\textbf{10.942}} & \textcolor{blue}{\textbf{1.975}} \\
            
            \bottomrule
        \end{tabular}
        }
    \end{minipage}
\end{table}

%% file: 6_Conclusion.tex
\section{Conclusion}
\label{sec:conclusion}

In this work, we propose a test-time iterative optimization framework that brings automatic control principles into controllable image generation.
By introducing a closed-loop feedback mechanism during inference, our method uses a modified PID controller to iteratively update latent control signals and reduce reference deviations.
This simple yet effective feedback design improves reference consistency in both identity fidelity and spatial structure, while requiring no model retraining.
Extensive experiments across diverse tasks and backbone models demonstrate the effectiveness and plug-and-play versatility of the proposed framework.
Beyond a specific optimization method, this work offers a new perspective: controllable generation can be viewed as a dynamic control problem, where classical control theory provides interpretable tools for improving generative models.

\vspace{1mm}
\noindent\textbf{Limitations and Future Work:}

\vspace{1mm}
\noindent -\textbf{Manual Parameter Tuning:}
The current PID controller uses manually selected hyperparameters.
Future work will explore adaptive control strategies to automate coefficient tuning.

\vspace{1mm}
\noindent -\textbf{Inference Latency:}
The iterative nature of closed-loop optimization increases inference latency and computational cost.
Future research will focus on accelerating convergence and reducing the required number of iterations.

\vspace{1mm}
\noindent -\textbf{Dependence on Sensor Accuracy:}
The feedback loop relies on external sensors, such as identity encoders and pose/depth estimators.
Future work will investigate robust control strategies to mitigate sensor noise and improve generation quality under imperfect feedback.


\section*{Acknowledgment}
The work was supported in part by the National Natural Science Foundation of China under Grant 62301310, 62572317 and 62225112.

%% file: X_Appendix_in_main.tex
\title{Supplementary Material\\
From Open Loop to Closed Loop: A Test-Time Iterative Optimization Framework for Reference-Consistent Image Generation}

\titlerunning{Supplementary Material}

\author{Baixuan Zhao\inst{1}\orcidlink{0000-0001-8929-8322} \and
Xinyu Zhang\inst{3}\orcidlink{0009-0007-3608-1673} \and
Huayu Zheng\inst{1}\orcidlink{0009-0002-4964-6761} \and
Shuaicheng Liu\inst{3}\orcidlink{0000-0002-8815-5335} \and \\
Xiongkuo Min\inst{1}\orcidlink{0000-0001-5693-0416} \and
Guangtao Zhai\inst{1}\orcidlink{0000-0001-8165-9322} \and
Xiaohong Liu\inst{1,2}\thanks{Corresponding author. Email: xiaohongliu@sjtu.edu.cn}\orcidlink{0000-0001-6377-4730}
}

\authorrunning{B.~Zhao et al.}

\institute{
Shanghai Jiao Tong University, Shanghai, China
\and
Shanghai Innovation Institute, Shanghai, China
\and
University of Electronic Science and Technology of China, Chengdu, Sichuan, China
}

\maketitle

\setcounter{footnote}{0}
\setcounter{figure}{0}
\renewcommand{\thefigure}{S\arabic{figure}}

\setcounter{table}{0}
\renewcommand{\thetable}{S\arabic{table}}

\setcounter{equation}{0}
\renewcommand{\theequation}{S\arabic{equation}}

\setcounter{algorithm}{0}
\renewcommand{\thealgorithm}{S\arabic{algorithm}}

\setcounter{section}{0}
\renewcommand{\thesection}{S\arabic{section}}
\renewcommand{\thesubsection}{S\arabic{section}.\arabic{subsection}}
\renewcommand{\thesubsubsection}{S\arabic{section}.\arabic{subsection}.\arabic{subsubsection}}


\section{Additional Implementation Details}

\subsection{PID Coefficients Tuning Techniques}
The tuning of PID coefficients relies on manual adjustment. We summarize the following tuning techniques.

\begin{itemize}
    \item \textbf{$K_p$ tuning}:
    \begin{itemize}
        \item Start with $K_p=0$ and gradually increase it until the system exhibits oscillations.
        \item Reduce $K_p$ to 50\% of the oscillatory value, or adjust it within $0.4$--$0.8$ of the critical gain for stability.
    \end{itemize}

    \item \textbf{$K_i$ tuning}:
    \begin{itemize}
        \item With $K_p$ fixed, start from $K_i=0$ and gradually increase $K_i$ until the steady-state error is reduced.
        \item Anti-windup measures can be adopted:
        \begin{itemize}
            \item Integral clamping: limit the integral term to prevent excessive accumulation.
            \item Integral separation: disable integration when the error exceeds a threshold.
        \end{itemize}
    \end{itemize}

    \item \textbf{$K_d$ tuning}:
    Gradually increase $K_d$ to suppress overshoot and improve response stability.
\end{itemize}

\subsection{Pseudocode}
This algorithm improves the reference consistency of generated images with low code modification cost.
Only a few core lines need to be added to the original inference logic of a basic generative model, as shown in Algorithm~\ref{algo:close_loop}.
The added code mainly focuses on error calculation and PID adjustment.

\textbf{Initialization.}
The target state $u^*$ is extracted from the reference image via the sensor $\mathcal{F}$, while the previous error $e_{\mathrm{prev}}$ and integral term $\Sigma_e$ are initialized.
No model reconstruction is required.

\textbf{Iterative optimization.}
After the model outputs image $I_k$, the tracking error $e_k$ is computed through $\mathcal{F}$.
The control input $u_{k+1}$ is then updated using the modified PID rule, and the historical error $e_{\mathrm{prev}}$ is refreshed.

\textbf{Result output.}
The original image output logic is kept unchanged to obtain the final generated image.

Overall, no large-scale modification to the generative model $\mathcal{G}$ or sensor $\mathcal{F}$ is required.
A small amount of additional code for closed-loop feedback and PID adjustment is sufficient to improve reference consistency.

\begin{algorithm}[H]
   \caption{Closed-loop optimization for reference-consistent image generation}
   \label{algo:close_loop}
\begin{algorithmic}[1]
   \STATE {\bfseries Input:} Reference image $I^*$, text prompt $c$, PID coefficients $(K_p, K_i, K_d)$, generative model $\mathcal{G}$, sensor $\mathcal{F}$, maximum iteration $N$.
   \STATE {\bfseries Output:} Generated image $I_N$.
   
   \STATE \textit{// Initialization}
   \STATE $e_{\mathrm{prev}} \leftarrow \mathbf{0}$, $\Sigma_e \leftarrow \mathbf{0}$, $k \leftarrow 1$
   \STATE $u^* \leftarrow \mathcal{F}(I^*)$
   \STATE $u_1 \leftarrow u^*$
   
   \STATE \textit{// Iterative optimization loop}
   \WHILE{$k \le N$}
      \STATE $I_k \leftarrow \mathcal{G}(u_k, c)$ \hfill \textit{// Generate image}
      \STATE $e_k \leftarrow u^* - \mathcal{F}(I_k)$ \hfill \textit{// Compute error}
      \STATE $\Sigma_e \leftarrow \Sigma_e + e_k$ \hfill \textit{// Update integral}
      
      \STATE \textit{// Modified PID control update}
      \STATE $u_{k+1} \leftarrow u^* + K_p e_k + K_i \Sigma_e + K_d (e_k - e_{\mathrm{prev}})$
      
      \STATE $e_{\mathrm{prev}} \leftarrow e_k$ \hfill \textit{// Save error}
      \STATE $k \leftarrow k+1$
   \ENDWHILE
   \STATE \textbf{return} $I_N$
\end{algorithmic}
\end{algorithm}

\section{More Qualitative Comparisons}

\subsection{Visualization of the Optimization Trajectory}
\begin{figure}[htbp]
    \centering
    \begin{subfigure}[b]{0.32\textwidth}
        \centering
        \includegraphics[width=\linewidth]{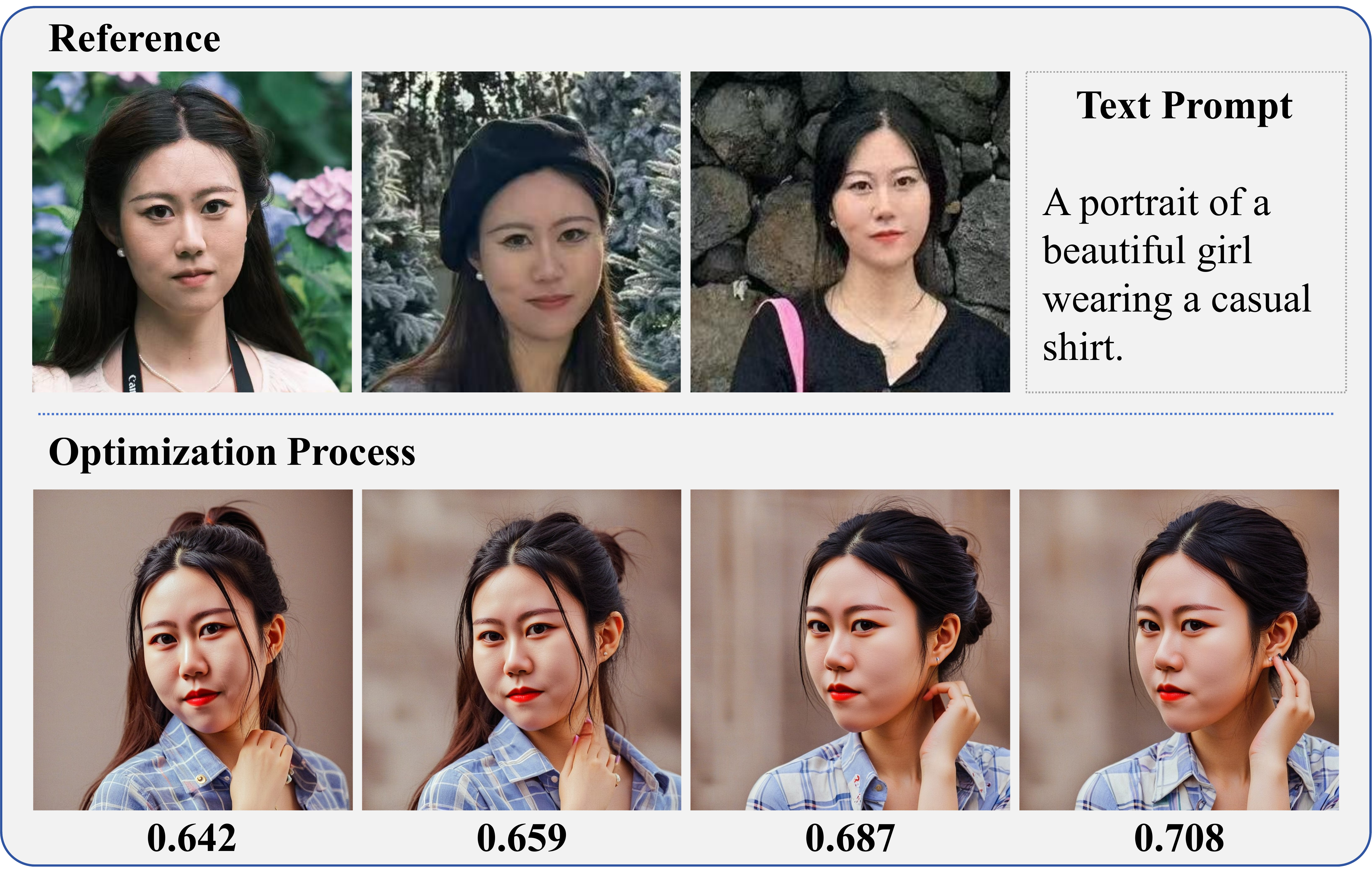}
        \caption{Identity}
        \label{fig:optimization_process_id}
    \end{subfigure}
    \hfill
    \begin{subfigure}[b]{0.32\textwidth}
        \centering
        \includegraphics[width=\linewidth]{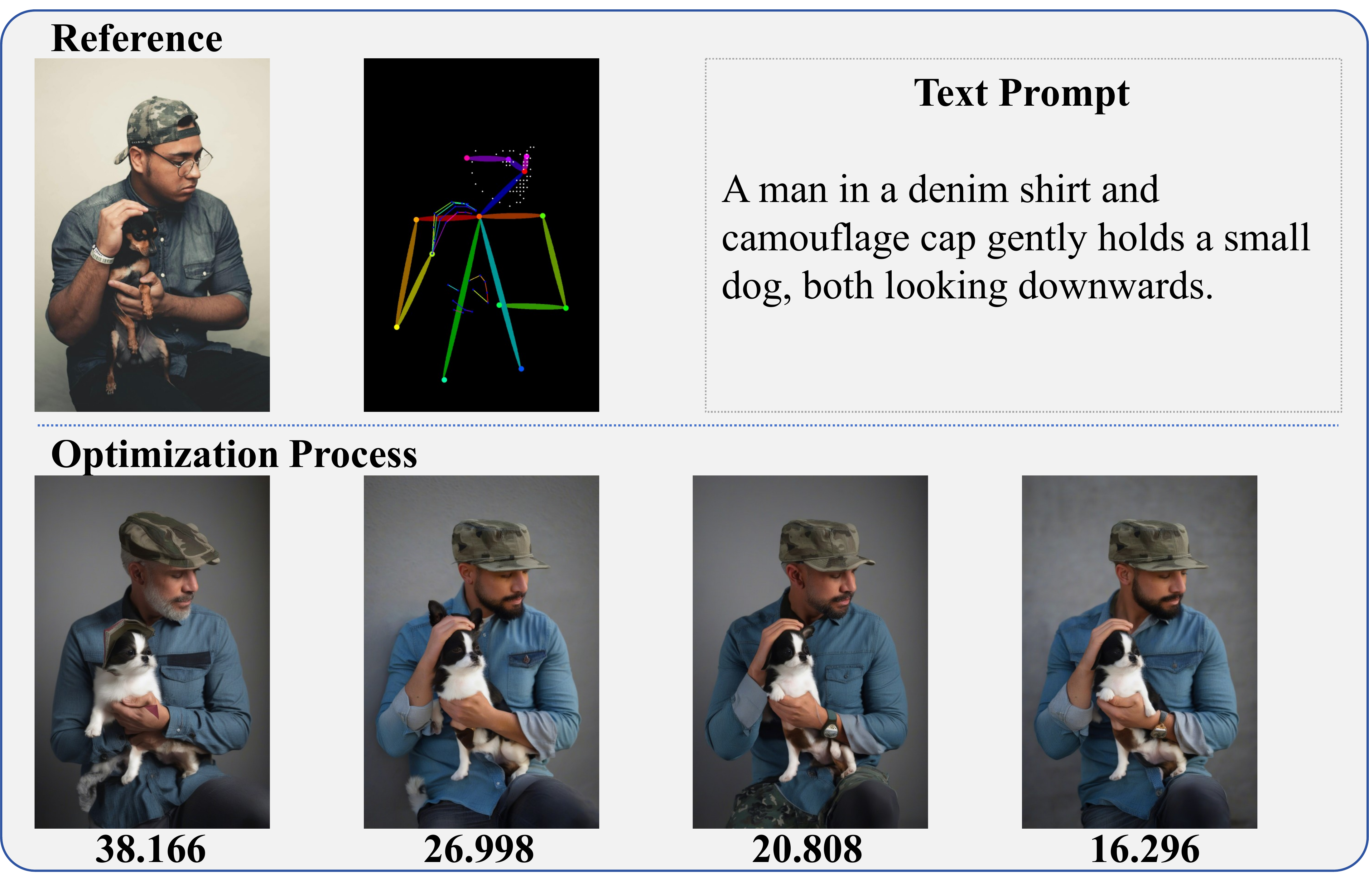}
        \caption{Pose}
        \label{fig:optimization_process_pose}
    \end{subfigure}
    \hfill
    \begin{subfigure}[b]{0.32\textwidth}
        \centering
        \includegraphics[width=\linewidth]{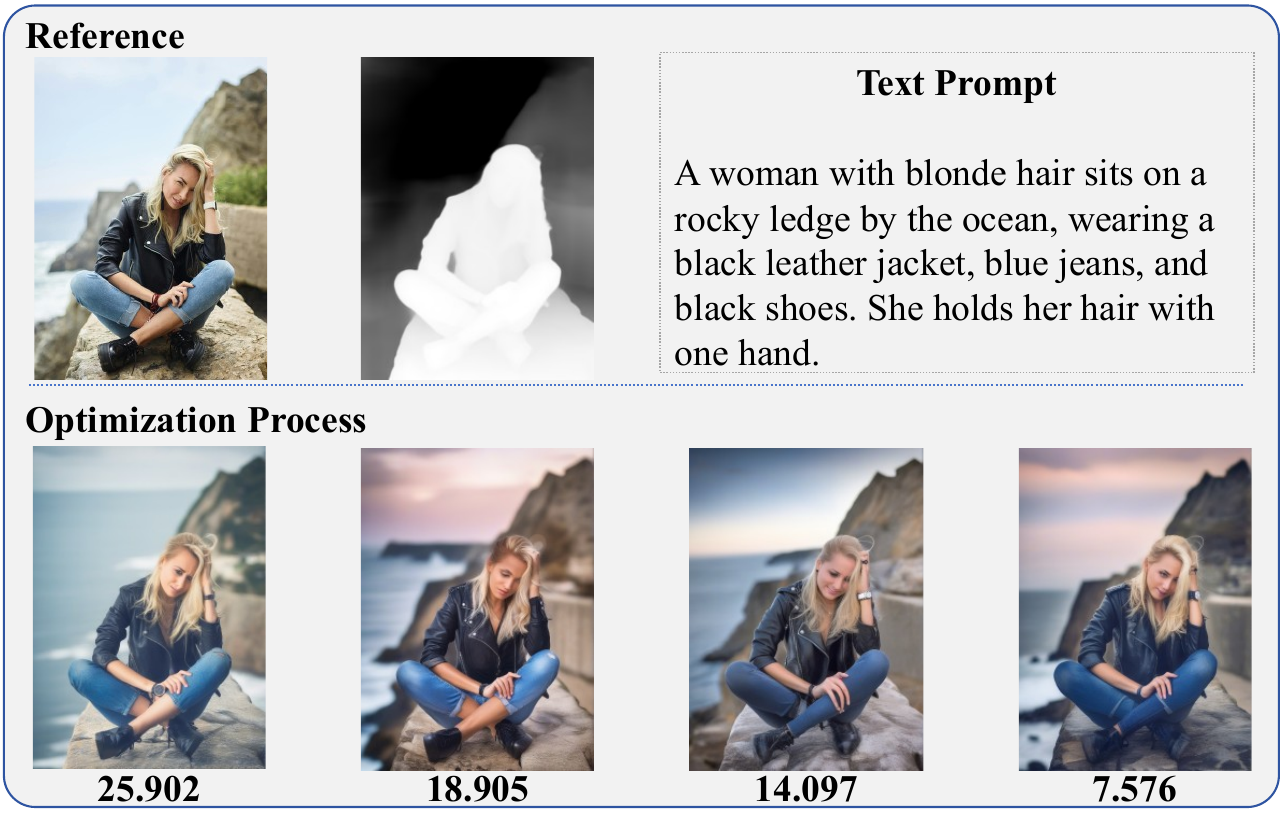}
        \caption{Depth}
        \label{fig:optimization_process_depth}
    \end{subfigure}
    
    \caption{Closed-loop optimization trajectories.
    From left to right within each panel, the controller iteratively refines the control signal to reduce the deviation from the reference condition.
    The improvement is reflected by the corresponding tracking metrics: similarity ($\uparrow$), MPJPE ($\downarrow$), and MAE ($\downarrow$).}
    \label{fig:optimization_process}
\end{figure}

Figure~\ref{fig:optimization_process} visualizes the intermediate generation results during closed-loop optimization.
Initial open-loop outputs, shown on the left of each sequence, typically exhibit noticeable deviations from the reference conditions.
Guided by the PID controller, the generated images are progressively refined over iterations.

For identity preservation, as shown in Fig.~\ref{fig:optimization_process}(a), the facial characteristics gradually move toward the reference identity.
For pose and depth control, as shown in Fig.~\ref{fig:optimization_process}(b,c), initial structural misalignments are corrected over successive iterations.
The corresponding metric scores, including similarity, MPJPE, and MAE, directly reflect this optimization process.

\newpage

\subsection{Additional Visualization Results}
\begin{figure}[t]
    \centering
    \includegraphics[width=0.93\linewidth]{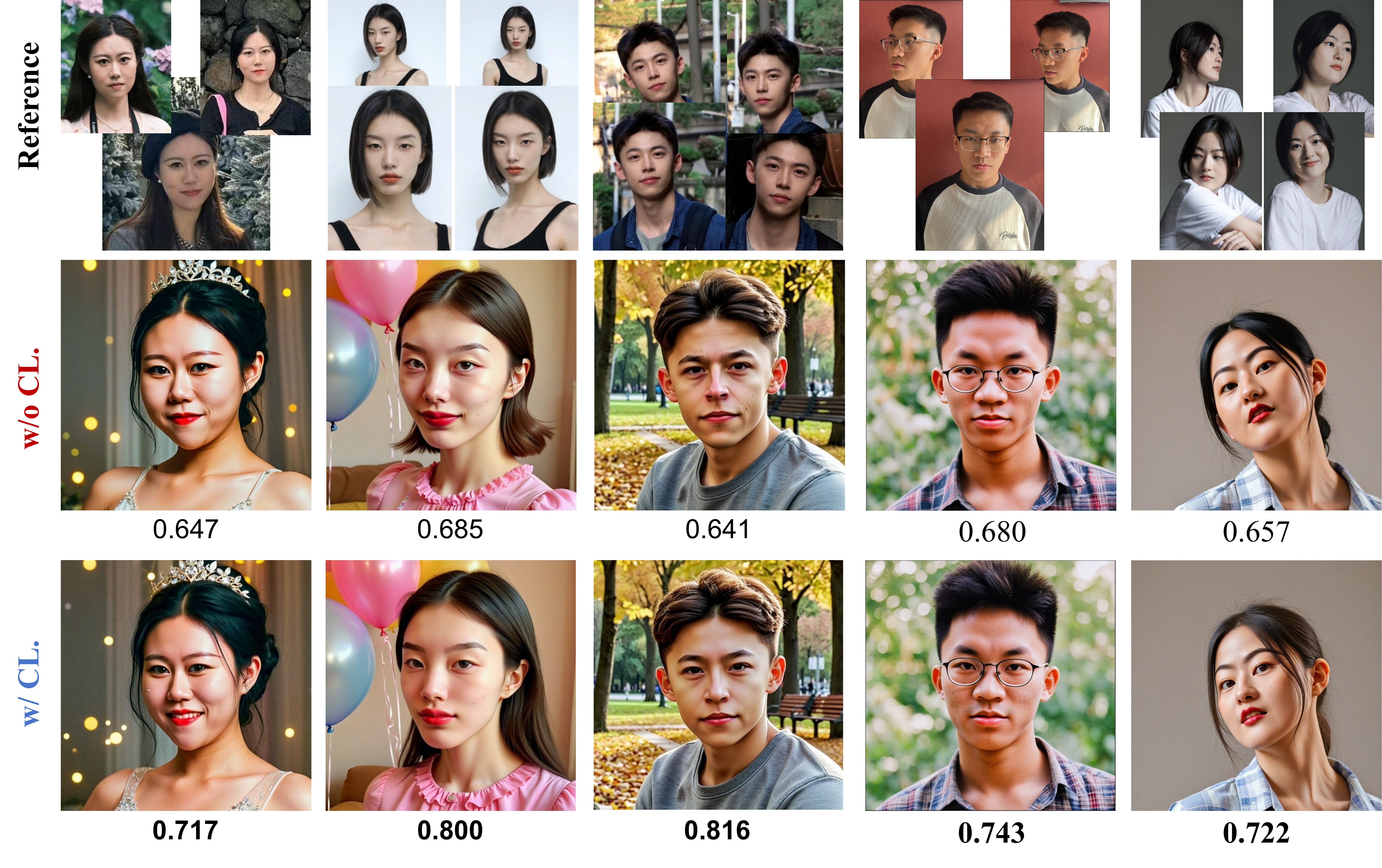}
    \caption{Qualitative comparison on ID-preserving portrait generation with IPA.}
    \label{fig:vis_portrait_sdxl2}
\end{figure}

\begin{figure}[t]
    \centering
    \includegraphics[width=0.93\linewidth]{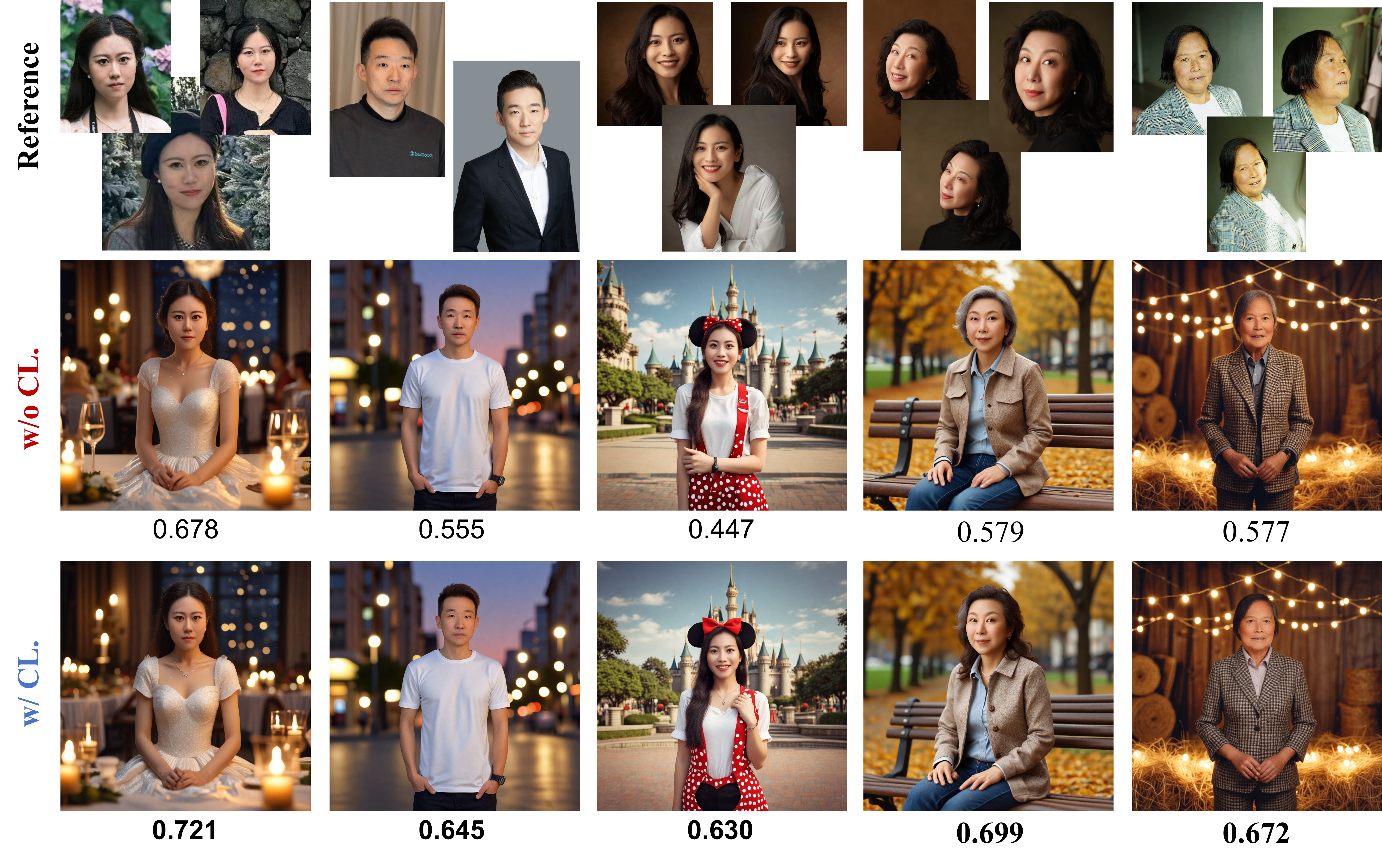}
    \caption{Qualitative comparison on ID-preserving portrait generation with PuLID.}
    \label{fig:vis_portrait_pulid}
\end{figure}

\begin{figure}[t]
    \centering
    \includegraphics[width=0.93\linewidth]{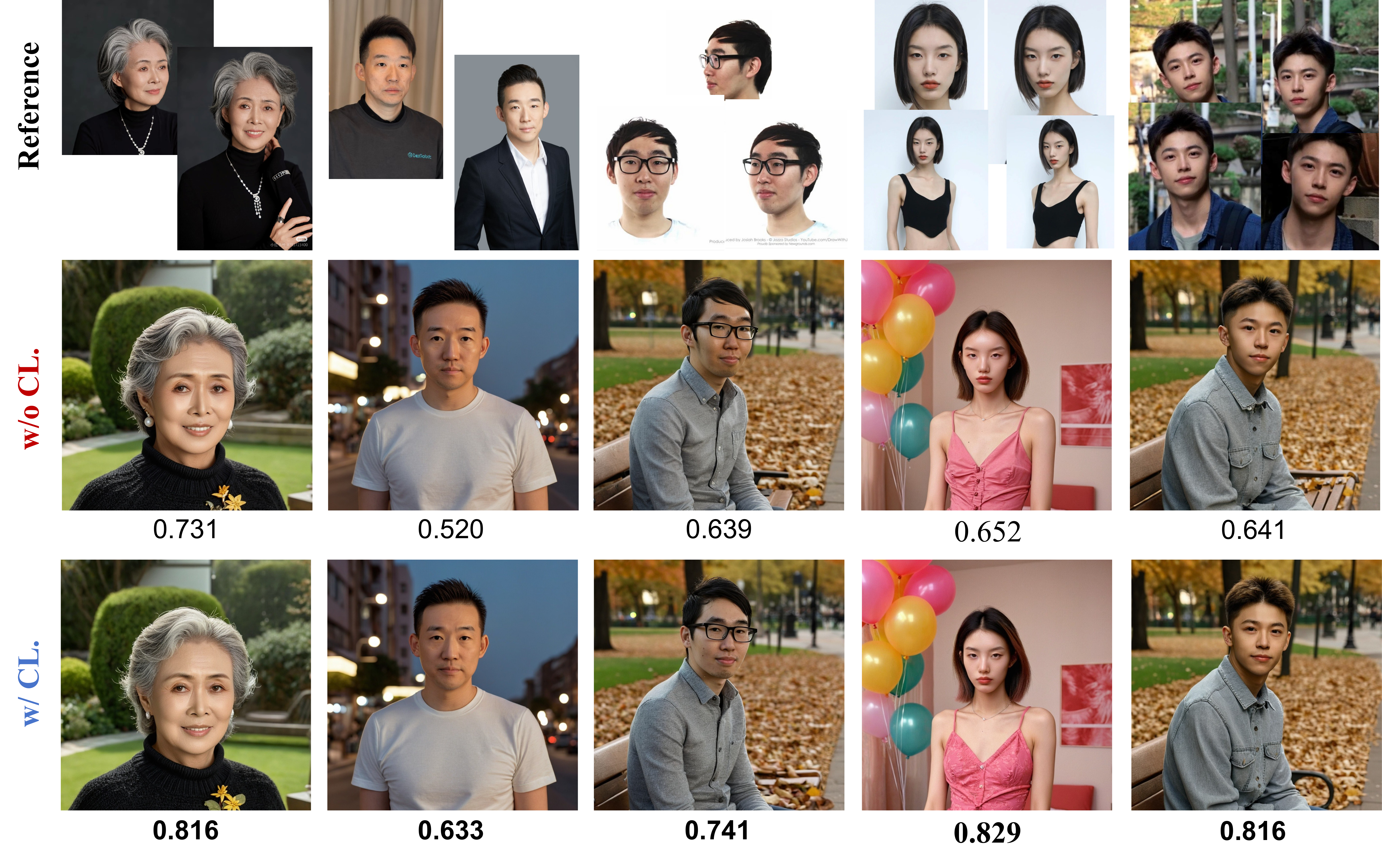}
    \caption{Qualitative comparison on ID-preserving portrait generation with PhotoMaker v2.}
    \label{fig:vis_portrait_pmv2}
\end{figure}

\begin{figure}[t]
    \centering
    \includegraphics[width=0.93\linewidth]{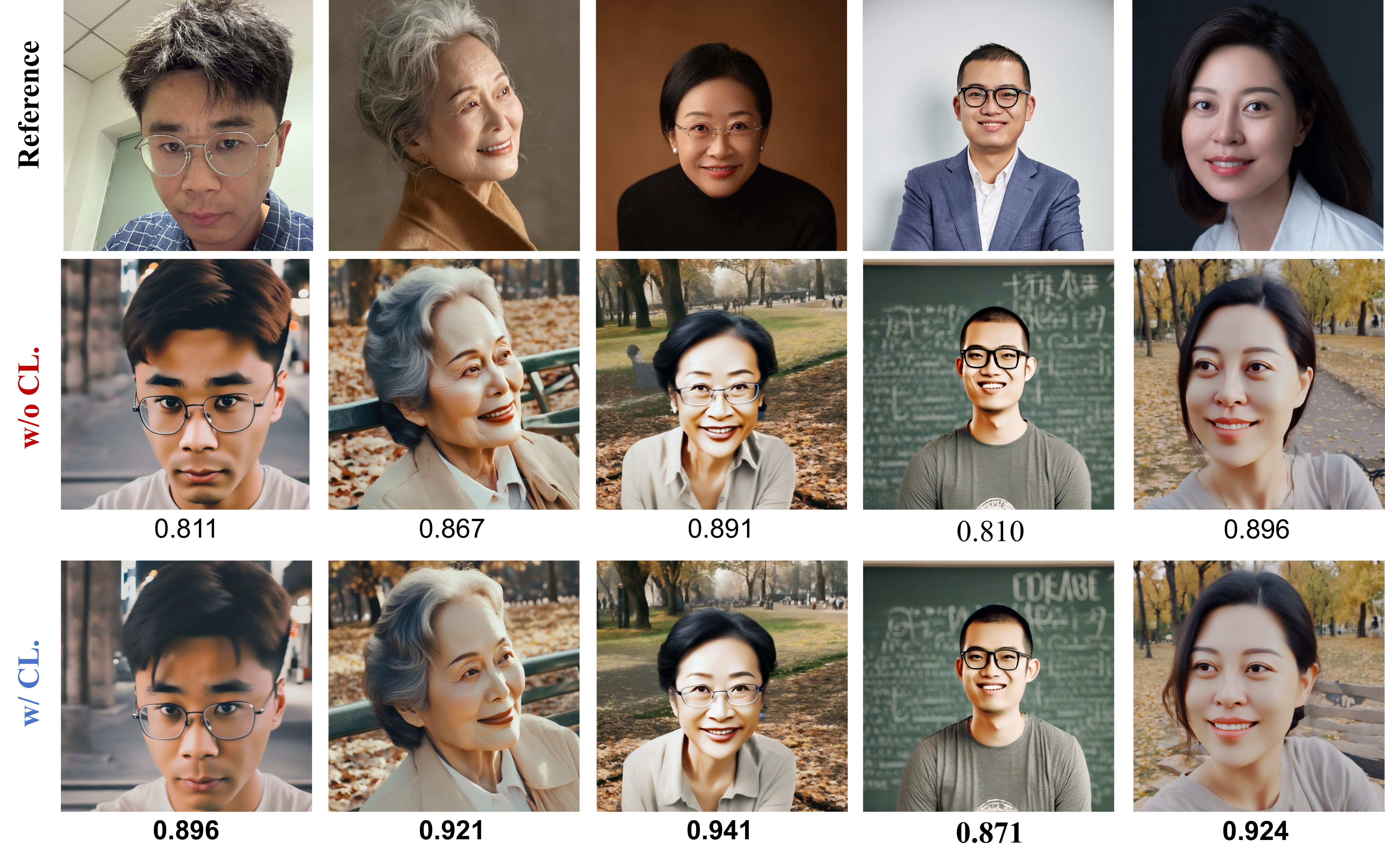}
    \caption{Qualitative comparison on ID-preserving portrait generation with InstantID.}
    \label{fig:vis_portrait_instantid}
\end{figure}

We provide additional qualitative results for identity preservation across IP-Adapter, PuLID, PhotoMaker v2, and InstantID in Figs.~\ref{fig:vis_portrait_sdxl2}--\ref{fig:vis_portrait_instantid}.
While open-loop baselines may suffer from identity degradation under complex prompts, our closed-loop optimization better preserves fine-grained facial characteristics.

\begin{figure}[t]
    \centering
    \begin{subfigure}[b]{0.45\textwidth}
        \centering
        \includegraphics[width=\textwidth]{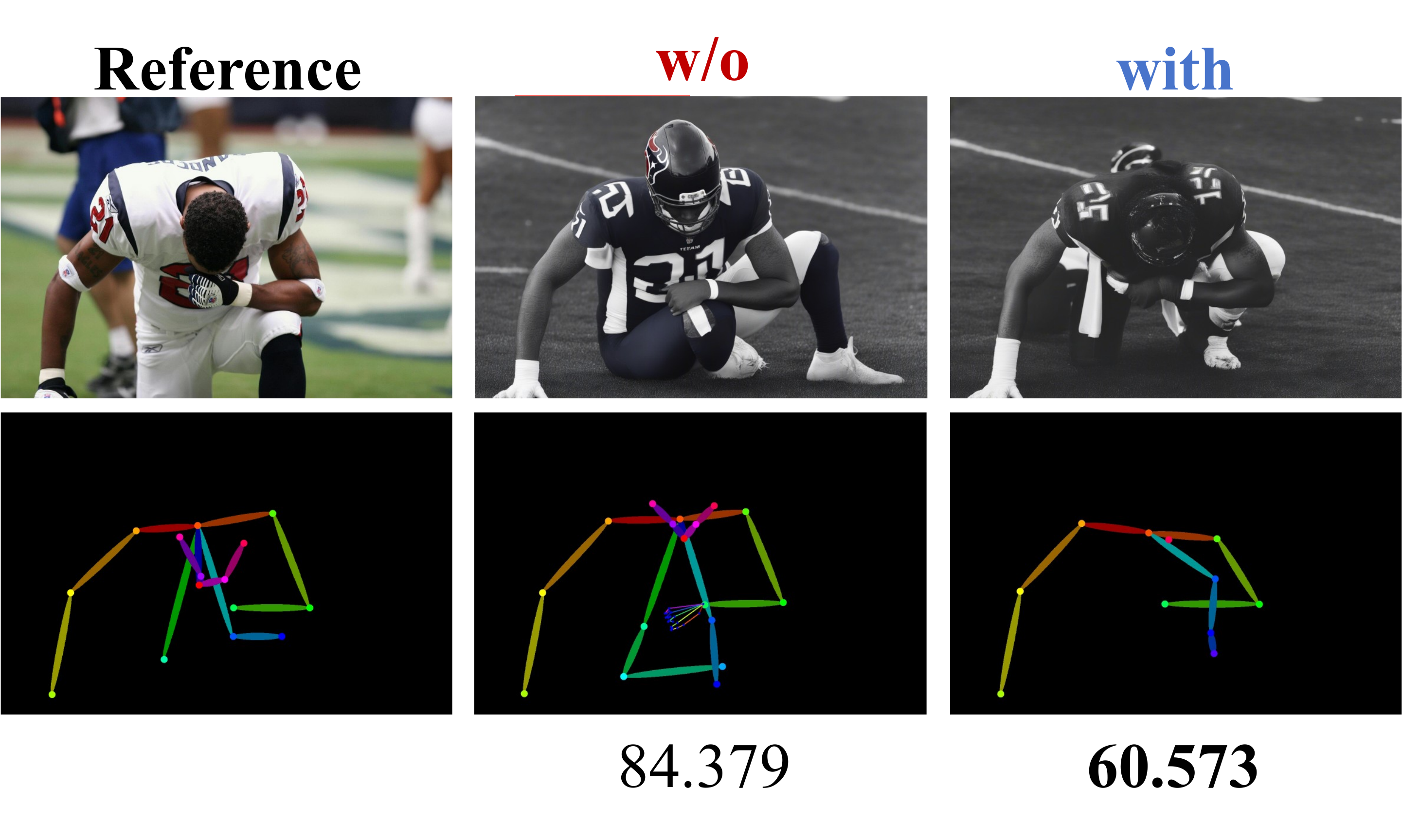}
    \end{subfigure}
    \hfill
    \begin{subfigure}[b]{0.45\textwidth}
        \centering
        \includegraphics[width=\textwidth]{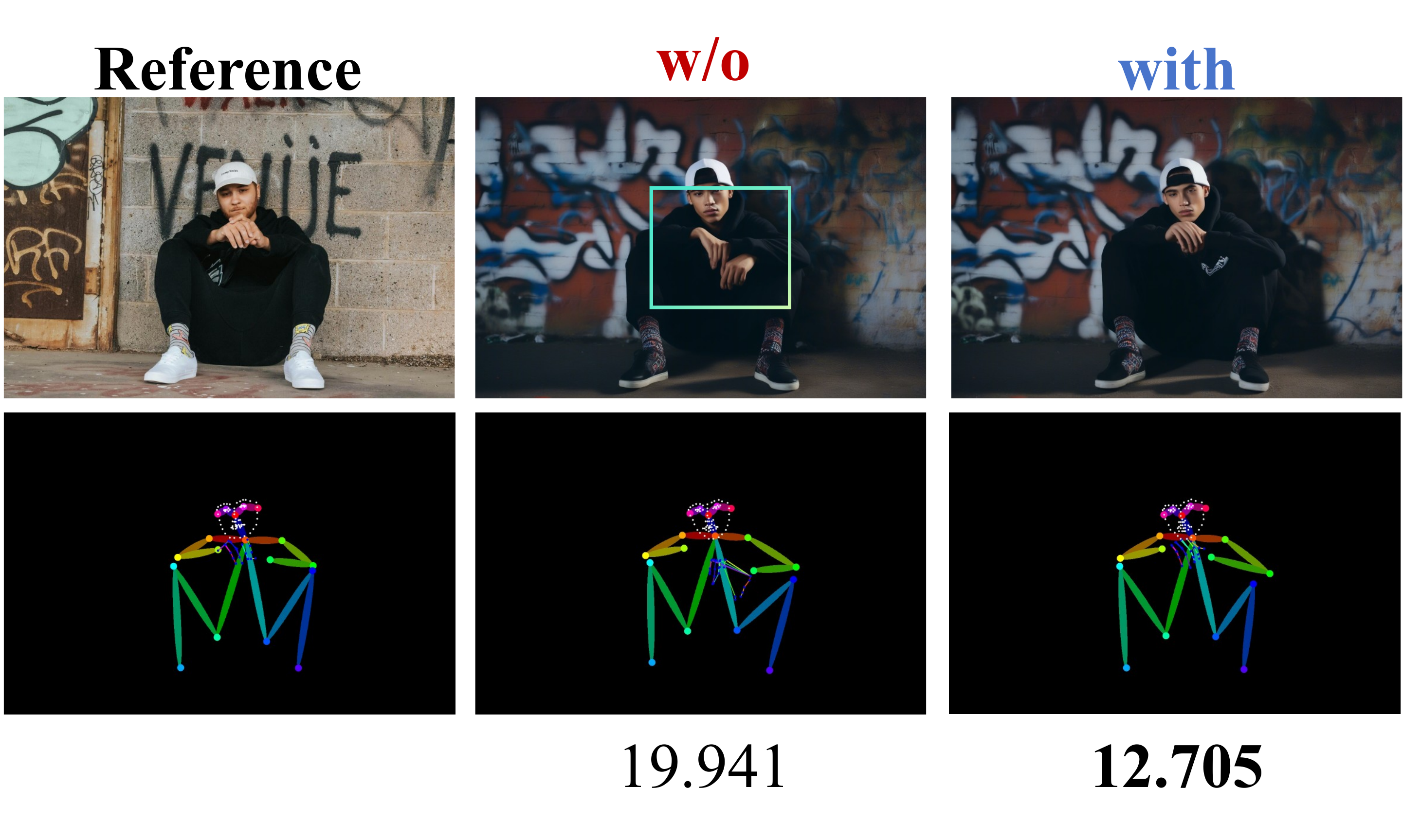}
    \end{subfigure}

    \vspace{-6pt}

    \begin{subfigure}[b]{0.3\textwidth}
        \centering
        \includegraphics[width=\textwidth]{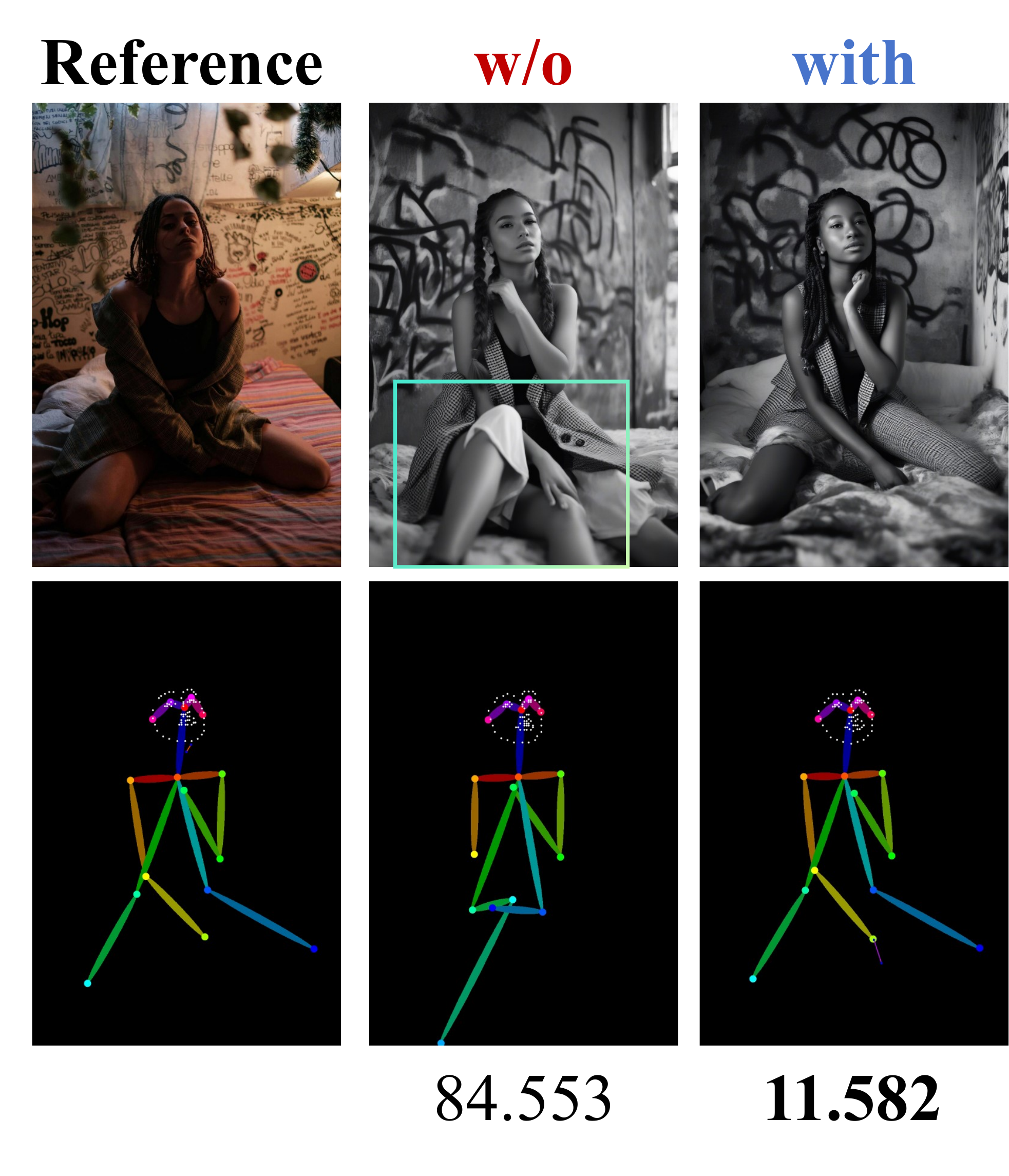}
    \end{subfigure}
    \hfill
    \begin{subfigure}[b]{0.3\textwidth}
        \centering
        \includegraphics[width=\textwidth]{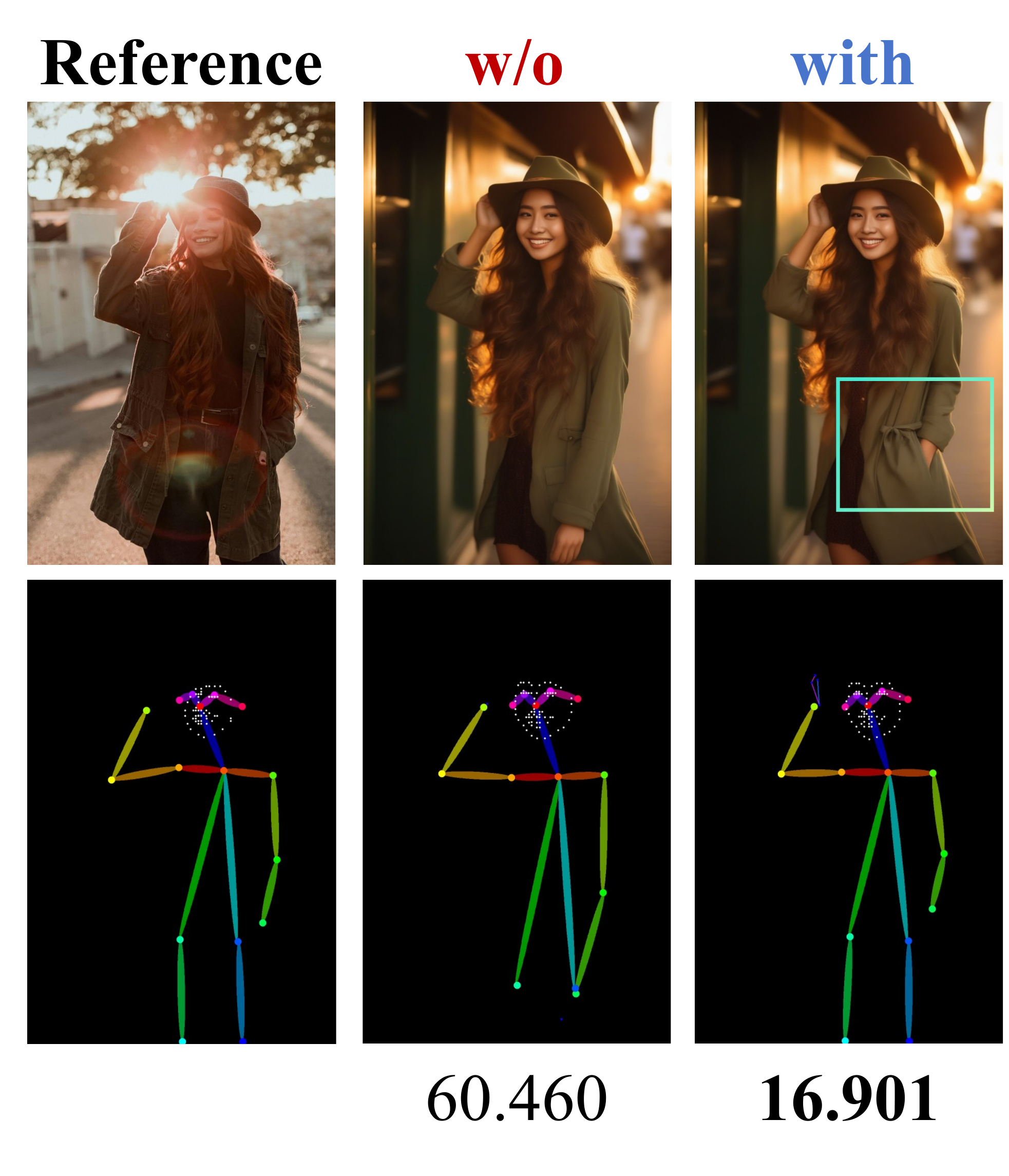}
    \end{subfigure}
    \hfill
    \begin{subfigure}[b]{0.3\textwidth}
        \centering
        \includegraphics[width=\textwidth]{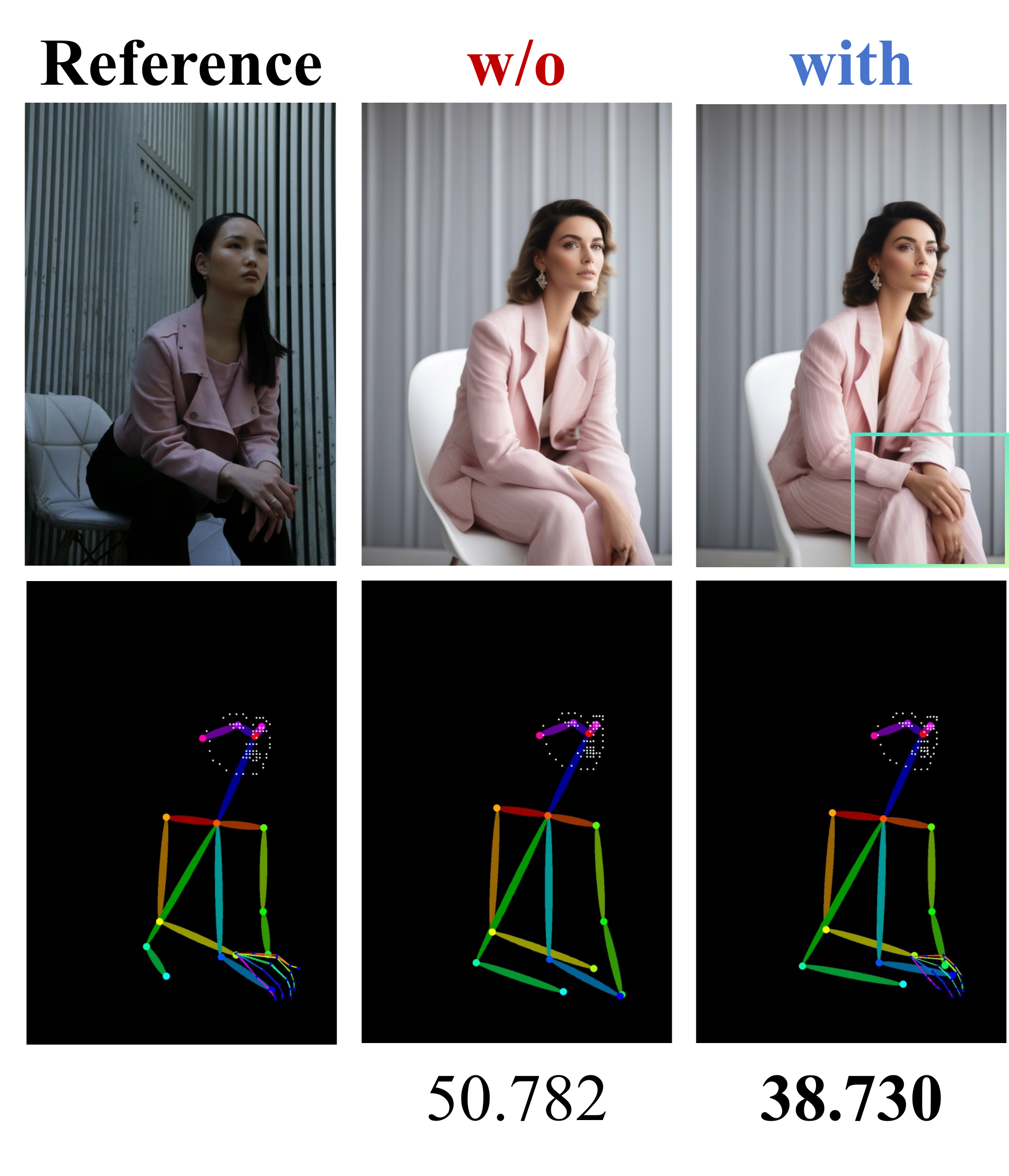}
    \end{subfigure}
    
    \caption{Qualitative comparison on pose-controlled generation.}
    \label{fig:vis_pose2}
    \vspace{-5pt}
\end{figure}

\begin{figure}[t]
    \centering
    \begin{subfigure}[b]{0.3\textwidth}
        \centering
        \includegraphics[width=\textwidth]{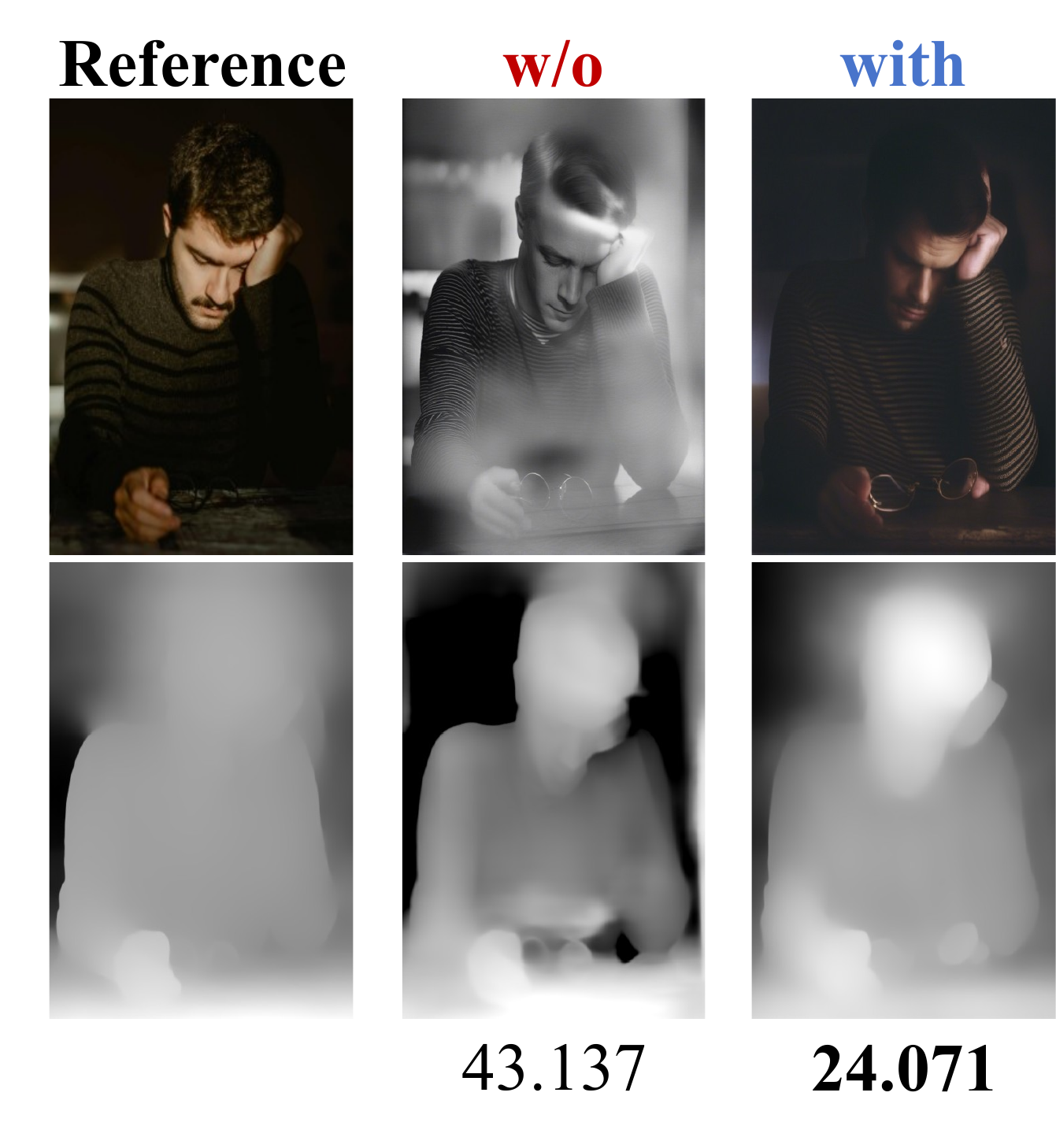}
    \end{subfigure}
    \hfill
    \begin{subfigure}[b]{0.3\textwidth}
        \centering
        \includegraphics[width=\textwidth]{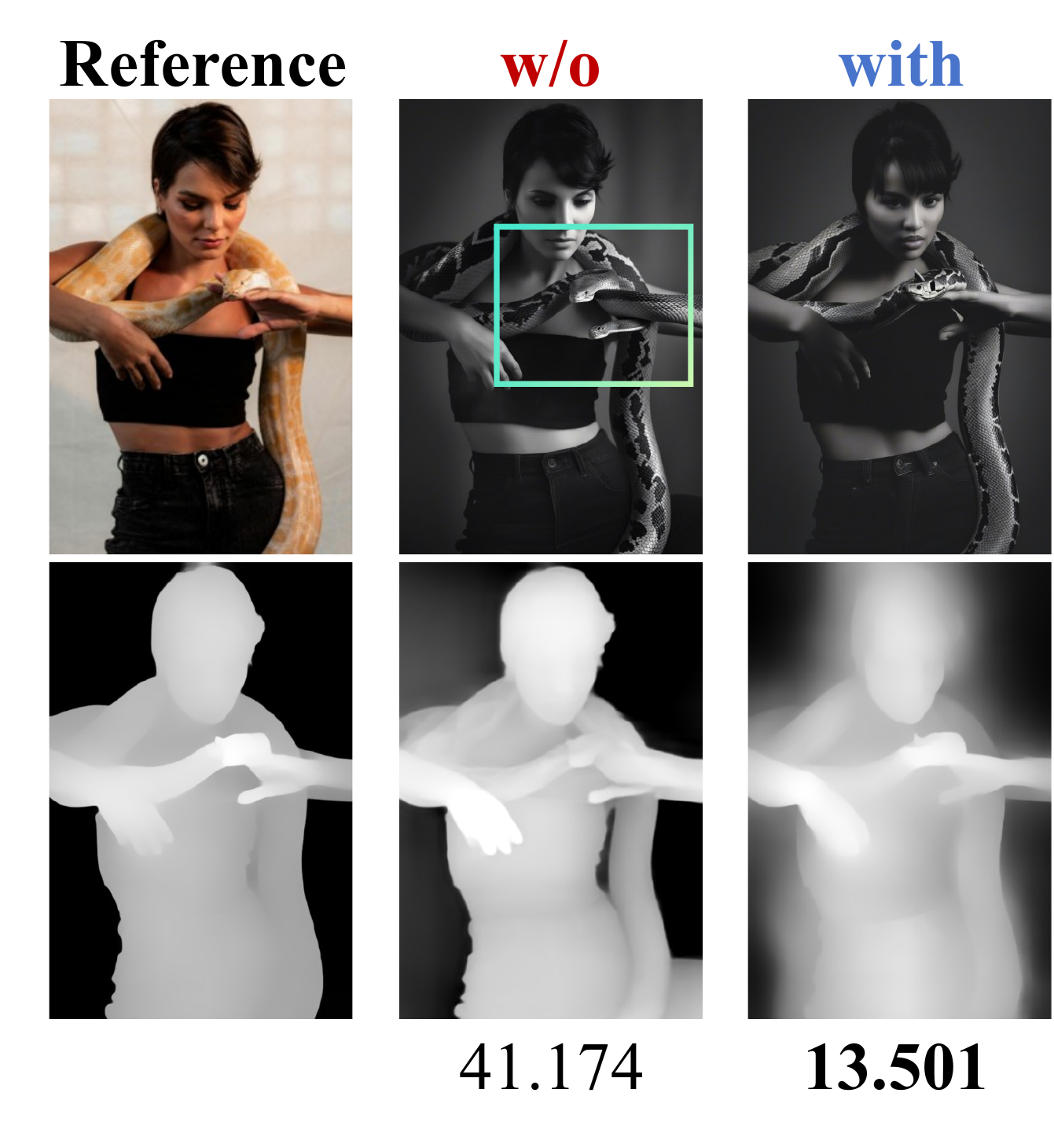}
    \end{subfigure}
    \hfill
    \begin{subfigure}[b]{0.3\textwidth}
        \centering
        \includegraphics[width=\textwidth]{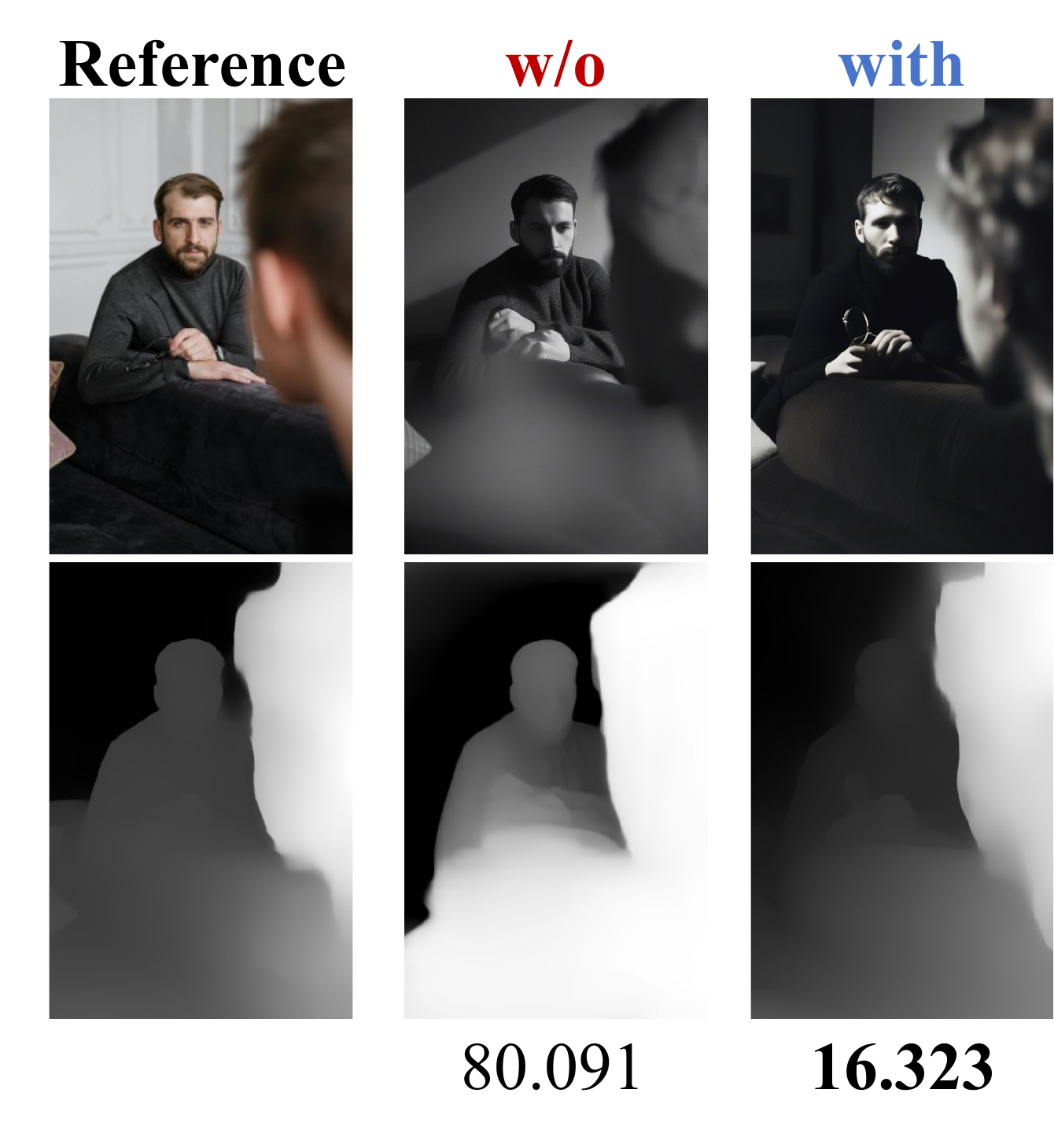}
    \end{subfigure}

    \vspace{-6pt}

    \begin{subfigure}[b]{0.3\textwidth}
        \centering
        \includegraphics[width=\textwidth]{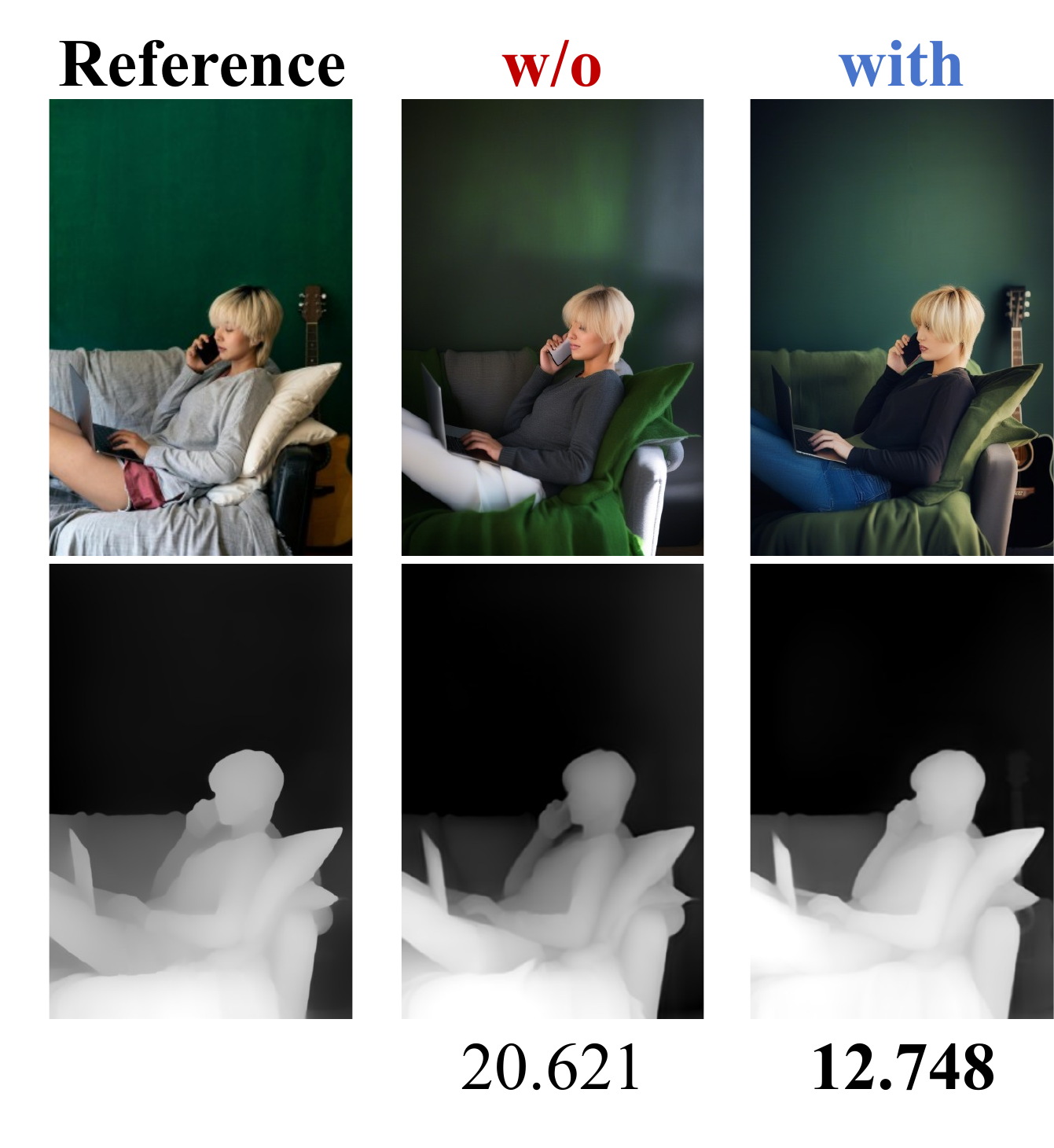}
    \end{subfigure}
    \hfill
    \begin{subfigure}[b]{0.3\textwidth}
        \centering
        \includegraphics[width=\textwidth]{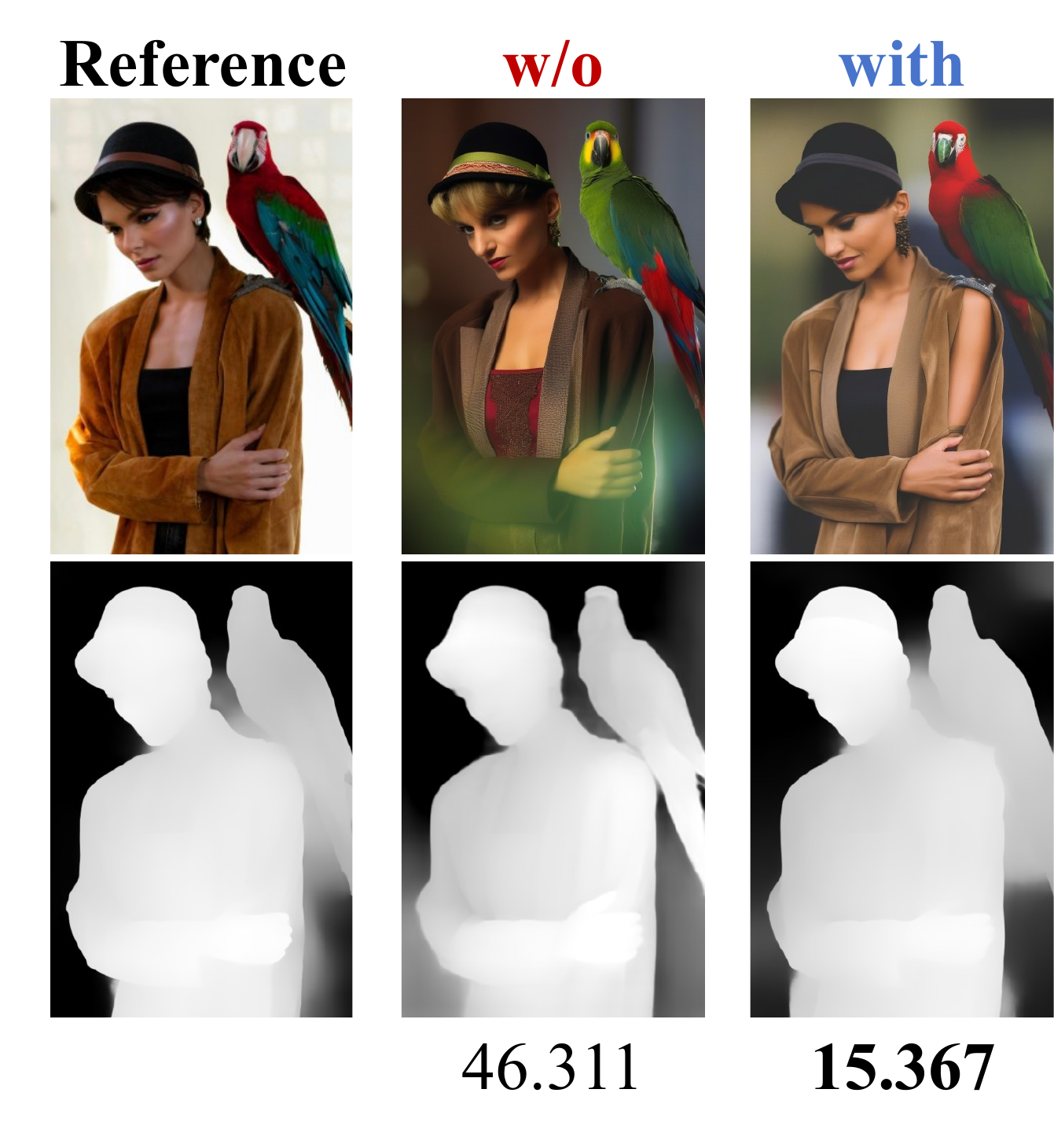}
    \end{subfigure}
    \hfill
    \begin{subfigure}[b]{0.3\textwidth}
        \centering
        \includegraphics[width=\textwidth]{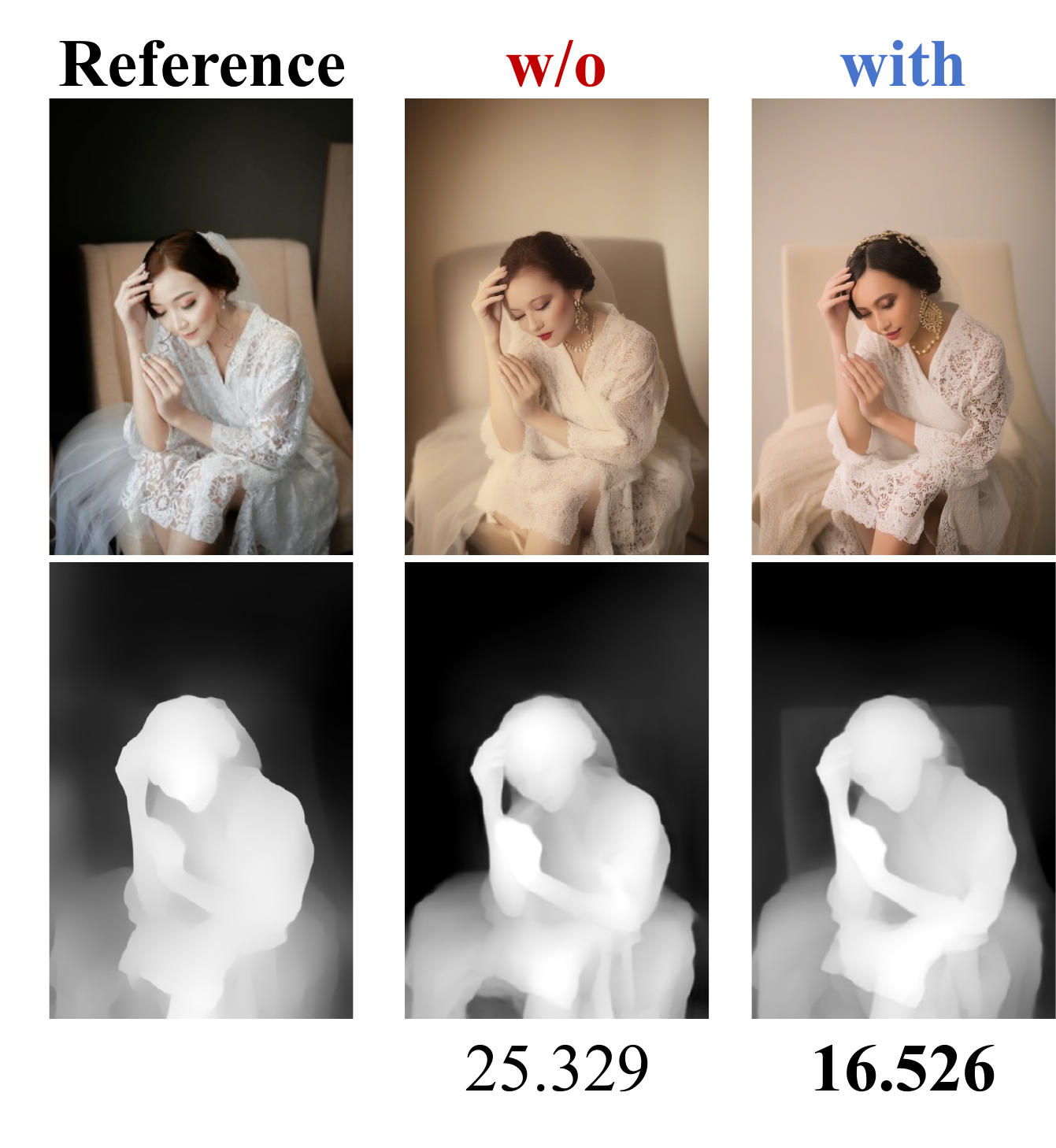}
    \end{subfigure}
    
    \caption{Qualitative comparison on depth-controlled generation.}
    \label{fig:vis_depth2}
\end{figure}

\begin{figure}[t]
    \centering
    \begin{subfigure}[b]{0.43\textwidth}
        \centering
        \includegraphics[width=\textwidth]{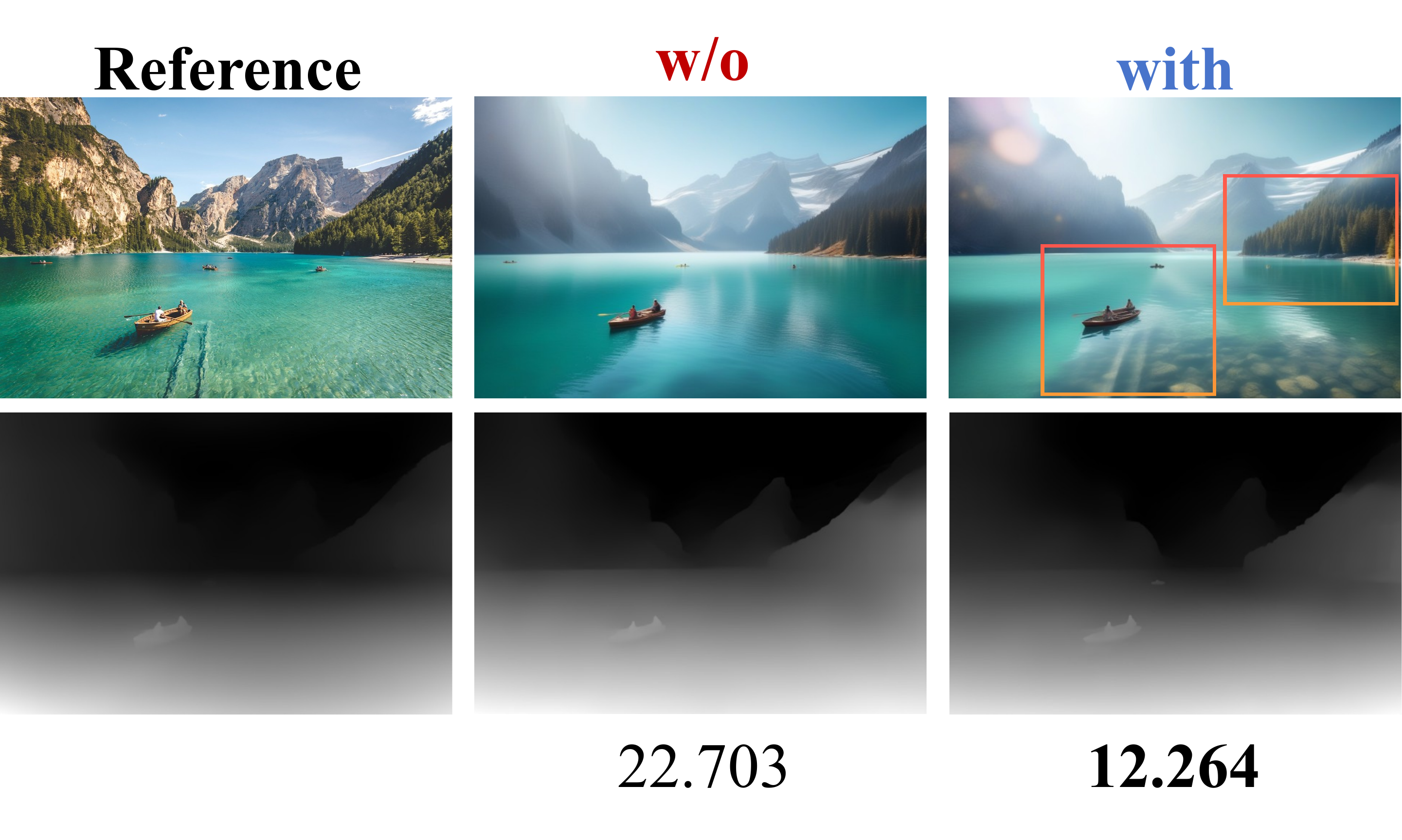}
    \end{subfigure}
    \hfill
    \begin{subfigure}[b]{0.43\textwidth}
        \centering
        \includegraphics[width=\textwidth]{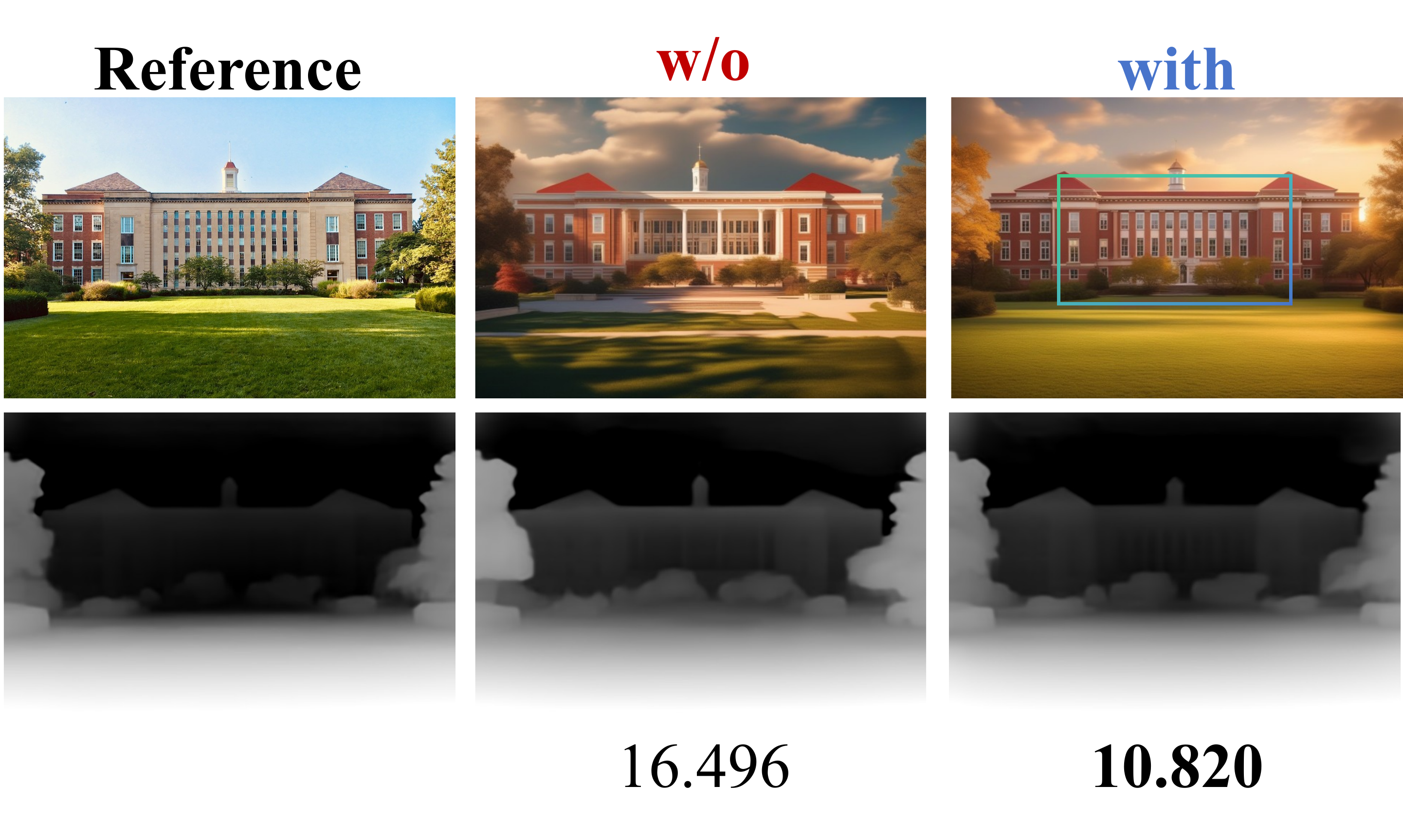}
    \end{subfigure}

    \vspace{-6pt}

    \begin{subfigure}[b]{0.3\textwidth}
        \centering
        \includegraphics[width=\textwidth]{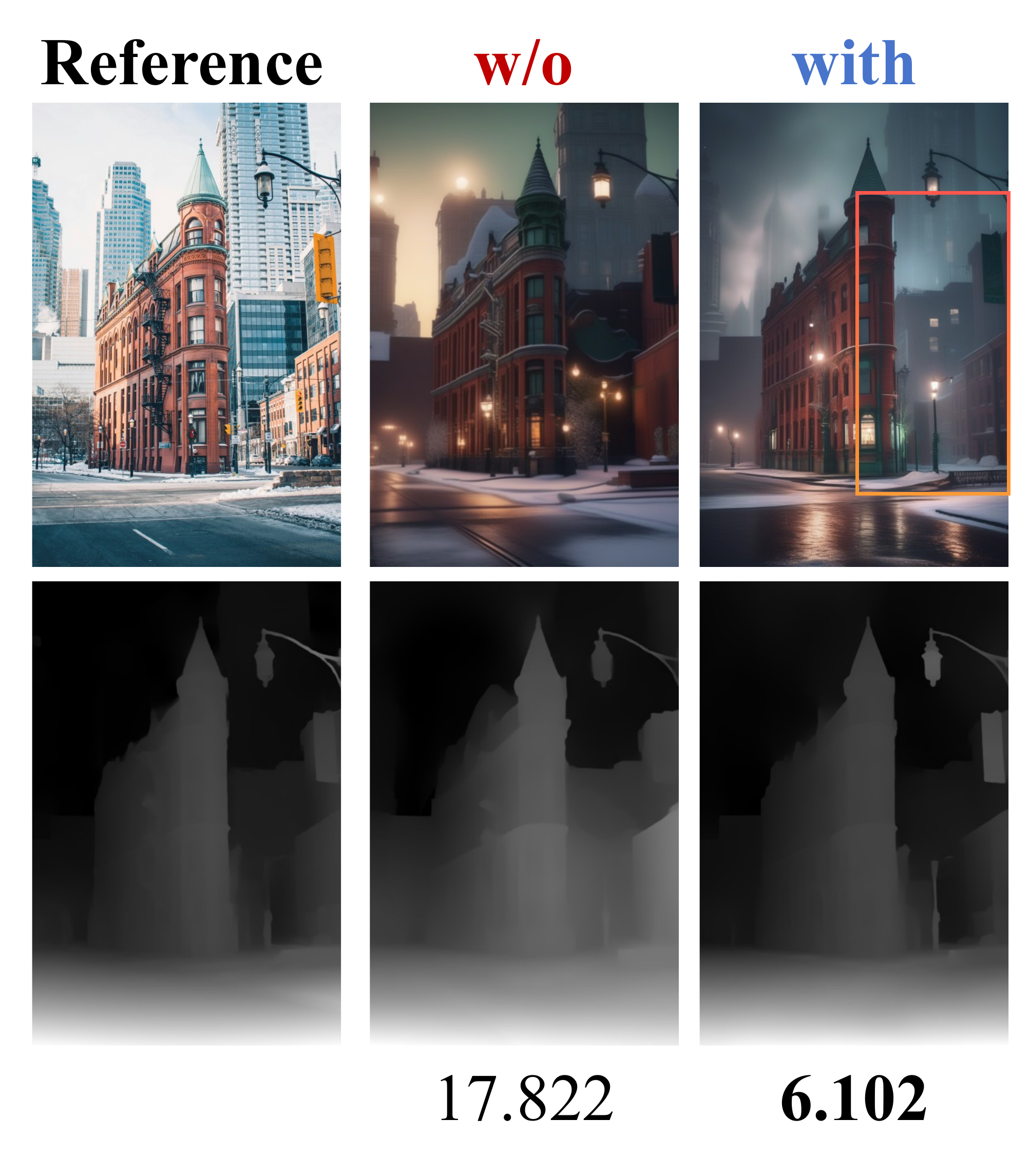}
    \end{subfigure}
    \hfill
    \begin{subfigure}[b]{0.3\textwidth}
        \centering
        \includegraphics[width=\textwidth]{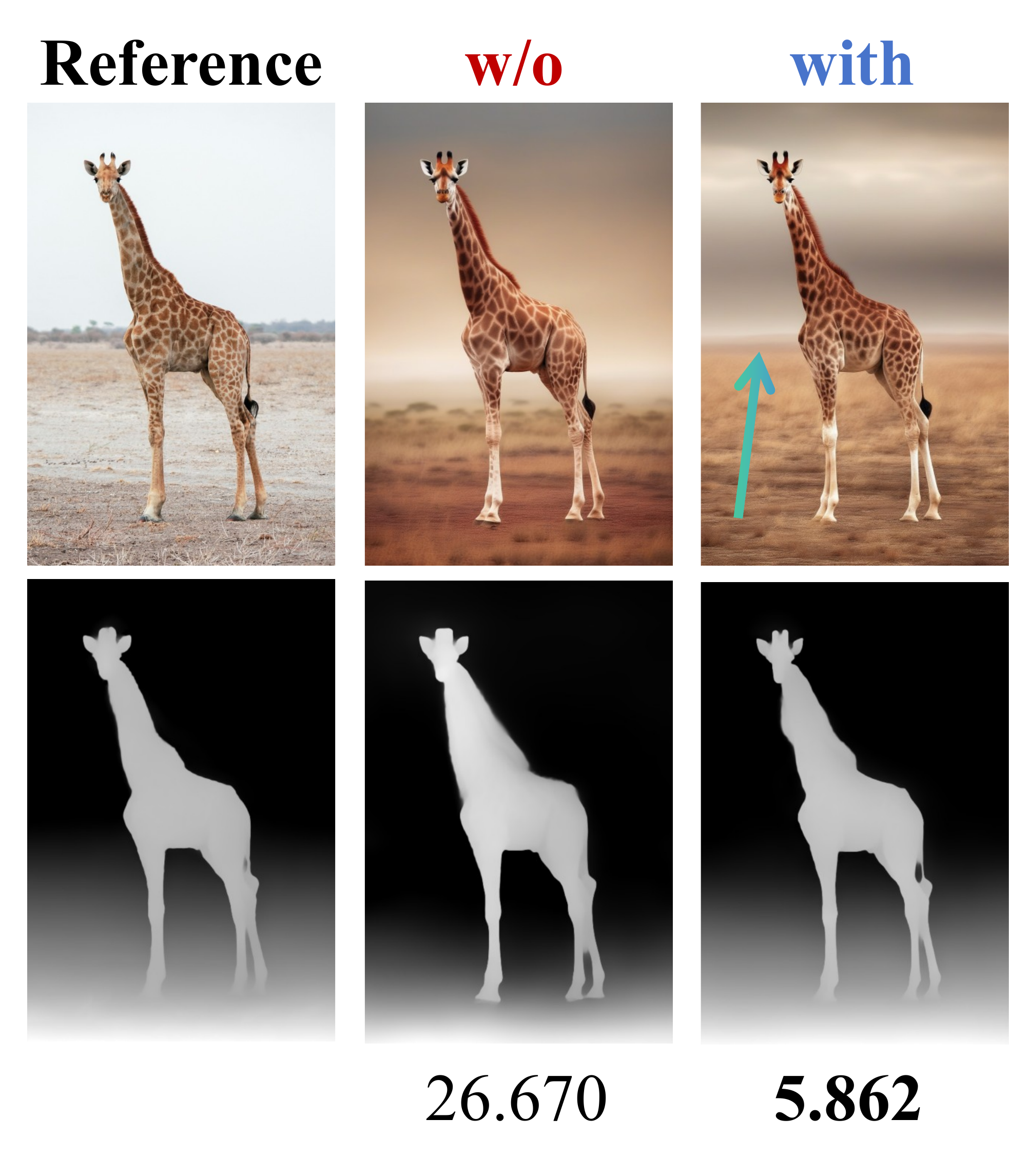}
    \end{subfigure}
    \hfill
    \begin{subfigure}[b]{0.3\textwidth}
        \centering
        \includegraphics[width=\textwidth]{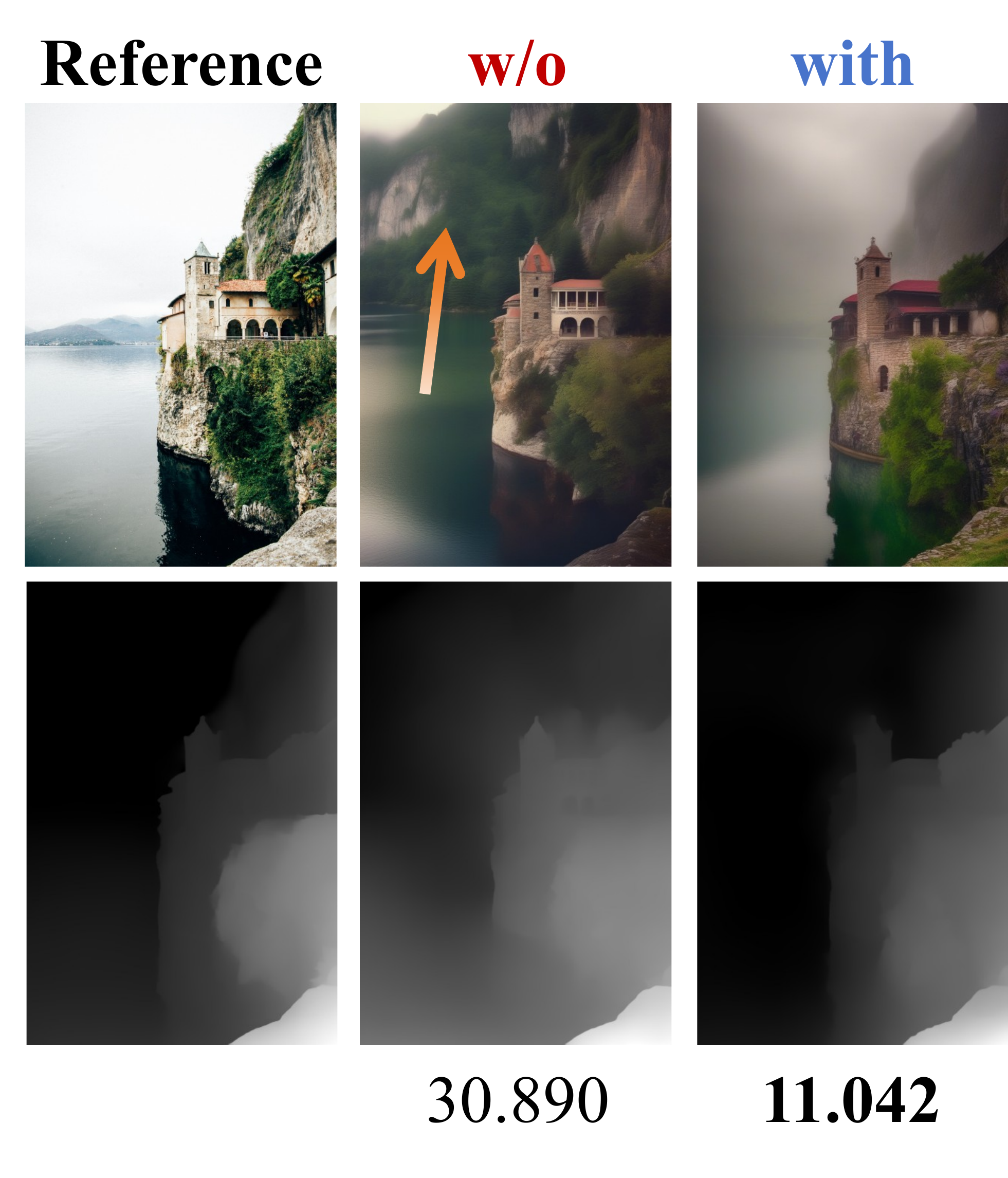}
    \end{subfigure}
    
    \caption{Qualitative comparison on depth-controlled generation with non-human samples.}
    \label{fig:vis_depth3}
\end{figure}

For spatial conditioning, supplementary comparisons of pose and depth control are presented in Figs.~\ref{fig:vis_pose2}, \ref{fig:vis_depth2}, and \ref{fig:vis_depth3}.
Our feedback mechanism rectifies typical feed-forward misalignments, such as unfaithful postures or distorted boundaries.
Figure~\ref{fig:vis_depth3} further extends the evaluation to non-human subjects, demonstrating structural alignment beyond human-centric generation.


